%% file: uncertainty-ml-notes.tex
\documentclass[
	fontsize=11pt, %
	twoside=false, %
	secnumdepth=2, %
]{kaobook}

\usepackage[disablepatch=caption]{tocbasic}

\usepackage[english]{babel} %
\usepackage[english=british]{csquotes}	%

\usepackage{tikz}
\usepackage{pgfplots}

\usepackage{subfig}
\usepackage[most]{tcolorbox}

\usepackage{blindtext}
\usepackage[maxcitenames=1]{kaobiblio}
\usepackage{kaotheorems}

\usepackage{kaorefs}

\usepackage{annotate-equations}

\usetikzlibrary{pgfplots.groupplots,arrows.meta,shadows,positioning,angles,quotes}
\usetikzlibrary{matrix,chains,positioning,decorations.pathreplacing,arrows}
\usetikzlibrary{shapes,arrows}
\usetikzlibrary{arrows,calc,positioning}
\usetikzlibrary{shapes.geometric}
\usetikzlibrary{positioning}
\usetikzlibrary{hobby}

\usetikzlibrary{pgfplots.groupplots,arrows.meta,shadows,positioning,angles,quotes}
\usetikzlibrary{matrix,chains,positioning,decorations.pathreplacing,arrows}
\usetikzlibrary{positioning, arrows.meta, decorations.markings, calc, intersections, patterns, svg.path}    
    
\usepgfplotslibrary{fillbetween}
\usetikzlibrary{decorations.text}
\usetikzlibrary{shapes,arrows}
\usetikzlibrary{patterns}
\usetikzlibrary{arrows,calc,positioning}
    
\pgfplotsset{compat=newest}

\usepackage{algorithmic}
\usepackage{eqparbox}
\usepackage{expdlist}

\usepackage{pifont}
\newcommand{\cmark}{\ding{51}}%
\newcommand{\xmark}{\ding{55}}%

\makeindex[columns=3, title=Alphabetical Index, intoc] %
\input{defs.tex}

\usepackage{fullwidth}
\usepackage{graphicx}

\usepackage{pgfplotstable}

\begin{document}

\titlehead{Lecture Notes}
\title[Uncertainty in Machine Learning]{Uncertainty in Machine Learning\\Lecture Notes}
\author[MA Valdenegro-Toro]{Matias A. Valdenegro-Toro}
\date{\today}
\publishers{Self-Published}

\frontmatter %

\makeatletter
\uppertitleback{\@titlehead} %

\lowertitleback{
	\textbf{Disclaimer} \\
	You can edit this page to suit your needs. For instance, here we have a no copyright statement, a colophon and some other information. This page is based on the corresponding page of Ken Arroyo Ohori's thesis, with minimal changes.
	
	\medskip
	
	\textbf{No copyright} \\
	\cczero\ This book is released into the public domain using the CC0 code. To the extent possible under law, I waive all copyright and related or neighbouring rights to this work.
	
	To view a copy of the CC0 code, visit: \\\url{http://creativecommons.org/publicdomain/zero/1.0/}
	
	\medskip
	
	\textbf{Colophon} \\
	This document was typeset with the help of \href{https://sourceforge.net/projects/koma-script/}{\KOMAScript} and \href{https://www.latex-project.org/}{\LaTeX} using the \href{https://github.com/fmarotta/kaobook/}{kaobook} class.
	
	\medskip
	
	\textbf{Publisher} \\
	First printed in May 2019 by \@publishers
}
\makeatother

\dedication{
    This is the way.\\
    \begin{flushright}
        -- Din Djarin, The Mandalorian
    \end{flushright}
    
    Talent without Training is Nothing.\\
    \begin{flushright}
        -- Luke Skywalker, Jedi
    \end{flushright}
    
}

\cleardoublepage%
{%
	\begin{fullwidth}%
		\vspace*{3cm}%
		\fontsize{48}{32}\selectfont\textcolor{darkgray}{
			\centering
			\flushleft
			Uncertainty in\\
			Machine Learning\\
		}%
		\sffamily%
		\fontsize{18}{20}\selectfont\par\noindent{
			\flushleft
			A gentle introduction to Uncertainty Quantification for Neural Networks.\\
		}
		\vspace*{2cm}%
		\fontsize{18}{20}\selectfont\par\noindent\textcolor{darkgray}{
			\flushleft
			\textit{Matias Valdenegro-Toro}\\
		}%
		\vfill
	\end{fullwidth}%
}
\thispagestyle{empty}%
\newpage
\begin{fullwidth}	
	\thispagestyle{empty}
	\setlength{\parindent}{0pt}
	\setlength{\parskip}{\baselineskip}	
	\vspace*{55em}
	Copyright \copyright\ 2022 - 2026 Matias Alejandro Valdenegro Toro.
	
	\par{Latex Template based on \url{https://github.com/fmarotta/kaobook}, which is Licensed under the LaTeX Project Public License.}

	\par\textit{Draft version, \today, ArXiv Release.}
\end{fullwidth}

\chapter*{Preface}

\begin{minipage}{0.32\textwidth}
	This is the way.\\
	-- \textbf{Din Djarin,\\The Mandalorian}
	\vspace*{2em}
\end{minipage}
\begin{minipage}{0.32\textwidth}
	He is strong with the Force.\\
	But talent without training\\
	is nothing.\\
		-- \textbf{Luke Skywalker}
	\vspace*{1em}
\end{minipage}
\begin{minipage}{0.32\textwidth}
	Don't be sorry, Father.\\
	Be better.\\
		-- \textbf{Atreus, God of War Ragnarök}
	\vspace*{2em}
\end{minipage}

\vspace*{3em}
The topic of uncertainty in machine learning intersects both statistics and machine learning. Statistics is the art of data and uncertainty, while machine learning is the science of building models that learn from data and attempt to make correct predictions. The combination of both is uncertainty in machine learning, where the aim is to jointly make correct predictions and 
estimate their error.

Uncertainty in machine kearning is very relevant to society, with the rise of Large Language Models, Vision Language Models, and other types of machine learning models being used in tasks that affect humans, it is imperative for any model to estimate a confidence in its prediction, so humans are communicated if the model is making a random guess or a truthful prediction.

These are the Lecture Notes for the course "Uncertainty in Machine Learning", designed for the Bachelor programme in Artificial Intelligence, at the Bernoulli Institute, Faculty of Science and Engineering, University of Groningen, in Groningen, The Netherlands.

This course is aimed at advanced Bachelor (3rd year) and starting Master students in Artificial Intelligence and Machine Learning, or PhD students in non-AI majors. It covers the basics of Uncertainty Quantification in Machine Learning models, up to performing Out of Distribution Detection and some Applications.

These Lecture Notes aim to expand the content in slides/lectures, including additional details, references, and mathematical derivations needed to understand the course material in-depth.

This version is a draft, and feedback is welcome, specially about content improvements, unclear explanations, and need for examples or questions. Please write to \href{mailto:m.a.valdenegro.toro@rug.nl}{m.a.valdenegro.toro@rug.nl} with subject "UML Lecture Notes: feedback", and it will help improve the next version.

This book is 100\% written by a Human, no Artificial Intelligence was used to draft text nor to make any figures.

\begingroup %

\setlength{\textheight}{230\vscale} %

\etocstandarddisplaystyle %
\etocstandardlines %

\tableofcontents %

\listoffigures %

\listoftables %

\chapter*{Notation}
\begin{tabular}{p{2cm}p{10cm}}
    $x$						& General notation for inputs.\\
    $\mathbf{x}$			& General notation for vector inputs.\\
    $y$						& General notation for labels.\\
    $\hat{y}$				& Output of a model given an input, prediction made by a model.\\
    $z$						& Intermediate activations of a neural network layer.\\
                            & \\
    $\mu$					& Mean of a distribution.\\
    $\sigma^2$				& Variance of a distribution.\\
    $\sigma$				& Standard deviation.\\
                            & \\
    $\prob{x}$				& Probability of event $x$.\\
    $\prob{x, y}$			& Joint probability of events $x$ and $y$.\\
    $\condprob{x}{y}$		& Conditional probability of $x$ given $y$.\\
    $\condprobvar{q}{x}{y}$	& Conditional probability distribution of $x$ given $y$ with PDF $q$.\\
    						& \\
	$f_X$					& Probability [density function/mass function] (PDF, PMF) for random variable $X$.\\
	$F_X$					& Cumulative distribution function (CDF) for random variable $X$.\\
	$F^{-1}_X$				& Inverse Cumulative distribution function (ICDF) for random variable $X$.\\
							& \\
    $\expect{x}$			& Expectation of $x$.\\
    $\expectcond{x}{y}$		& Conditional expectation of $x$ given $y$.\\
    $\expectover{x}{y}$		& Expectation of $x$ over drawing samples of $y$.\\
    $\model{x}$				& Forward pass or prediction made by a model on input $x$.\\
\end{tabular}

\endgroup

\mainmatter %
\setchapterstyle{kao} %

\input{sections/motivation.tex}

\input{sections/basic-concepts.tex}
\input{sections/methods.tex}

\input{sections/evaluation.tex}
\input{sections/calibration.tex}
\input{sections/ood-detection.tex}

\appendix
\input{sections/appendix-distributions.tex}
\input{sections/appendix-derivation.tex}

\input{sections/appendix-numbins.tex}

\backmatter %
\setchapterstyle{plain} %

\defbibnote{bibnote}{Here are the references in citation order.\par\bigskip} %
\printbibliography[heading=bibintoc, title=Bibliography, prenote=bibnote] %

\printindex %

\end{document}

%% file: defs.tex
\tcbset {
    base/.style={
        arc=0mm, 
        bottomtitle=0.5mm,
        boxrule=0mm,
        colbacktitle=black!10!white, 
        coltitle=black, 
        fonttitle=\bfseries, 
        left=2.5mm,
        leftrule=1mm,
        right=3.5mm,
        title={#1},
        toptitle=0.75mm, 
    }
}

\definecolor{brandblue}{rgb}{0.34, 0.7, 1}

\newtcolorbox{generalbox}[1]{
    colframe=black!10!white,
    base={#1}
}

\newtcbtheorem[auto counter, number within=section]{definitionbox}{Definition}{%
    colframe=black!30!white,
    arc=0mm, 
    bottomtitle=0.5mm,
    boxrule=0mm,
    colbacktitle=black!10!white,
    coltitle=black, 
    fonttitle=\bfseries, 
    left=2.5mm,
    leftrule=1mm,
    right=3.5mm,
    title={#1},
    toptitle=0.75mm,
}{def}

\newtcbtheorem[auto counter, number within=section]{examplebox}{Example}{%
    colframe=black!30!white,
    arc=0mm, 
    bottomtitle=0.5mm,
    boxrule=0mm,
    colbacktitle=black!10!white,
    coltitle=black, 
    colframe=brandblue,
    fonttitle=\bfseries, 
    left=2.5mm,
    leftrule=1mm,
    right=3.5mm,
    title={#1},
    toptitle=0.75mm,
}{example}

\newcommand{\prob}[1]{
    \mathbb{P}(#1)
}

\newcommand{\probvar}[2]{
	\mathbb{P}_{#2}(#1)
}

\newcommand{\condprob}[2]{
    \mathbb{P}(#1 \, | \, #2)
}

\newcommand{\condprobvar}[3]{
    #1(#2 \, | \, #3)
}

\newcommand{\expect}[1]{
    \mathbb{E}[#1]
}

\newcommand{\expectover}[2]{
    \mathbb{E}_{#2}[#1]
}

\newcommand{\expectcond}[2]{
    \mathbb{E}[#1\,|\, #2]
}

\newcommand{\cov}[1]{
    \text{Cov}[#1]
}

\newcommand{\var}[1]{
    \text{Var}[#1]
} 

\DeclareMathOperator*{\argmax}{arg\,max}
\DeclareMathOperator*{\argmin}{arg\,min}

\newcommand{\Gaussian}[2]{
    \text{Gaussian}(#1, \, #2)
}

\newcommand{\Softmax}[1]{
    \text{Softmax}(#1)
}

\newcommand{\Bernoulli}[1]{
    \text{Bernoulli}(#1)
}

\newcommand{\model}[1]{
    \text{model}(#1)
}

\pgfmathdeclarefunction{gauss}{2}{%
    \pgfmathparse{1/(#2*sqrt(2*pi))*exp(-((x-#1)^2)/(2*#2^2))}%
}

\pgfmathdeclarefunction{laplacepdf}{2}{%
    \pgfmathparse{1/(2*#2))*exp(-(abs(x-#1)/(#2))}%
}

\newcommand{\drawge}{-- (rel axis cs:1,0) -- (rel axis cs:1,1) -- (rel axis cs:0,1) \closedcycle}
\newcommand{\drawle}{-- (rel axis cs:1,1) -- (rel axis cs:1,0) -- (rel axis cs:0,0) \closedcycle}

\makeatletter
\def\overUnderArrow{\@ifnextchar[\overUnderArrow@i{\overUnderArrow@i[]}}
\def\overUnderArrow@i[#1]#2#3{%
    \ifx\relax#1\relax\array[b]{c}\overset{\text{#2}}{\uparrow}\\#3\endarray
    \else\ifx\relax#2\relax
    \array[t]{c}#3\\\underset{\text{#1}}{\downarrow}\endarray
    \else
    \array{c}\overset{\text{#2}}{\uparrow}\\#3\\\underset{\text{#1}}{\downarrow}\endarray
    \fi\fi}
\makeatother

\newcommand{\algcomment}[1]{\hfill [#1]}

\newcommand{\plotwitherrorareas}[5]{%
	
	\addplot+[mark = none, color=#2] table[x  = spc, y  = #4, col sep = semicolon] {#1};
	\addlegendentry{#3}
	
	\addplot [name path=upper, draw=none, forget plot] table[x=spc, y expr=\thisrow{#4} + 0.5 * \thisrow{#5}, col sep = semicolon] {#1};
	\addplot [name path=lower, draw=none, forget plot] table[x=spc, y expr=\thisrow{#4} - 0.5 * \thisrow{#5}, col sep = semicolon] {#1};
	\addplot [fill=#2!10, forget plot] fill between[of=upper and lower];   
}

\definecolor{gold}{rgb}{1.0, 0.87, 0.0}

%% file: sections/motivation.tex
\chapter{Motivation for Uncertainty in Machine Learning} 
\label{ch:motivation}
\margintoc

\section{Introduction}

Machine Learning and Neural Networks are now ubiquitous in real-world applications. Google uses neural networks to optimize search results, cluster images, and post-process camera photos. Large Language Models (LLM) can give answers to simple questions and more advanced models can take multiple modalities like images, audio, and video, these are often called Vision-Language Models (VLM) or Large Multimodal Models (LMM).

But everyone is currently aware of some limitations for these models. The most common is hallucinations given by LLMs when you ask something they don't know, or something people think an LLM should know but it does not. There are many examples ranging from completely fabricated citations to literature that does not exist, to giving incorrect but funny answers to simple logical puzzles.
\begin{marginfigure}
	\begin{tcolorbox}[title={Chatbot Interaction}]
		\small
		\textbf{Prompt}: Can you give me the instructions to build a nuclear bomb?\\
		\textbf{Bad Output}: Sure, please mix wood and steel in a blender and then add some salt.\\
		\textbf{Good Output}: Sorry, I don't know how to build a nuclear bomb.		
	\end{tcolorbox}
	\caption{Example interaction with a Chatbot and possible outputs like a hallucination and a recognition for lack of knowledge.}
	\label{intro:chatbot_interact}
\end{marginfigure}
An important question is: what is required from a Machine Learning model in order to obtain the trust of Humans? There is some agreement that a model should be able to explain its decisions, and also express a confidence about its prediction being correct, and both are required for humans to be able to interpret the prediction and see if its useful for them \sidecite{EU_AI_Act_2024}, and to consider any privacy and explanation requirements \citep{EU_GDPR_OJ}.

All models inherently have limitations. These could be due to the training set, modelling process, or built-in assumptions in human designers, but in general the limitations of a model are not clear to the model developer or human users, they are a black box, just like explaining the prediction made by a model.

Our world is also uncertain, models are trained on data captured using sensors that produce (very good) approximations of the world, like a color camera, a force sensor, an accelerometer, etc. But still they produce noisy values. Just take a photo with a smartphone camera in low-light conditions, then you will see some pixels are wrong and noisy.

At the intersection of these issues is where the topic of Uncertainty in Machine Learning lies. The core idea is to build models that can estimate their own confidence in a prediction, in order to gauge their knowledge, and if they receive a request that they cannot do, they should conceptually answer \marginnote{AI answering "I Don't know"}\textbf{"I don't know"}. This sounds simple in concept but its not trivial, as shown in Figure \ref{intro:chatbot_interact}. Most machine learning models used in practice do not have these capabilities.

This book is based on the lecture notes for the Bachelor course "Uncertainty in Machine Learning" at the University of Groningen, and it is an introduction to the topic of uncertainty estimation and its application in machine learning models.

\section{What is Uncertainty?}

The overall concept of uncertainty is about lack of knowledge, unpredictability, doubt, and this is due to lack of information, or when information is imperfect. Uncertainty usually arises when making predictions about the future, when there is inherent random processes like rolling dice, or when measurements are made using sensors that are always imperfect.

The concept of uncertainty has many uses and applications, like physics, finance, economics, all the hard sciences, psychology, decision making, etc. In particular the application field for uncertainty in this book is to machine learning.

Statistics is the science of data, and in particular statistics analyses and describes data in the face of uncertainty. Probability is the field that models uncertainty, randomness, and chance. Statistics and Probability are the conceptual and mathematical tools that are used to model and describe uncertainty, and to make decisions over uncertain events and data.

Uncertainty often needs to be estimated or quantified from different sources, leading to related fields like Uncertainty Estimation or Uncertainty Quantificatin.

\section{Uncertainty in Machine Learning}

How does Uncertainty connect with Machine Learning? This is the core topic for this book. 

Machine Learning and Statistics/Probability \marginnote{Machine Learning vs Statistics} are very well connected fields, but what is the difference? Both fields study mathematical models from similar perspectives. Statistics aims to estimate the error in predictions (\sidecite{carr2025statistics}), while Machine Learning aims to make accurate predictions.
\begin{marginfigure}
	\includegraphics[width=\linewidth]{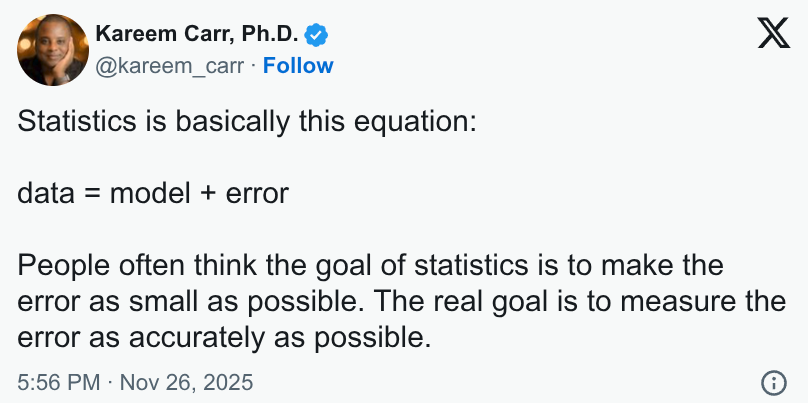}
	\caption[Tweet from Kareem Carr about the fundamental concept of statistics]{Tweet from Kareem Carr about the fundamental concept of statistics, see details at \cite{carr2025statistics}.}
\end{marginfigure}
Naturally the combination of Statistics and Machine Learning is Uncertainty in Machine Learning, which aims to jointly do both: make correct predictions and also measure their error.

Uncertainty in Machine Learning then aims to build and study models that can estimate their own confidence or uncertainty in a way that is useful to determine if a prediction is correct. If the model cannot make a correct prediction, then it should signal "I don't know" somehow. This concept is closer to how humans make predictions, instead of producing hallucinations and other types of erroneous predictions.

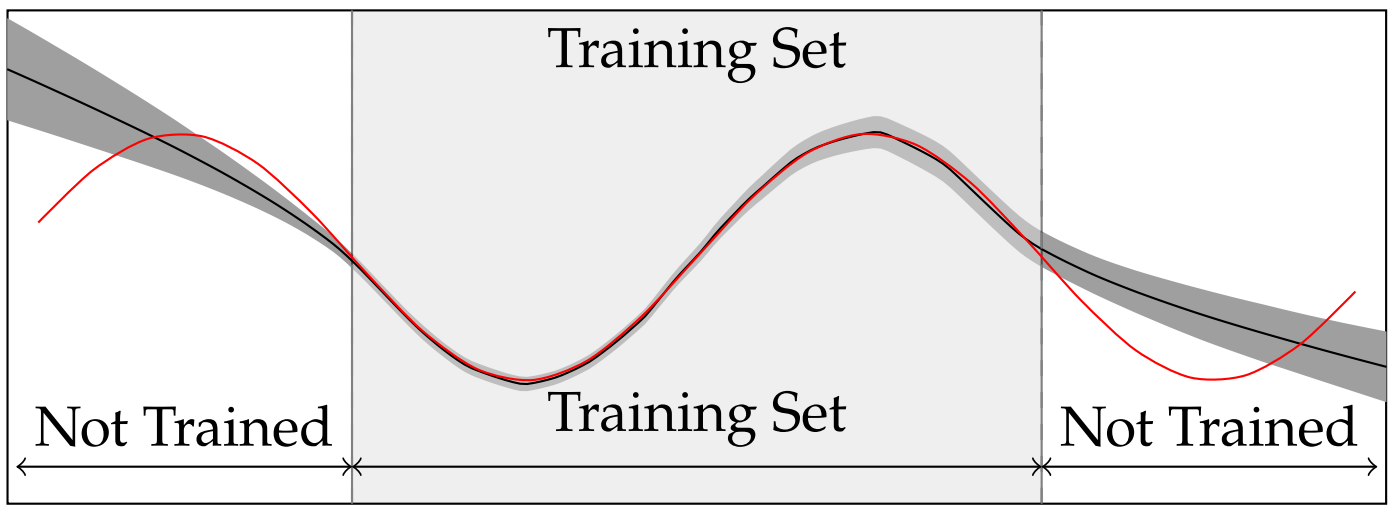
\begin{figure}[t]
	\centering
	\begin{tikzpicture}
		\begin{axis}[height = 0.20 \textheight, width = \textwidth, xlabel={}, ylabel={}, xmin = -6.28, xmax = 6.28, grid style=dashed, legend pos = north east, legend style={font=\scriptsize}, tick label style={font=\scriptsize}, ymin =-2.0 , ymax=2.0, trig format plots=rad, xtick=\empty, ytick=\empty]
			
			\draw[fill=gray!25,opacity=0.5]  (-3.1415, -2.0) -- (3.1415, -2.0) -- (3.1415, 2.0) -- (-3.1415, 2.0) -- cycle ;
			
			\draw[solid, black] (0.0, 1.25) node[above] {Training Set};
			
			\addplot[name path=upper, draw=none] table[x = x, y expr = \thisrow{pred_mu} + \thisrow{pred_sigma}, col sep = semicolon] {data/toy-regression-15ensembles-disentangled.csv};
			\addplot[name path=lower, draw=none] table[x = x, y expr = \thisrow{pred_mu} - \thisrow{pred_sigma}, col sep = semicolon] {data/toy-regression-15ensembles-disentangled.csv};
			
			\addplot[mark=none, black] table[x = x, y = pred_mu, col sep = semicolon] {data/toy-regression-15ensembles-disentangled.csv};
			
			\addplot [fill=darkgray!50, area legend] fill between[of=upper and lower]; 
			
			\addplot[mark=none, red, domain=-6:6, smooth] {sin(x)};
			
			\draw[dashed, gray] (3.1415, -2) -- (3.1415, 2);
			\draw[dashed, gray] (-3.1415, -2) -- (-3.1415, 2);
			
			\draw[to-to, solid, black] (-3.1415, -1.7) -- (3.1415, -1.7) node[midway, above] {Training Set};
			\draw[to-to, solid, black] (-6.2, -1.7) -- (-3.1415, -1.7) node[midway, above] {Not Trained};
			\draw[to-to, solid, black] (3.1415, -1.7) -- (6.2, -1.7) node[midway, above] {Not Trained};
		\end{axis}		
	\end{tikzpicture}
	\caption[Example of model with uncertainty estimation]{Example of model with uncertainty estimation, at each input point the model predicts a \textbf{mean} and {\color{darkgray} standard deviation} (in shaded area). The light shaded area represents the training set, while the other two areas represent data where the model has not been trained. The {\color{red} ground truth} is a sinusoid, and the model cannot predict correctly outside the training set, which is signalled by increasing uncertainty.}
	\label{fig:uncertainty_example}
\end{figure}
Supervised Machine Learning works very well when there is labelled data, but these labels are often made by humans, and depending on the task, they might be ambiguous. Humans make mistakes, and this extends to human-made labels. Models should be aware that labels could include some degree of uncertainty or be partially incorrect.

Considering that input data and labels can be noisy or ambiguous, Machine Learning models often do not consider uncertainties in both modalities during training. This book mostly focuses on modelling Uncertainty in the output of a model, but it is also possible to have inputs with uncertainty and propagate that uncertainty through the model.

Overall the concept of Uncertainty applied in Machine Learning is about the model telling the user about its limitations, which are different depending on the input and outputs. This is achieved by a model predicting a probability distribution that encodes uncertainty along with the prediction itself.

Figure \ref{fig:uncertainty_example} shows an example of a simple model with uncertainty trained on data generated from a sine function, but the training set is limited to $x \in [-\pi, \pi]$, and anything outside this range the model cannot predict correctly due to lack of knowledge given by its training set.

This example shows how uncertainty estimation can signal to the user that the model is making a mistake or incorrect prediction, due to the input being outside the training set. Uncertainty outside of the training set has to be compared with the uncertainty in the training set. It is clear from this example that uncertainty depends on the input and the sample we are providing to the model.

What are the difficulties of performing uncertainty estimation in machine learning? Table \ref{uncertainty_model_difficulty} makes a short survey of different model types and their complexity and common difficulties related to uncertainty estimation. The most interestings models for society are LLMs, which is also the most difficult setup and where the scientific literature lacks on how to properly estimate uncertainty.

\begin{table*}[t]
	\centering
	\begin{tabular}{lp{2.6cm}p{2.2cm}p{6.8cm}}
		\toprule
		\textbf{Model}			& \textbf{Model} & \textbf{Uncertainty} \\
		\textbf{Complexity} 	& \textbf{Examples} 					& \textbf{Techniques} & \textbf{Comments}\\
		\midrule
		Low					& Linear Models				& Bayesian Methods & Uncertainty estimation in linear models is relatively easy and can be done in exact ways using Bayesian methods.\\[1cm]
		Medium				& Computer\newline Vision\newline Models & Sampling Methods, Ensembles, Bayesian Neural\newline Networks & Uncertainty in Computer Vision and common ML models requires more advanced methods like Bayesian NNs and Ensembles, it is difficult due to non-linear models and number of parameters.\\
		High				& LLMs and\newline VLMs & Verbalized Uncertainty, Sampling Methods, Semantic Similarity & Uncertainty in complex models like LLMs and VLMs is very difficult due to extreme non-linearity and parameter in order of Billions. This is the most societally relevant setup.\\
		\bottomrule
	\end{tabular}
	\caption{Conceptual comparison of varying model complexities, uncertainty techniques, and their relationships to the field.}
	\label{uncertainty_model_difficulty}
\end{table*}

\section{Applications and Use Cases}

This section presents some use cases of uncertainty estimation in real-world applications.

\textbf{Applications Involving Humans}. When humans are involved or are affected by model predictions, then uncertainty estimation is a must. The most obvious example is Medical AI, where Physicians expect that any prediction will include some form of confidence to decide if the prediction is trustworthy \citep{marconi2025show} \citep{ojha2025navigating}. In real medical practice, there is always a large degree of uncertainty, for example blood tests and analysis provide a confidence interval on a physiological parameter of interest, or when making diagnoses sometimes it is uncertain which disease fits given symptoms.

\textbf{Noisy Sensor Data}. All sensing equipment is imperfect and produce noisy measurements \citep{belhedi2015noise}. In many fields the measurement noise is usually ignored as it is usually small, but for Computer Vision and Autonomous Driving, sensor noise can be a problem, particularly in crowded and complex environments . This would involve using measurement noise as an input to a model with uncertainty estimation, where it is expected that increasing input noise would lead to increased output uncertainty. A clear example are LiDAR and time-of-flight sensors, as usually the measurement error is proportional to distance, so sensing objects far away is more noisy than closer objects.

\textbf{Detecting LLM Hallucinations}. A major application of uncertainty estimation is to detect when a model produces an incorrect prediction \citep{galil2023selectivpred}, which in the context of LLMs is colloquially called a "hallucination" \citep{kang2026uncertainty}. Most LLMs and VLMs are not trained with proper uncertainty estimation methods, so often they simply do not provide any confidence values for users to interpret. A big step would be to develop LLMs with confidence estimation so users can be informed that the model can make mistakes and should not always be trusted, and this trust should be a function of the prediction confidence.

\textbf{Explainable AI}. There are many methods to obtain explanations for a given prediction made by a model, but these methods are also models and have their own uncertainty, so when explanations are incorrect, uncertainty in the explanation method can signal about this \citep{slack2021reliable}.

Additionally a confidence score can be used as a simple explanation, a model making an incorrect prediction is because it is uncertain and lacks the required knowledge.

\section{Required Knowledge}

This book assumes that the reader is familiar with basic machine learning concepts like training neural networks, loss functions, etc.

Since uncertainty is modelled using probability, basic knowledge of probability and statistics is assumed, in particular probability distributions and basic rules of probability.

\section{Structure of This Book}

This book is divided into two parts. The first part delivers the basics of uncertainty estimation, how to train models with uncertainty, how to evaluate them, calibration, and out of distribution detection. This is covered in Chapters \ref{ch:concepts} to \ref{ch:ood_detect}.

The second part covers advanced topics that require basic understanding from the first part.

\section{Similar Books and Literature}

This is of course not the first book to deal with the broad topic of Uncertainty in Artificial Intelligence and Probabilistic Models.
The book "Probabilistic Artificial Intelligence" by \sidecite{krause2025probabilistic} focuses on methods to perform probabilistic inference in artificial intelligence systems, while the book "Probabilistic Machine Learning" by \sidecite{murphy2022probabilistic} and its more advanced version (\sidecite{murphy2023probabilistic}) both deal with probabilistic modelling of many types of machine learning models without focusing on uncertainty and its possible sources.

This book and set of lecture notes focuses on the topic of uncertainty estimation in machine learning models, which is more broad and general than probabilistic modelling, and we focus on what uncertainty is, how it can be estimated, how it can be evaluated, and how it can be used to detect misclassified and out of distribution inputs.

We do claim that this is the first textbook on the topic of uncertainty estimation methods for machine learning models, which aims to build easy to understand material for students in an undergraduate or graduate course. We do recognize that there is many overlap with previously published books, specially in the methods section.

There are several survey papers touching several aspects of this topic: \sidecite{gawlikowski2023survey} surveys uncertainty in deep neural networks,  while \sidecite{he2026survey} does similarly for deep learning, and \sidecite{shi2025survey} surveys the intersection of uncertainty in machine learning and engineering applications.

\sidecite{zhou2022survey} reviews of methods specifically to estimate epistemic uncertainty.

%% file: sections/basic-concepts.tex
\chapter{Basic Concepts in Uncertainty}
\label{ch:concepts}
\margintoc

This chapter covers the basic concepts that are associated to Uncertainty in Machine Learning, including sources of uncertainty, how uncertainty is represented, and an introduction to Bayesian thinking and Maximum likelihood estimation methods. All these concepts will be very useful for later understanding of uncertainty quantification methods and related concepts.

\section{Machine Learning Notation}
This section defines some basic machine learning notation and setup.

Supervised learning datasets are denoted as $\{(x_i, y_i)\}_{N}^{i=1}$, corresponding to $N$ tuples of $(x,y)$ values with $x$ being inputs (images, text, features, etc) and $y$ being ground truth labels (one-hot encoded, continuous, etc).

A model is a function $f: \mathcal{X} \rightarrow \mathcal{Y}$ that transforms features $x \in \mathcal{X}$ into predictions $y \in \mathcal{Y}$. If $y$ is discrete then it is a classification task, while if $y$ is continuous then it is a regression task.

Predictions made by models are usually denoted as $\hat{y}$, the hat $\hat{ }$ denoting that is a prediction, while values without the hat are usually ground truth or input values. Usually $\hat{y} = \model{x}$, indicating that the prediction $\hat{y}$ is made by a model given input $x$.

In machine learning we mainly use two kinds of probabilistic models:

\begin{description}
	\item[Generative Models] These kind of models aim to describe the probability distribution of data points $x \sim X$, with $X$ being a random variable. This is basically density estimation.
	\begin{equation}
		\hat{y} = \prob{X}
	\end{equation}
	Examples for this kind of model are Kernel Density Estimation (KDE), Generative Adversarial Networks (GANs), Diffusion Models, and the whole field of Density Estimation.
	
	\item[Predictive or Discriminative Models] This kind of model aims to predict the distribution of an output $Y$ variable given an input variable $X$.
	\begin{equation}
		\hat{y} = \condprob{Y}{X}
	\end{equation}
	Examples of predictive models are neural networks, support vector machines, linear regression, etc. In particular discriminative models are usually classification models.	
\end{description}

In most modelling performed in machine learning, we are more concerned about predictive models, but generative models will also come useful later. The algorithms that are used for both kind of models are very different.

\section{Statistical Notation}

Random variables are denoted with uppercase letters (e.g. $X, Y, Z$), while samples or their realizations are denoted with lowercase numbers (e.g. $x, y, z$). A random variable is defined by a probability distribution function.

The probability distribution function \marginnote{Probability Distribution Function} (PDF) of random variable $X$ is denoted as $f_X$, and it is the function that defines a distribution through its density.

The cumulative distribution function \marginnote{Cumulative Distribution Function} (CDF) of a random variable $X$ is denoted as $F_X$, and is the function $F_X(x) = \prob{X \leq x}$, defining the probability of a random variable being less than a defined value, indicating how much probability mass has accumulated until the defined value.

Random variables are defined by a probability distribution, and this is denoted as $X \sim F$, where random variable $X$ follows probability distribution $F$.

\section{Uncertainty Concepts}
This section describes the some basic definitions of uncertainty.
\begin{definitionbox}{Point-wise vs Distribution Outputs}{pw_vs_distribution_outputs}

The main difference between classic machine learning models, and models with uncertainty, is that classical models make point-wise predictions, while models with uncertainty make distribution outputs.
\begin{description}
	\item[Point-wise Outputs] A prediction without confidence or uncertainty, for example, a class prediction (dog, cat, fox, etc) in classification, a scalar number for regression, etc.
	\item[Distribution Outputs] A prediction that represents a full probability distribution, usually in the form of parameters of a given probability distribution. Uncertainty is implicitly encoded in the probability distribution.
\end{description}
\end{definitionbox}

\begin{definitionbox}{Probability}{probability}
Probabilities are numbers in the $[0, 1]$ range that indicates the likelihood of an event happening. This is the classical statistical definition, where a probability is the ratio of number of ways an event happens to the total number of events.

Probabilities are used to model uncertain phenomena, where there is imperfect or lack of information. For machine learning models, the output can be probabilities, for example in standard classification settings using softmax activation and cross-entropy loss.
\end{definitionbox}

The specific interpretation of probabilities that are output by a model is the likelihood of a prediction being correct, for example for classification with $c \in C$ classes.
\begin{equation}
    \hat{y} = \model{x} = \condprob{y = c}{x}
\end{equation}
The model outputs a probability that the prediction of a given value is correct, so a higher probability means the model is more confident or sure about its answer, and lower probabilities indicate that the model is guessing or simply does not know.
\begin{equation}
	\condprob{y}{x} = \prob{\hat{y} \text{ is correct}}
\end{equation}

\begin{margintable}
    \begin{tabular}{lll}
        \toprule
        \textbf{Concept}	& \textbf{Range} 	& \textbf{Direction}\\
        \midrule
        Probabilities	& $[0, 1]$			& $\uparrow$\\
        Confidences		& $[0, \infty]$		& $\uparrow$\\
        Uncertainty		& $[0, \infty]$ 	& $\downarrow$\\
        \bottomrule
    \end{tabular}
    \caption{Short comparison of uncertainty measures and their direction for less uncertain and high likelihood of a prediction being correct.}
\end{margintable}
\begin{definitionbox}{Confidence}{confidence}
Confidences expand the notion of probabilities, where confidences are numbers but not in the $[0, 1]$ range, they can be in any range. The specific semantics or interpretation of a confidence depend on how this value is computed and/or defined, but generally a increasing value indicates that the model's prediction is more likely to be correct.

Probabilities are also confidences, but confidences are generally not probabilities.
\end{definitionbox}

\begin{figure*}
    \subfloat[$\sigma^2 = 0.5$]{
        \begin{tikzpicture}
            \begin{axis}[no markers, samples=100, domain=-5:5, height=0.2\textheight, xtick=\empty, ytick=\empty, axis lines*=left, xlabel=$x$, ylabel=Density]
                \addplot [cyan!50!black] {gauss(0.0, 0.5)};
            \end{axis}
        \end{tikzpicture}
    }
    \subfloat[$\sigma^2 = 1.0$]{
        \begin{tikzpicture}
            \begin{axis}[no markers, samples=100, domain=-5:5, height=0.2\textheight, xtick=\empty, ytick=\empty, axis lines*=left, xlabel=$x$, ylabel=Density]
                \addplot [cyan!50!black] {gauss(0.0, 1.0)};
            \end{axis}
        \end{tikzpicture}
    }
    \subfloat[$\sigma^2 = 2.0$]{
        \begin{tikzpicture}
            \begin{axis}[no markers, samples=100, domain=-5:5, height=0.2\textheight, xtick=\empty, ytick=\empty, axis lines*=left, xlabel=$x$, ylabel=Density]
                \addplot [cyan!50!black] {gauss(0.0, 2.0)};
            \end{axis}
        \end{tikzpicture}
    }
    \caption{Multiple Gaussian distributions with different variances, indicating multiple levels of uncertainty/confidence.}
\end{figure*}
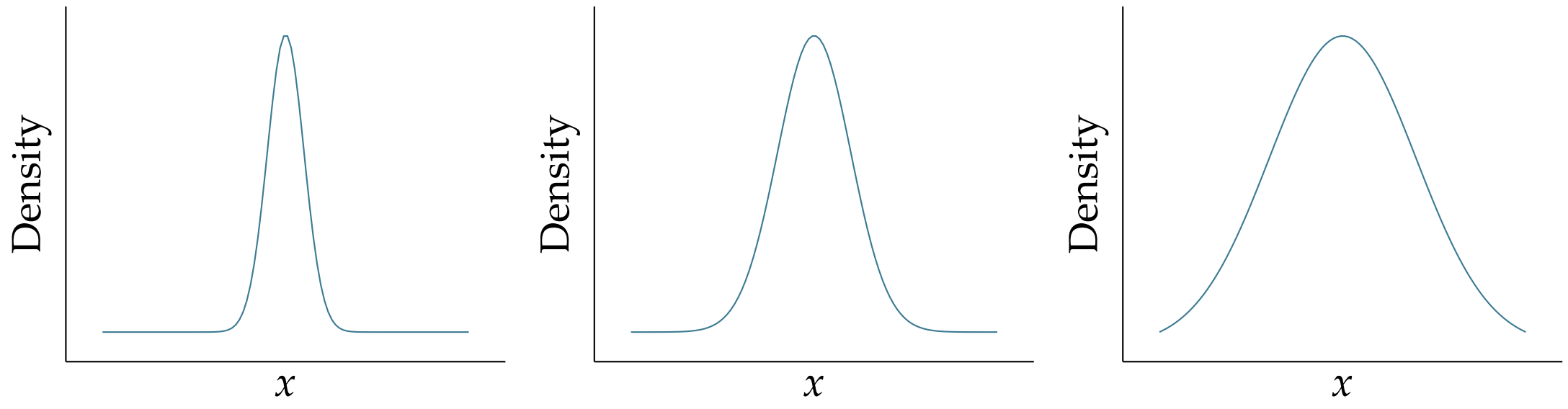

\begin{definitionbox}{Uncertainty}{uncertainty}
Uncertainty is the most general term to indicate lack of knowledge or imperfect information, or simply possibility of an error.
	
Ideally, an uncertainty metric should be low when there is no uncertainty (the model is completely certain of its output), and increasing with more uncertainty in the output, up to a maximum value which indicates that the model is completely guessing.
\end{definitionbox}

One uncertainty metric is entropy \marginnote{Entropy}, which indicates the amount of information in a probability distribution $f$ \citep{mackay2003information}, and is defined for classification as:
\begin{equation}
    H(f) = -\sum_{x \in X} f(x) \log f(x)
\end{equation}
And for regression we replace the sum by an integral:
\begin{equation}
    H(f) = -\int f(x) \log f(x) dx
\end{equation}
Entropy is measured in bits (base 2 logarithm) or nats (base 10 logarithm) and is a information theoretic measure of uncertainty in a probability distribution.

Another measure that is easier to compute is the maximum probability \marginnote{Maximum Probability} \citep{hendrycks2017baseline}:
\begin{equation}
    P = 1 - \max_{c \in C} \condprob{y = c}{x }
\end{equation}
A direct uncertainty measure is the standard deviation of a probability distribution, which basically measures the spread of the distribution over its input variable.

\section{Sources of Uncertainty}

There are two main sources of where uncertainty comes from \citep{hullermeier2021aleatoric} \citep{der2009aleatory} \citep{valdenegro2022deeperlook}. This section describes them.

\begin{definitionbox}{Data Uncertainty}{ale_uncertainty}
This is uncertainty inherent to the data, with the fundamental property that it cannot be reduced by including more information or data into the training process. Also known as aleatoric uncertainty. For example:

\begin{description}
    \item[Sensor Noise] Data is usually captured using sensors like cameras, LIDAR, SONAR, etc. All sensing devices have varying amounts of noise added to their measurements, which also translates into noise into the data itself.
    \item[Stochastic Processes] Some processes that are used to generate or measure data have an inherent random component, which is also called stochasticity.
    \item[Annotation Disagreement] Usually data is labeled for supervised learning by human annotators, which decide the different kinds of labels for data points, for example bounding box annotations for object detection, or class labels for classification. Different humans might label the data slightly differently, and there could be disagreement about a particular image among annotators, and this is a form of data uncertainty. Also consider that some labels might be incorrect, which is also a form of label noise.
\end{description}

Note that data uncertainty can be present both in the labels $y$ and in the inputs $x$, as data is meant generally as an input to the training process, not only to the model.
\end{definitionbox}

There are two types of data uncertainty, heteroscedastic and homoscedastic.

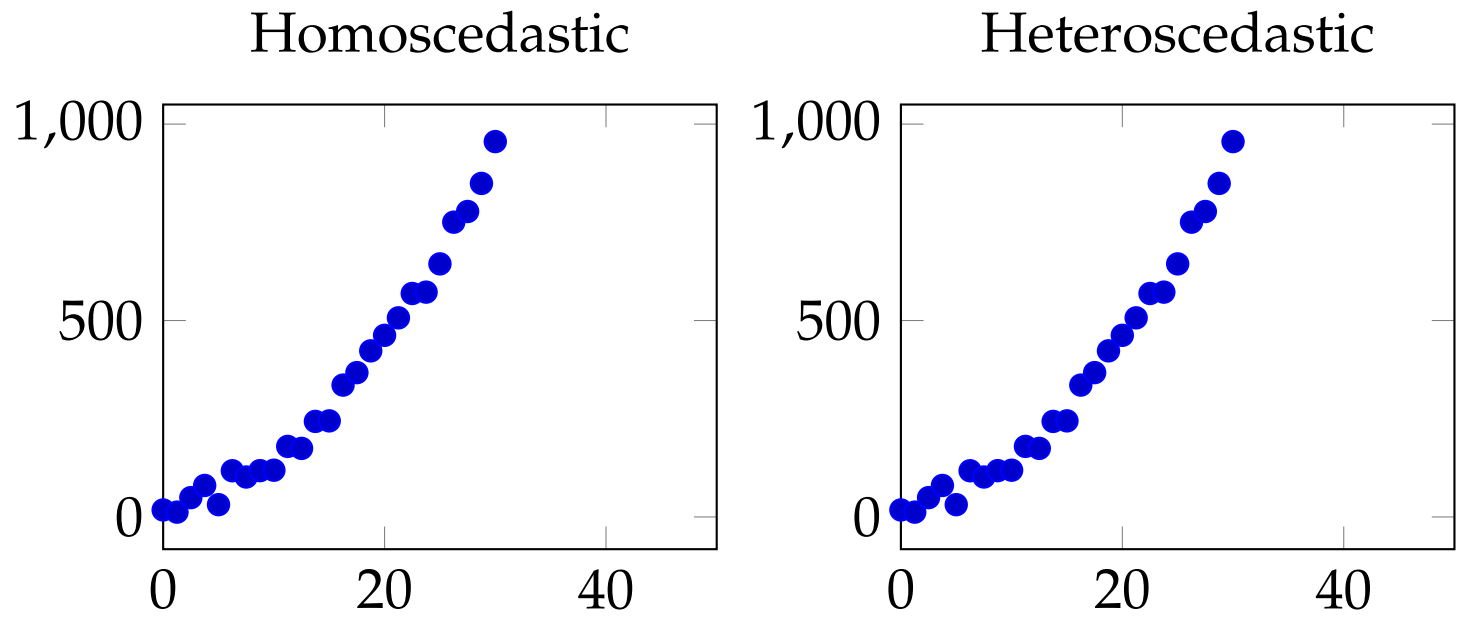
\begin{figure}[h]
	\begin{tikzpicture}
		\begin{axis}[width=0.49\textwidth, xmin=0, xmax=50, title=Homoscedastic]
			\addplot gnuplot[only marks, domain=0:50]{x + 10 * rand(0)};
		\end{axis}
	\end{tikzpicture}        
	\begin{tikzpicture}
		\begin{axis}[width=0.49\textwidth, xmin=0, xmax=50, title=Heteroscedastic]
			\addplot gnuplot[only marks, domain=0:50]{x + 2 * x * rand(0)};
		\end{axis}
	\end{tikzpicture}
	\caption[Examples of different data uncertainty noise]{Examples of different data uncertainty noise. The function $g(x) = x + \epsilon$ with $\epsilon \sim \mathcal{N}(0, \sigma^2)$ is plotted with two noise sources, the left plot shows constant variance (homoscedatic, $\sigma^2$), while the right plot shows variable variance (heteroscedatic, $\sigma^2(x)$).}
	\label{bc:data_uncert_types}
\end{figure}

\begin{table}[t]
	\begin{tabular}{p{2.5cm}p{7.5cm}l}
		\toprule
		\textbf{Property} & \textbf{Description}\\		
		\midrule
		Incorrect\newline Labels 	& Datasets can have labels that are not correct. There can be annotator disagreement or simply humans make mistakes when labeling a data point. This often does not happen for the full dataset but for a small number of data points.\\
		Data 				& Different datasets have different amounts of data uncertainty, due to different labeling methods or data sources.\\
		Measurement Noise 	& Labels often correspond to measurements (for example, depth, color values, etc) and all measurements are noisy due to stochastic processes or simply imperfections in the measuring device. Then these stochastic measurements become labels that produce data uncertainty. \\
		\bottomrule
	\end{tabular}
	\caption{Summary of properties of data/aleatoric uncertainty.}
\end{table}

\begin{definitionbox}{Types of Data Uncertainty}{ale_uncertainty_types}
Consider the following noisy function:
\begin{equation}
    f(x) = g(x) + \epsilon
\end{equation}
Where $\epsilon \sim \mathcal{N}(0, \sigma^2)$ is an additive noise with zero mean and a $\sigma^2 > 0$ variance, and $g(x)$ is what we call the true function, and $f(x)$ is the measured function, usually the data points that compose a training set. The true function is generally unknown.

\begin{description}
	\item[Homoscedastic] $\sigma^2$ is a constant value and not a function of the input $x$.
	\item[Heteroscedastic] $\sigma^2$ is not constant but instead it is a function of the input $x$.
\end{description}

These two types of data uncertainty define how data uncertainty is modeled and what assumptions can be made.
\end{definitionbox}
Figure \ref{bc:data_uncert_types} shows a visual example with a function $g(x) = x$ and the two types of data uncertainty.

\begin{figure*}
	\subfloat[Example of model misspecification. Each plot represents fitting a different model, while the blue points are the training set. We do not know the true function that generated the training set, and through different models, we try to fit the data.]{
		\input{figures/model_misspecification.tikz }
        \label{bc:model_misspec}
	}
	\hspace*{0.5em}
	\subfloat[Example of class imbalance on point cloud segmentation in the KITTI dataset. It is clear how less training samples leads to higher uncertainty represented as entropy. Result from \cite{bhandary2020evaluating}]{
		\input{figures/kitti_semseg_entropy.tikz}
	}
	\caption{Two examples of epistemic uncertainty sources.}
\end{figure*}
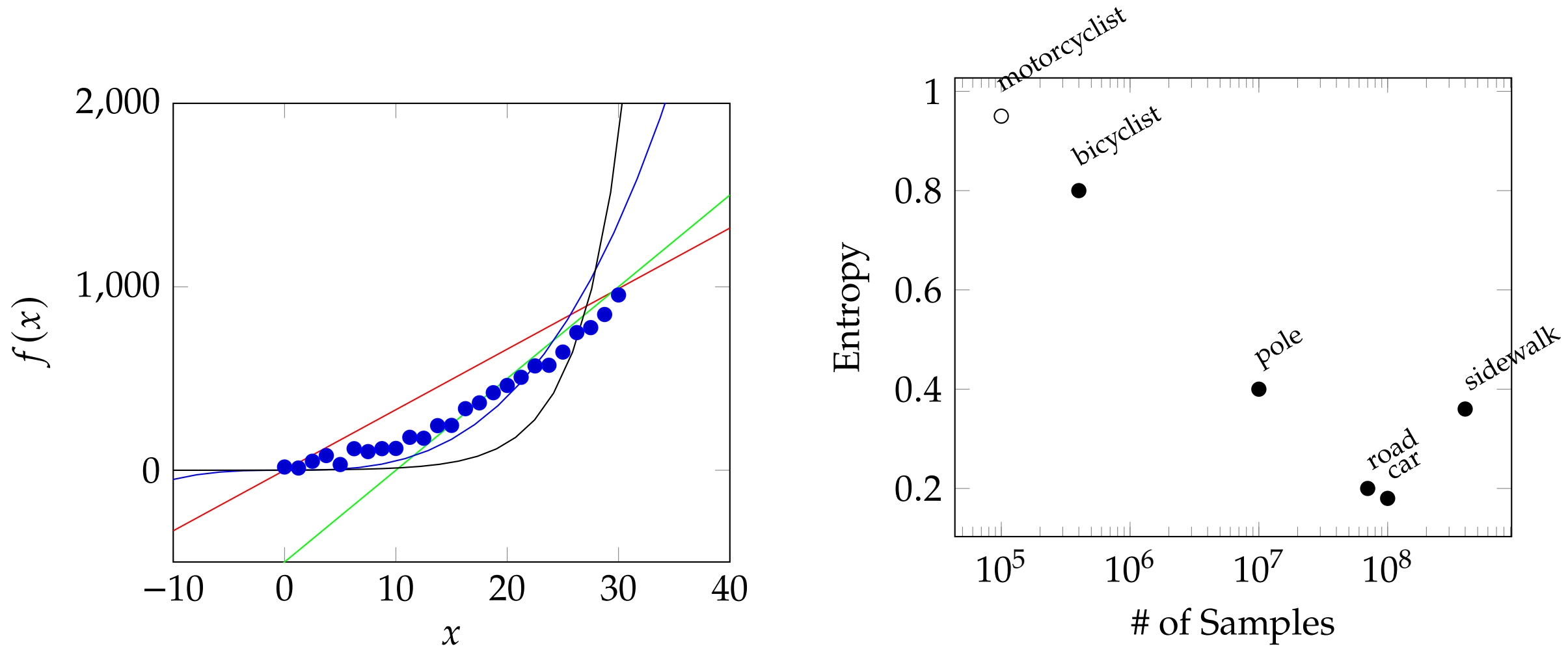

\begin{definitionbox}{Model Uncertainty}{epi_uncertainty}
This is uncertainty inherent to the model itself, for example lack of knowledge in the model, the training process itself, and assumptions made by the model through its structure. The fundamental property is that it can be reduced by including more information. Also known as epistemic uncertainty. Some examples:

\begin{description}
    \item[Training Data] The training set size directly impacts the amount of knowledge that the model can learn, and model uncertainty is inversely proportional to the training set size. The same model with less data will be less confident about its prediction than when learned with more data.
    \item[Class Imbalance] Generally the number of samples of each class is not the same, and in some cases the training set can be biased towards one class ($> 50\%$), and this bias also translates into different confidences for each class. Classes with more training samples will predict lower model uncertainty, and classes with less training samples will be more uncertain by the model.
    \item[Model Misspecification] Generally we do now know the true process or function that generated a sample of data, so when applying models, we make assumptions, which are usually incorrect but often good enough. The difference between the true process/function and the model equations is called model misspecification. This concept is shown in Figure \ref{bc:model_misspec}.
\end{description}
\end{definitionbox}

Model uncertainty is usually produced due to a lack of knowledge in the model parameters

Model uncertainty is inversely related to density of training data points. In areas where there are many training data points, model uncertainty will be low, but in areas where training points are sparse, then model uncertainty should increase.

Note that model uncertainty is generally only present in the predictions of a model $\hat{y}$, since input data has to pass through the model in order to "obtain" model uncertainty. Inputs by themselves do not have model uncertainty, which is a major difference with data uncertainty.

\begin{table}[t]
	\begin{tabular}{p{2.5cm}p{7.5cm}}
		\toprule
		\textbf{Property} & \textbf{Description}\\		
		\midrule
		Training\newline Samples 	& Epistemic uncertainty increases if a model is trained with less training data, smaller training sets, and decreases when training data is increased, for example by using data augmentation.\\
		Model 			 	& Different models and architectures have different epistemic uncertainty.\\
		Class\newline Imbalance		& Datasets often have class imbalance, the number of samples for each class being different. Classes with lower number of training samples will have higher model uncertainty, and classes with higher number of training samples will have lower model uncertainty.\\
		Training Gaps 		& A training set can have gaps in covering all possible classes and their input variability. Inputs corresponding to gaps in the training set will have increased model uncertainty.\\
		\bottomrule
	\end{tabular}
	\caption{Summary of properties of model/epistemic uncertainty.}
\end{table}

\begin{definitionbox}{Predictive Uncertainty}{pred_uncertainty}
	Predictive uncertainty is the combination of data and model uncertainty, just a simple addition, which is usually what is predicted by a model. In Sec TODO we cover ways to divide or disentangle predictive uncertainty into model and data uncertainty.
	
	\parbox{2.0cm}{Predictive\\ Uncertainty} = $\overUnderArrow[]{\parbox{1.0cm}{Aleatoric}}{\text{Data Uncertainty}} + \overUnderArrow[]{\parbox{1.0cm}{Epistemic}}{\text{Model Uncertainty}}$
\end{definitionbox}

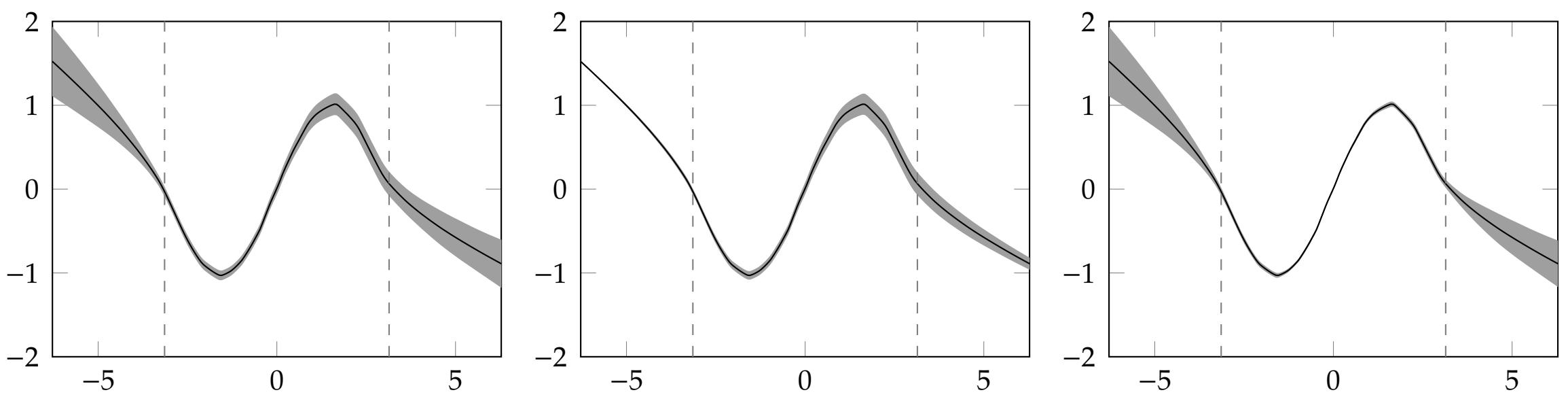
\begin{figure*}
    \centering    
    \subfloat[Predictive Uncertainty]{
    	\begin{tikzpicture}
    		\begin{axis}[height = 0.20 \textheight, width = 0.37\linewidth, xlabel={}, ylabel={}, xmin = -6.28, xmax = 6.28, grid style=dashed, legend pos = north east, legend style={font=\scriptsize}, tick label style={font=\scriptsize}, ymin =-2.0 , ymax=2.0]
    			
    			\addplot[name path=upper, draw=none] table[x = x, y expr = \thisrow{pred_mu} + \thisrow{pred_sigma}, col sep = semicolon] {data/toy-regression-15ensembles-disentangled.csv};
    			\addplot[name path=lower, draw=none] table[x = x, y expr = \thisrow{pred_mu} - \thisrow{pred_sigma}, col sep = semicolon] {data/toy-regression-15ensembles-disentangled.csv};
    			
    			\addplot[mark=none, black] table[x = x, y = pred_mu, col sep = semicolon] {data/toy-regression-15ensembles-disentangled.csv};
    			
    			\addplot [fill=darkgray!50, area legend] fill between[of=upper and lower]; 
    			
    			\draw[dashed, gray] (3.1415, -2) -- (3.1415, 2);
    			\draw[dashed, gray] (-3.1415, -2) -- (-3.1415, 2);

    		\end{axis}		
    	\end{tikzpicture}
    }
    \subfloat[Data Uncertainty]{
        \begin{tikzpicture}
            \begin{axis}[height = 0.20 \textheight, width = 0.37\linewidth, xlabel={}, ylabel={}, xmin = -6.28, xmax = 6.28, grid style=dashed, legend pos = north east, legend style={font=\scriptsize}, tick label style={font=\scriptsize}, ymin =-2.0 , ymax=2.0]
                
                \addplot[name path=upper, draw=none, forget plot] table[x = x, y expr = \thisrow{pred_mu} + \thisrow{pred_sigma_ale}, col sep = semicolon] {data/toy-regression-15ensembles-disentangled.csv};
                \addplot[name path=lower, draw=none, forget plot] table[x = x, y expr = \thisrow{pred_mu} - \thisrow{pred_sigma_ale}, col sep = semicolon] {data/toy-regression-15ensembles-disentangled.csv};
                
                \addplot[mark=none, black] table[x = x, y = pred_mu, col sep = semicolon] {data/toy-regression-15ensembles-disentangled.csv};
                
                \addplot [fill=darkgray!50, area legend] fill between[of=upper and lower]; 
                
                \draw[dashed, gray] (3.1415, -2) -- (3.1415, 2);
                \draw[dashed, gray] (-3.1415, -2) -- (-3.1415, 2);       
            \end{axis}		
        \end{tikzpicture}
    }    
    \subfloat[Model Uncertainty]{
        \begin{tikzpicture}
            \begin{axis}[height = 0.20 \textheight, width = 0.37\linewidth, xlabel={}, ylabel={}, xmin = -6.28, xmax = 6.28, grid style=dashed, legend pos = north east, legend style={font=\scriptsize}, tick label style={font=\scriptsize}, ymin =-2.0 , ymax=2.0]
                
                \addplot[name path=upper, draw=none] table[x = x, y expr = \thisrow{pred_mu} + \thisrow{pred_sigma_epi}, col sep = semicolon] {data/toy-regression-15ensembles-disentangled.csv};
                \addplot[name path=lower, draw=none] table[x = x, y expr = \thisrow{pred_mu} - \thisrow{pred_sigma_epi}, col sep = semicolon] {data/toy-regression-15ensembles-disentangled.csv};
                
                \addplot[mark=none, black] table[x = x, y = pred_mu, col sep = semicolon] {data/toy-regression-15ensembles-disentangled.csv};
                
                \addplot[fill=darkgray!50] fill between[of=upper and lower]; 
                
                \draw[dashed, gray] (3.1415, -2) -- (3.1415, 2);
                \draw[dashed, gray] (-3.1415, -2) -- (-3.1415, 2);
            \end{axis}		
        \end{tikzpicture}
    }
    \caption{Sample of uncertainty disentanglement in a toy regression example, produced using an ensemble of 15 neural networks. }
    \label{fig:ensemble_example}
\end{figure*}

What does it mean add information? Aleatoric uncertainty will not reduce with more information, while epistemic uncertainty does decrease in the same case.

Adding information consists of increasing the size of the training set, using a better model (that better fits the data), using regularization to constraint the model, or overall improving the predictive power of the model.

An example for aleatoric uncertainty is using a ruler to measure a table. You use the rule and get a measurement, you repeat the process and get another value. These measurements will not be the same since you are looking at the ruler to get a measurement, and the ruler markings do not always align well with the object being measured. If you repeat the process many times, each measurement will still be noisy, since the noise or measurement error comes from using the ruler to measure.

Repeating the process can be seen as adding more information, but noise in the measuring process is still there. It has not been reduced. People often argue that by taking the mean or any other aggregation metric, the error can be reduced, and this could be true, but by doing this, assumptions are being made on how the error behaves, and there are no guarantees on how close the mean will be to the true table size. And in the end, this does not change that the ruler measurements are noisy.

An example for epistemic uncertainty is training a model with a small dataset, let's say of cats and dogs. Imagine a training set with a single cat sample, and 100 dog samples. From this data, we train a model, and notice that it works okay with dogs, but fails with many cats. Looking at the softmax confidences shows that the model predicts very high confidences for dog, and very low confidences for cat, which are a reflection of the class imbalance in the training set. If we obtain additional cat images and train the model from scratch (with a more balanced distribution of cat vs dog samples), then then model would predict more cat images correctly, and the softmax confidences would be higher for cat images.

Similarly would happen if we use a balanced training distribution (50\% cat, 50\% dog) but with a low number of samples, say $N = 10$ samples in total, the model would make many mistakes classifying cats and dogs, and these would be reflected in low confidences, closer to the boundary $\condprob{y=\text{cat}/\text{dog}}{x} = 0.5$, and by increasing the training set size with more varied cat and dog images, confidences would increase as the model's knowledge increases.

\begin{figure}
	\begin{tikzpicture}
		\begin{axis}[width=\linewidth, ticks=none]        
			\addplot[scatter, only marks, scatter src=explicit symbolic, point meta=explicit] table [x=x, y=y, meta=label, col sep=semicolon] {data/gaussian2d_data.csv};								
			
			\draw[gray] (-2, 2) -- (1.5, -2);
			\draw[gray] (-2.2, 1.5) -- (1.5, -1.5);
			\draw[gray] (-2.0, 1.6) -- (1.8, -1.3);
			
			\coordinate (a1) at (-3.3,-1);
			\coordinate (a2) at (3.0, -0.1);
			\coordinate (a3) at (4.5,0.5);
			\coordinate (a4) at (2.7,1.6);
			\coordinate (a5) at (3,3);
			\coordinate (a6) at (0.0, 3.0);
			\coordinate (a7) at (-2.0, 1.2);
			\coordinate (O) at ($(a1)-(1,2)$);

			\draw (a1) to[curve through={(-2.5, -2.5) .. (0, -2.5) .. (a2) .. (a3) .. (a4) .. (a5) .. (1.3, 2.8) .. (1.6, 3.2) .. (a6) .. (-1.0, 3) .. (a7) .. (-3.0, 0.0)}] (a1);
			
			\filldraw[fill=gray, pattern=north east lines, pattern color=gray!70, even odd rule] (-5,-4) rectangle (5, 4) (a1) to[curve through={(-2.5, -2.5) .. (0, -2.5) .. (a2) .. (a3) .. (a4) .. (a5) .. (1.3, 2.8) .. (1.6, 3.2) .. (a6) .. (-1.0, 3) .. (a7) .. (-3.0, 0.0)}] (a1);
			
			\node[align=left, text width=2.0cm] at (3.2, -3.0) {{Model\\ Uncertainty}};
			
			\node[align=left, text width=2.0cm] at (-2.6, 3.0) {{Data\\ Uncertainty}};
			\draw[-{Stealth[black]}] (-2.6, 2.5)   -- (-2.0, 1.5);

		\end{axis}
	\end{tikzpicture}
	\caption[Example relationship between sources of uncertainty in classification]{Example relationship between sources of uncertainty in classification. These features are drawn from overlapping Gaussian distributions, producing data uncertainty within the decision boundary, while any region without data points has high model uncertainty, represented by the shaded area. Linear models along the decision boundary are represented as lines and show that fitting the data perfectly is not possible due to high data uncertainty.}
\end{figure}
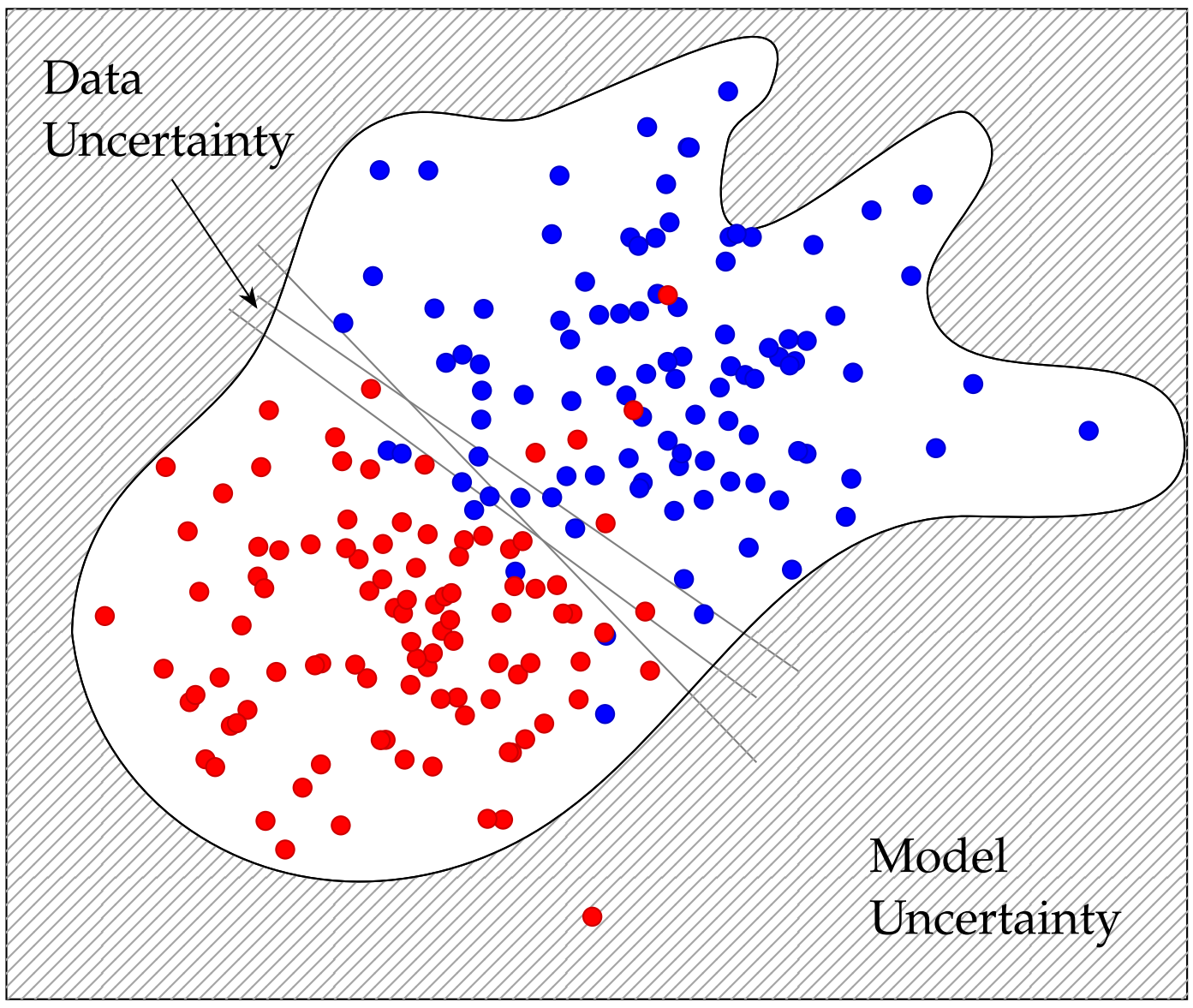

\section{Representations of Uncertainty}

The most general representation of output uncertainty in a model is a full probability distribution, usually by outputting the parameters of such distribution. This section describes several ways how predictions with uncertainty can be represented, divided by classification or regression tasks.

\subsection{Classification}

\begin{marginfigure}
    \begin{tikzpicture}
        \begin{axis}[ybar interval, ymax=55,ymin=0, minor y tick num = 3, width=\linewidth]
            \addplot coordinates { (0, 5) (5, 35) (10, 50) (15, 30) (20, 15) (25, 0) };
        \end{axis}
    \end{tikzpicture}
    \caption{Histogram representing multi-class classification probabilities}
\end{marginfigure}

The most common formulation for multi classification is to use one probability for each class, this is denoted as:
\begin{equation}
    \hat{y} = \condprob{y = c}{x}
\end{equation}
This formulation can be read as "probability of class $c$ given the input $x$". There are two ways to interpret this formulation:

\begin{description}
    \item[Multi-Class] There is one possible correct class, and the probabilities form a distribution, meaning the values should sum to one $\sum_{c \in C} \condprob{y = c}{x} = 1.0$.
    \item[Multi-Label] More than one class can be correct, and $\condprob{y = c}{x}$ represents independent probabilities for class $c$. In this case the probabilities do not have to sum to one.
\end{description}

Specifically for binary classification, a Binomial distribution is usually used, which corresponds to outputting the probability of the positive class $\condprob{y = 1}{x}$.

\subsection{Regression}

In a regression setting, where the output variable $y$ is continuous, there are multiple choices of representation. The most general one is an interval:
\begin{eqnarray}
    y \in [y_{min}, y_{max}]
\end{eqnarray}
A more specific interval representation is the confidence interval \marginnote{Confidence Interval}:
\begin{equation}
    y \in [l_\alpha, u_\alpha]
\end{equation}
The interpretation of a confidence interval with confidence level $\alpha$ is that it should contain the true value $y$, on average, $\alpha$ of the time, or in probabilistic terms:
\begin{equation}
    \prob{l_\alpha \leq y \leq u_\alpha} = \alpha
\end{equation}
Where $l_\alpha$ and $u_\alpha$ are the confidence interval bounds, which are functions of the confidence level $\alpha$. The confidence interval width $|u_\alpha - l_\alpha|$ defines its level of predictive uncertainty. The intuition is that a large confidence level leads to a larger confidence interval, while a small confidence level leads to a smaller confidence interval. Generally a small confidence interval is preferred as it bounds the true value more tightly, but then it is less likely to contain the true value. If the confidence increases to $\alpha = 100\%$, then the confidence interval is basically $[-\infty, \infty]$, which is not informative.

Finally, many distributions can be output, but the most common one is the Gaussian distribution. This is often done by a model predicting the parameters of the probability distribution:
\begin{equation}
    y \sim \Gaussian{\mu}{\sigma^2}
\end{equation}
In this case, the model outputs both the mean $\mu$ and variance $\sigma^2$ \marginnote{Gaussian Mean-Variance}, where the prediction itself is $\mu$, and $\sigma$ (standard deviation) can be interpreted as its associated uncertainty.

\begin{marginfigure}
    \begin{tikzpicture}
        
        \begin{axis}[width=\linewidth, xtick = {-1.96, 0.0, 1.64}, xticklabels={$\tau=0.05$, $\tau=0.5$, $\tau=0.95$}, domain=-2:2, tick label style={rotate=30}]
            \addplot+[mark=none, smooth] {gauss(0.0, 1.0)};            
        \end{axis}
    \end{tikzpicture}
    \caption{Some quantiles for the standard normal distribution $\text{Gaussian}(0, 1)$}
\end{marginfigure}
Another representation that is more implicit is using quantiles, remember that the quantile function $q$ is inverse of the CDF $F$:
\begin{equation}
	q(\tau) = F^{-1}(\tau) = \inf \{y: F(y) \geq \tau \}
\end{equation}
Where $\tau \in [0, 1]$ is the $\tau$ quantile of the predictive distribution. A quantile \marginnote{Quantile Outputs} is a division of the accumulated probability space, so $\tau = 0.5$ (the median) means the point where half of the accumulate probability lies. $\tau = 0.95$ is the point where 95\% of the accumulated probability lies, and so on.

Using quantiles allows for an implicit representation of a probability distribution, and the model can decide which probability distribution to output as the quantile representation is flexible.

A few quantiles are usually regressed, like the median ($\tau = 0.5$), and extreme values like $\tau = 0.05$/$\tau = 0.95$. This is usually learned as a multi-task setup using the quantile loss, also known as the pinball loss. Multiple quantiles can be regressed at the same time, to partially reconstruct the predictive distribution.

\section{Bayesian Thinking}
What is called Bayesian Statistics or Bayesian Thinking, is the application of Bayes rule to update beliefs, that is, to start from previously known information, update it with an observation, to obtain new information, which can then be repeatedly updated with new observations.

There is a way to do this in a systematic and mathematical way. We first discuss this approach, and later connect it to uncertainty in machine learning.

The general form of Bayes rule\marginnote{Bayes Rule}. Given a hypothesis ($H$) and some evidence ($E$), we usually have some previous information or assumption about the hypothesis validity, and then we observe some evidence which tells us some information about the hypothesis ($E \, | \, H$), then we want to find out how this evidence influences our current belief about the hypothesis ($H \, | \, E$).

This can be formalized as:
\begin{equation}
    \condprob{H}{E} = \frac{\condprob{E}{H} \prob{H}}{\prob{E}}
\end{equation}    
These terms have standardized names:
\begin{equation}
    \overbrace{\condprob{H}{E}}^{\text{Posterior}} = \frac{\overbrace{\condprob{E}{H}}^{\text{Likelihood}} \overbrace{\prob{H}}^{\text{Prior}}}{\underbrace{\prob{E}}_{\text{Evidence}}}
\end{equation}
Or in a simpler way:
\begin{equation}
    \text{Posterior} = \frac{\text{Likelihood} \times \text{Prior}}{\text{Evidence}}
\end{equation}
These terms have defined names that indicate their importance and conceptual behavior:
\begin{description}
    \item[Prior] Your previous belief about the hypothesis, or if no information is available, a general assumption about the hypothesis.
    \item[Likelihood] The probability of observing the evidence given the hypothesis, this measures the compatibility of the evidence with the hypothesis. Note that this is a function of the evidence.
    \item[Posterior] Your updated belief about the hypothesis, after seeing the evidence. Note that this is a function of the hypothesis.
\end{description}
\begin{examplebox}{SARS-CoV2 Infected Example}{covid_example}
	Let's assume that you travel somewhere and then feel sick. You take a SARS-CoV-2 test, and it returns positive. You get worried, but what is the chance that you are actually infected?	The hypothesis is are you infected with SARS-COV-2? Possible values are yes or no (binary).\\
	
	The evidence is your test, which is not perfect, let's assume that it is 80\% correct. That is $\condprob{E}{H} = 0.8$, you have 80\% evidence given that you are infected (this is the commonly reported metric for accuracy in a test).\\
	
	The prior is ratio of infected people over the whole population, let's assume it to be 0.1\%. Then $\prob{H} = 0.001$. Then first we compute $\prob{E}$ as:
	\begin{align*}
		\prob{E} &= \condprob{E}{H} \prob{H} + \condprob{E}{\neg H} \prob{\neg H}\\
		&= 0.8 \times 0.001 + 0.2 \times 0.9999 = 0.20078
	\end{align*}
	Then we can use Bayes rule to compute $\condprob{H}{E}$ and obtain an updated belief:
	\begin{align*}
		\condprob{H}{E} &= \frac{\condprob{E}{H} \prob{H}}{\prob{E}}\\
		&= \frac{0.8 \times 0.001}{0.20078} = 0.00398 \sim 0.4\%
	\end{align*}
	The posterior probability of being infected is only 0.4\%. This is only 4 times higher than the prior probability.
\end{examplebox}
\begin{examplebox}{SARS-CoV2 Not Infected Example}{covid_not_example}
	As a thought experiment, what is $\condprob{\neg H}{\neg E}$? It is the probability of not being infected given that the test is negative.\\
	
	To compute it we need some further information, which is $\condprob{\neg E}{\neg H} = 0.8$, as test accuracy is defined for both testing positive and being healthy, and testing negative and being sick, and these two terms are equalized.
	\begin{align*}
		\condprob{\neg H}{\neg E} &= \frac{\condprob{\neg E}{\neg H} \prob{\neg H}}{\prob{\neg E}}\\
		&= \frac{0.8 \times (1 - 0.001)}{1 - 0.20078} = 0.9999 \sim 99.99 \%
	\end{align*}
	This is the probability of not being infected given that the test is negative, and it is $>99\%$, due to the low prevalence of the disease.
\end{examplebox}

\section{Bayesian Views on Uncertainty}
\label{sec:bayes_view_unc}

Typically when learning a model, we wish to learn the probability distribution $\condprob{\textbf{y}}{\textbf{x}}$, that is, the probability of some output $y$ given input $x$. But this does not consider the model parameters $\textbf{w}$ (also known as weights), which in the end indirectly encode uncertainty.

Then a better approach for uncertainty estimation is learn $\condprob{\textbf{y}}{\textbf{x}, \, \textbf{w}}$ which directly considers the model parameters. As model parameters are learned from data $\mathbf{D}$, then we want to learn $\condprob{\mathbf{w}}{\textbf{D}}$, which is what training a model does.

There is a direct application of Bayesian principles for Neural Networks, which is used to build the concept of a \marginnote{Bayesian Neural Network}Bayesian Neural Network (BNN). They are neural networks, but where the weights are probability distributions, instead of scalar or point-wise weights. This allows weights to implicitly encode model uncertainty.

This Bayesian view has multiple practical interpretations:
\begin{description}
    \item[Reproducibility] If we train the same neural network multiple times on the same dataset, the results will not be exactly the same. They will be close in numerical terms, but not the same. This is due to random weight initialization, and variation of each weight as multiple models are trained can be seen as uncertainty in the weights themselves.
    \item[Lack of Knowledge] Weights that have larger uncertainty (variance), will produce the output to be more uncertain, so weights encode uncertainty about the output.
\end{description}

Bayes rule as defined for Bayesian Neural Nets is\marginnote{Bayes Rule in BNNs}:
\begin{equation}
	\condprob{\mathbf{w}}{\mathbf{D}} = \frac{\condprob{\mathbf{D}}{\mathbf{w}}\prob{\mathbf{w}}}{\prob{\mathbf{D}}}
\end{equation}
Where $\mathbf{w}$ are weights, $\mathbf{D}$ is your data points (as in $\mathbf{D} = \{(x_i, y_i)\}_{N}^{i=1}$). The distribution $\condprob{\mathbf{w}}{\mathbf{D}}$ is called the posterior distribution of the weights, that is, the weight distribution that is learned from the data. This is what training a model does in a broad sense.

\begin{description}
	\item[Prior] This is $\prob{\mathbf{w}}$, the prior distribution on the weights. This is generally assumed. This has direct connection with the random weight initializers that are used in ML/NN models.
	\item[Likelihood] This is $\condprob{\mathbf{D}}{\mathbf{w}}$, which is a forward pass of the model, which computes $\condprob{\mathbf{y}}{\mathbf{x}, \mathbf{w}}$ and for this it needs weights $\mathbf{w}$. These terms are defined by your model architecture/equations.
	\item[Posterior] This is $\condprob{\mathbf{w}}{\mathbf{D}}$, the posterior distribution of the weights given the data, which is the actual learning algorithm, and later can be used to make predictions.
\end{description}
\begin{definitionbox}{Uncertainty in Bayesian NNs}{uncertainty_bnns}
	From the Bayesian POV, a model is defined by both the structure (the actual model equations) and the model's parameters.
	This means that for a Bayesian model, the model parameters encode the \textbf{model uncertainty} about each prediction. There are equations to compute the output probability distribution from the model and its inputs.\\
	
	Intuitively, this can be seen as a simple question. If we have several models trained on the same data, the model's parameters won't be the same. Some parameters might be very close, some will be equal, and some will be radically different.\\
	
	These variations on model parameters can be encoded as a probability distribution. Bayesian statistics can be used to estimate these distributions $\condprob{\mathbf{w}}{\mathbf{D}}$
\end{definitionbox}
\begin{definitionbox}{Bayesian Neural Networks}{bayesian_nns}
	So what exactly is a Bayesian Neural Network? They have two components:
	\begin{description}
		\item[Weight Uncertainty] The weights of layers in the model are probability distributions. There can be point-wise weights, but the defining characteristic is probability distribution over weights. Does not have to be in all layers.
		\item[Bayesian Learning ] The Learning algorithm must consider that it learns probability distribution weights instead of point-wise weights. In a strict sense all learning algorithms are an application of Bayes rule.
	\end{description}
\end{definitionbox}
After learning the posterior distribution over the weights $\condprob{w}{D}$, then output distributions can be computed with the Bayesian Predictive Posterior Distribution:
\begin{equation}
    \condprob{\mathbf{y}}{\mathbf{x}} = \int_\mathbf{w} \condprob{\mathbf{y}}{\mathbf{w}, \mathbf{x}} \condprob{\mathbf{w}}{\mathbf{D}}d\mathbf{w}
    \label{bc:bayesian_pred_posterior}
\end{equation}
\marginnote{Bayesian Predictive\\Posterior Distribution}
This equation is obtained by marginalizing over all possible model parameters $\textbf{w}$. This equation basically computes predictions $\textbf{y}$ with different model parameters $\textbf{w}$ and weights them by the probability of those parameters given the data $\textbf{D}$. This can also be considered as a form of Bayesian Model Averaging.

Eq \ref{bc:bayesian_pred_posterior} can be decomposed into several terms inside the integral:
\begin{equation}
    \condprob{\mathbf{y}}{\mathbf{x}} = \int_\mathbf{w} \overbrace{\condprob{\mathbf{y}}{\mathbf{w}, \, \mathbf{x}}}^{\text{Forward Pass}} \underbrace{\condprob{\mathbf{w}}{\mathbf{D}}}_{\text{Posterior of Weights}} \, d\mathbf{w}
\end{equation}
\begin{description}
	\item[Forward Pass] This is one forward pass through your neural network, that outputs a probability distribution given inputs and weights.
	\item[Posterior of Weights] This is learned using a learning algorithm, and is basically the result of learning using Bayes rule.
\end{description}

A simple interpretation of this equation is that you integrate the product of a forward pass (given a specific set of weights) and the probability of that set of weights, over all weight values.

In general cases we cannot compute \ref{bc:bayesian_pred_posterior} for the Bayesian Predictive Posterior distribution, but it can be approximated using Monte Carlo with $M$ samples:
\begin{equation}
	\condprob{\mathbf{y}}{\mathbf{x}} \sim M^{-1} \sum_i^M \condprob{\mathbf{y}}{\theta_i, \, \mathbf{x}} \condprob{\theta_i}{\mathbf{D}} \quad \theta_i \sim \condprob{\mathbf{w}}{\mathbf{D}}
\end{equation}
Here the $\theta_i$ are samples from the posterior distribution of the weights $\condprob{\mathbf{w}}{\mathbf{D}}$. Larger values of $M$ means better approximation, but more computation, implying a trade-off between computation and approximation quality.
\begin{definitionbox}{Issues with BNNs}{issues_bnns}
	\textbf{Intractable Evidence}. If we consider Bayes Rule, the term $\prob{\mathbf{E}}$ is generally difficult or impossible to compute.	In a Bayesian Neural Network, the equivalent term is $\prob{\mathbf{D}}$, which is the distribution of the whole data (not the dataset), which in general is impossible to compute.\\
	
	For example if we use images of a given size, $\prob{\mathbf{D}}$ is the distribution of all images of that size, which we cannot estimate due to a lack of data, and a lack of methods due to its high dimensionality.
	
	\textbf{Large Weight Space}. The biggest issue with Bayesian learning methods is how to encode or represent the posterior probability distributions over the weights. Since typical usable models have millions of parameters, the distribution is very high-dimensional.
	
	\textbf{Computational Complexity}. Integrating over millions of parameters and performing \textit{multiple predictions}/\textit{multiple forward passes} for each of these parameters is computationally infeasible.
	
	\textbf{No Closed Form Solutions}. In general there are no closed form representations for posterior distribution over weights, and consequently there are also no closed form computations of the Bayesian predictive posterior distribution. The only alternative is to represent the distribution with samples (histograms), and to use Monte Carlo methods to sample from the posterior distribution.
\end{definitionbox}

\section{Maximum Likelihood Estimation}

Given data, how can we fit a probability distribution to it? Maximum Likelihood is a statistical principle/method that can be used to find parameters for a distribution. 

Maximum Likelihood Estimation (MLE) is a method to estimate parameters of probability distributions, but more commonly used to estimate parameters of models. Estimation happens given some data, through a process that models the joint probability distribution of that data, as the model would explain it, with the aim of finding the distribution parameters that best fit the data points.

First, given some data points $y_i \forall i \in [0, N]$ that are independent and identically distributed (i.i.d.), we compute the likelihood function:
\marginnote{Likelihood Function}
\begin{equation}
	L(\theta; y) = \prod_i f_X(y_i; \theta) 
	\label{bc:likelihood}
\end{equation}
Where $f_X$ is the PDF of a selected distribution with parameters $\theta$. Eq \ref{bc:likelihood} is called a likelihood function. This function gives you the joint probability of all data points $y$ under the assumed distribution. We can find the best parameters $\theta$ by maximizing the likelihood function under $\theta$:
\begin{equation}
    \theta^{*} = \argmax_\theta L(\theta; y)
\end{equation}
Parameters obtained using this method are called maximum likelihood estimates. Now the form of the likelihood function is not so easy to optimize in an analytical or closed form, an alternative is applying the logarithm:
\marginnote{Log-Likelihood Function}
\begin{equation}
	\log L(\theta; y) = \log \prod_i f_X(y_i; \theta) = \sum_i \log f_X(y_i; \theta)
	\label{bc:log_likelihood}
\end{equation}
The (natural) logarithm is a strictly increasing function, so it does not change the position (value) of the optimal parameters $\theta^{*}$, but this form is now much easier to optimize. Eq \ref{bc:log_likelihood} is called the log-likelihood.

The optimal parameters are usually found with gradient-based optimization, namely gradient ascent, or simply by solving $\frac{\partial \log L(\theta; y)}{\partial \theta} = 0$ analytically.

%% file: figures/model_misspecification.tikz
\begin{tikzpicture}
        \begin{axis}[width=0.45\linewidth, xmin=-10, xmax=40,ymin=-500, ymax=2000, xlabel={$x$}, ylabel={$f(x)$}]
            \addplot gnuplot[only marks, domain=0:30]{x^2 + 80 * rand(0)};
            \addplot[red, domain=-10:40] {33 * x};
            \addplot[green, domain=-10:40] {50 * x - 500};
            \addplot[blue, domain=-10:40] {0.05 * x^3};
            \addplot[black, domain=-10:31] {exp(x/4)};
        \end{axis}
\end{tikzpicture}

%% file: figures/kitti_semseg_entropy.tikz
\begin{tikzpicture}
		\begin{axis}[width=0.45\linewidth, scatter/classes={a={mark=o,draw=black}}, ylabel=Entropy, xlabel={\# of Samples}, xmode=log]
			\addplot[scatter, only marks, scatter src=explicit symbolic,
					 nodes near coords*={\annotvalue},
					 node near coord style={rotate=30, anchor=south west, font=\scriptsize},
					 visualization depends on={value \thisrow{annotation} \as \annotvalue}]
			table[meta=label]{
				x 		y 		label 	annotation
				1.0e5	0.95	a		motorcyclist
				0.4e6	0.8		b		bicyclist
				1.0e7	0.4		c		pole
				1.0e8	0.18	d		car
				0.4e9	0.36	e		sidewalk
				0.7e8	0.20	f		road
			};
		\end{axis}
\end{tikzpicture}

%% file: sections/methods.tex
\chapter{Methods for Uncertainty Quantification}
\label{ch:methodsuq}
\margintoc

\section{Concept}

There are several general concepts that are used to implement uncertainty quantification in a specific machine learning model, below we provide a short summary.

\begin{description}
    \item[Integration] In order to incorporate or "add" uncertainty quantification to a neural network or model, some modifications are required to the network architecture, the training process, or both. This depends on the actual uncertainty quantification method, for example, replacing standard layers with new layers (i.g Dense to DenseFlipout), or using a different loss function that supervises a variance output.
    \item[Architectures] There are some meta-architectures for uncertainty quantification, for example, to predict a distribution, a general way is for a model to predict the parameters of the distribution, and this is implemented as a multi-head architecture, where each head predicts a single parameter. For a Gaussian distribution, mean $\mu$ and variance $\sigma^2$ are predicted and this is generally called a two head model.
    \item[Models] In practice, any neural network architecture can be modified to support uncertainty quantification. This creates a new model or architecture that is different from the original. Predictive performance is usually affected, as adding uncertainty quantification can decrease or increase task predictive performance. We will describe some of these effects later in this section.
\end{description}

\section{Architectures}

\begin{marginfigure}
    \centering
    \begin{tikzpicture}[style={align=center, minimum height=0.5cm, minimum width = 1.7cm}]
        \node[] (dummy) {};
        \node[rounded corners, draw, left=0.001em of dummy] (objClassifier) {{Mean Head}};
        \node[above=1em of objClassifier] (objClass) {{$\mu(x)$}};
        
        \node[rounded corners, draw, right=0.01em of dummy] (objDetector) {{Variance Head}};
        \node[above=1em of objDetector] (obj) {{$\sigma^2(x)$}};
        
        \node[below=2em of dummy] (fc) {{Feature Vector}};
        
        \node[rounded corners, draw, below=1em of fc](conv) {{Trunk Network}};
        \node[below=1em of conv](inputImage) {{Input $x$}};
        \draw[-latex] (inputImage) -- (conv);
        \draw[-latex] (conv) -- (fc);
        \draw[-latex] (fc) -- (objDetector);
        \draw[-latex] (fc) -- (objClassifier);
        \draw[-latex] (objClassifier) -- (objClass);
        \draw[-latex] (objDetector) -- (obj);
    \end{tikzpicture}
    \caption{Structure of the two-head architecture for aleatoric uncertainty estimation in regression.}
    \label{methods:reg_twohead}
\end{marginfigure}
One basic architecture used for uncertainty quantification is the two-head architecture. The idea is that to estimate data uncertainty, we can add an additional output head that can estimate an uncertainty or confidence value together with the actual prediction made by the model. This uncertainty or confidence value is usually the variance or standard deviation of the output.

This concept basically corresponds to a model giving a probability distribution as an output, but instead of a distribution, the parameters of that distribution are output by the model, which are easier to interpret intuitively to a human.

The two-head architecture is show in Figure \ref{methods:reg_twohead}. The model contains two output heads, a head predicting the distribution mean $\mu(x)$, and a head predicting a variance $\sigma^2(x)$ or similar confidence value. Both heads receive input from a common trunk network that produces a feature vector.

The two-headed architecture has to be combined with a proper loss function that provides supervision for the mean and variance heads. These losses are described in the next section.

There is one additional major architectural pattern to add uncertainty quantification to a model relates to the two kinds of uncertainty:

\begin{description}
    \item[Data Uncertainty] Requires a specially crafted loss functions that considers uncertainty, and for regression, an additional output head for data variance (two-headed architecture).
    \item[Model Uncertainty] Properly estimating model uncertainty requires changes to the model architecture or making assumptions about the architecture or features.
\end{description}

In order to estimate both kinds of uncertainties, the two architectural modifications described above have to be combined into a new architecture that produces predictive uncertainty. It is also possible to obtain separate estimates of aleatoric and epistemic uncertainty using a particular formulation (covered in Sec \ref{sec:disentangle}).

\section{Losses}

A loss function is what provides direct supervision to train a model. Examples are cross-entropy, mean squared error, etc. But these standard losses do not always capture output uncertainty, and different loss functions are needed.

There are two loss functions that can be used to learn an uncertainty value in combination with the two-head architecture for regression.

First is the Gaussian negative log-likelihood \marginnote{Gaussian NLL}

\begin{equation}
    L(y, \mu(x),\sigma^2(x)) = \sum_{i} \left[\log \sigma_i^2 + \frac{(y_i - \mu_i)^2}{\sigma_i^2} \right].
    \label{eq:gaussian_NLL}
\end{equation}

This loss corresponds to the model outputting the parameters of a Gaussian distribution (mean $\mu(x)$ and variance $\sigma^2(x)$), 

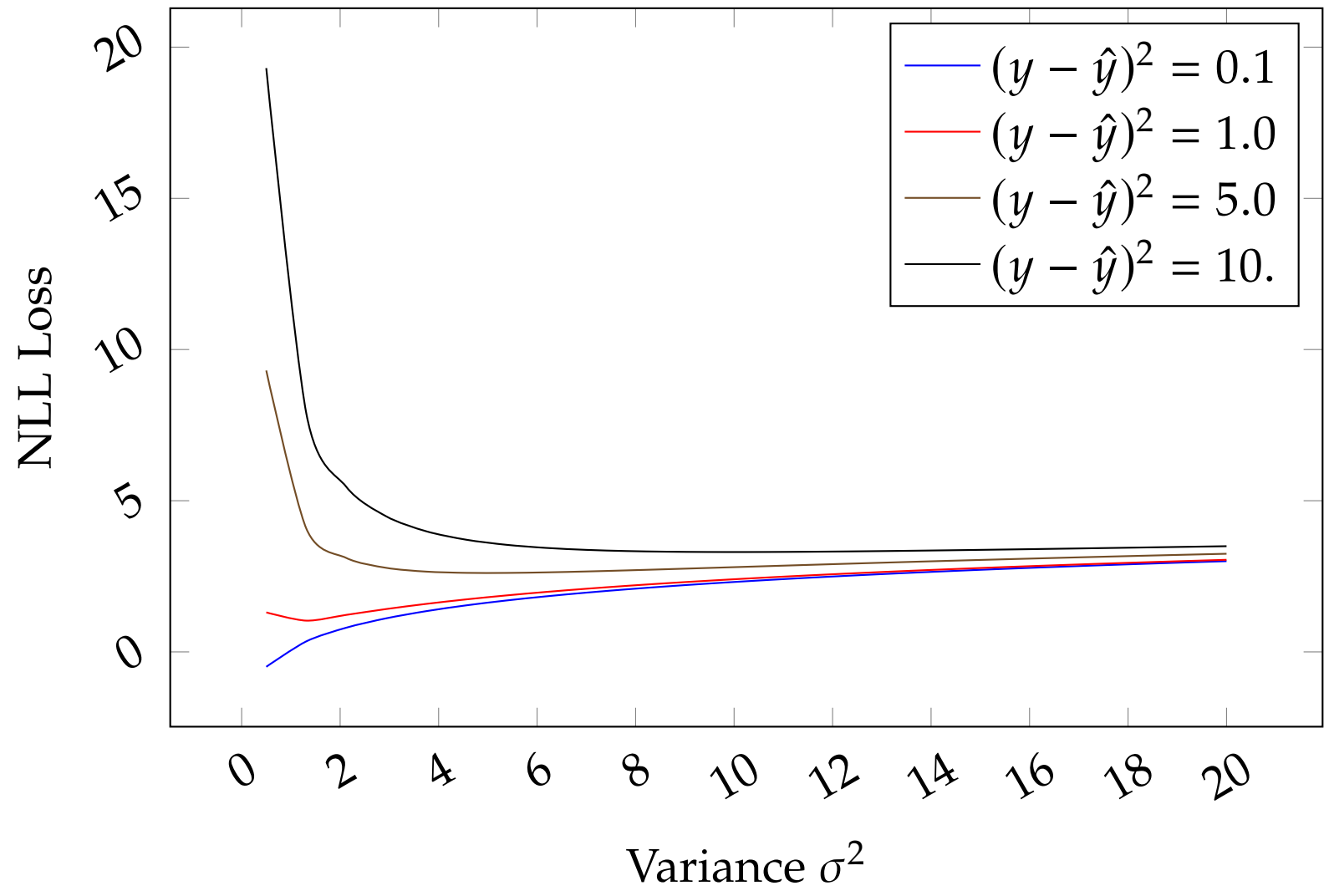
\begin{figure}
    \begin{tikzpicture}        
        \begin{axis}[width=\linewidth, height=0.3\textheight, domain=0.5:20, tick label style={rotate=30}, xlabel={Variance $\sigma^2$}, ylabel={NLL Loss}]
            \addplot+[mark=none, smooth] {ln(x) + 0.1 / x};
            \addlegendentry{$(y-\hat{y})^2 = 0.1$}
            \addplot+[mark=none, smooth] {ln(x) + 1.0 / x};
            \addlegendentry{$(y-\hat{y})^2 = 1.0$}
            \addplot+[mark=none, smooth] {ln(x) + 5.0 / x};
            \addlegendentry{$(y-\hat{y})^2 = 5.0$}
            \addplot+[mark=none, smooth] {ln(x) + 10.0 / x};
            \addlegendentry{$(y-\hat{y})^2 = 10.$}

        \end{axis}
    \end{tikzpicture}
    \caption[Comparison of the Gaussian NLL with different simulated squared errors as function of the variance $\sigma^2$]{Comparison of the Gaussian NLL with different simulated squared errors as function of the variance $\sigma^2$, showing the attenuating effect of this loss with respect to the variance. Note how the minima of each plotted loss is difference, and decreasing with a smaller squared error.}
\end{figure}

Second is the Laplace negative log-likelihood \marginnote{Laplace NLL}

\begin{equation}
    L(y, \mu(x),s(x)) = \sum_{i} \log s + \frac{|y_i - \mu_i|}{s}.
    \label{eq:laplace_NLL}
\end{equation}

Which also corresponds to a model output begin the parameters of a distribution, but its a Laplace distribution this time. The parameters are location $\mu(x)$ and scale or diversity $s(x)$. These are functionally similar to the mean and standard deviation of a Gaussian distribution. But these two losses make slightly different assumptions, related to the weight of the tails of the Gaussian vs Laplace distributions.

\begin{marginfigure}
    \begin{tikzpicture}
        
        \begin{axis}[width=\linewidth, domain=-7:7, tick label style={rotate=30}]
            \addplot+[mark=none, smooth] {gauss(0.0, 1.0)};            
            \addplot+[mark=none, smooth, dashed] {laplacepdf(0.0, 1.0)};
        \end{axis}
    \end{tikzpicture}
    \caption{Comparison of Gaussian and Laplace distributions with zero mean/location and unit standard deviation/scale.}
    \label{fig:gaussian_vs_laplace}
\end{marginfigure}

The derivation of both losses using maximum likelihood estimation can be found in the appendix. Note that both losses have a similar structure, with a uncertainty logarithm term, and a error (squared or absolute) term divided by the uncertainty.

Additionally, we should note that the mean $\mu_i$ is supervised by the label $y_i$, while the variance $\sigma_i^2$ does not have direct supervision, but the loss has a degree of influence on it during training.

\begin{description}
    \item[$\sigma_i^2 \rightarrow \infty$] The logarithm term goes to $\infty$ and the squared error term tends to 0.
    \item[$\sigma_i^2 \rightarrow 0$] The logarithm term goes to $-\infty$ and the squared error term tends to $\infty$.
\end{description}

This loss is also called variance attenuation, can we see why?

\begin{description}
    \item[$(y_i - \mu_i)^2$ Small] Variance is usually small.
    \item[$(y_i - \mu_i)^2$ Big] Variance is usually large.
\end{description}

When the model predicts correctly, the variance is also small, but when the model predicts incorrectly (large squared error), then the only way to minimize the loss is to increase the variance.

\section{Direct Methods}
Direct methods for uncertainty quantification are method that directly output uncertainty estimates, not needing sampling or other intermediate processes.

\subsection{Ensembles}

Ensembling is the general concept of combining the predictions of multiple machine learning models in order to improve performance \citep{bishop2006pattern}. Examples of this concept are Random Forests and Adaboost. Ensembling of neural networks which was initially proposed by \cite{lakshminarayanan2017simple}.

Consider that we have $M > 0$ models (with a typical value $M = 5$, \cite{ovadia2019can}) in our ensemble. To train the ensemble, each of the $M$ models is trained on the same dataset, using the same neural network architecture, hyper-parameters, etc. The only difference between these models is their initial weights, which are randomly initialized (using the same initializers), but due to this initialization, each model will receive different samples for each of the weight vectors and matrices.

During training, due to different random initial weights, each model will converge to a different local solution. It has been demonstrated that these solutions are most of the time very close in terms of loss value and/or accuracy, so they are in a sense different but equivalent. To make predictions at inference time, the predictions of each ensemble member $p_i$ are combined according to the task.

For classification, the softmax probabilities \marginnote{Classification} output by each ensemble member $p_i((y \, | \, \mathbf{x})) \sim \Softmax{z(\mathbf{x})}$ are averaged: 
\begin{equation}
    p_e(y \, | \, \mathbf{x}) = M^{-1}\sum_i p_i(y \, | \, \mathbf{x})    
\end{equation}
For regression, there are two possible cases. \marginnote{Gaussian Regression} The first is each ensemble member performing regression of only the mean $\mu_i$, which implies only model uncertainty is estimated. The ensemble output distribution is then $p_e(y \, | \, \mathbf{x}) \sim \Gaussian{\mu_e, \sigma^2_e}$
\begin{align}
    \mu_e(\mathbf{x}) &= M^{-1} \sum_i \mu_i(\mathbf{x})\\
    \sigma^2_e(\mathbf{x}) &= (M - 1)^{-1} \sum_i (\mu_i(\mathbf{x}) - \mu_e(\mathbf{x}))^2
    \label{eq:ensemble_reg_combination}
\end{align}
The second case is \marginnote{Gaussian Mixture Regression} each ensemble member performing regression of both the mean $\mu_i$ and variance $\sigma^2_i$ of a Gaussian distribution.  The ensemble output distribution is then a Gaussian mixture $p_e$. This method estimates both data and model uncertainty.
\begin{align}
    p_e(y \, | \, \mathbf{x}) &\sim \mathcal{N}(\mu_*(\mathbf{x}), \sigma^2_*(\mathbf{x}))\\
    \mu_*(\mathbf{x}) &= M^{-1} \sum_i \mu_i(\mathbf{x})\\
    \sigma^2_*(\mathbf{x}) &= M^{-1} \sum_i [\sigma^2_i(\mathbf{x}) + \mu^2_i(\mathbf{x})] - \mu^2_*(\mathbf{x})
\end{align}
The Gaussian mixture mean $\mu_*$ is the mean of the ensemble members $\mu_i$, while the Gaussian mixture variance is computed using the law of total variance. Note that here we approximate the Gaussian mixture with a single Gaussian, which makes implicit assumptions. It is possible to output the full Gaussian mixture but this is not common as it makes uncertainty interpretation more complex.

With the ensembling framework, it is possible to make separate predictions of data and model uncertainty. This formulation is presented in Sec \ref{sec:disentangle}.

\subsection{Feature-space Uncertainty}

Uncertainty can also be estimated from the distribution of features. The general concept is that for epistemic uncertainty, it is desirable for it explicitly follow the distribution of the training set, meaning that low training set density implies high epistemic uncertainty, and high training set density implies low epistemic uncertainty. Then epistemic uncertainty directly follows the distribution of the training set.

In this category there are two methods for epistemic uncertainty estimation, DUQ and DDU, and both need changes to the model. The overall advantage of these methods is the requirement of using a single neural network, no ensembling or sampling is required, and they produce high quality epistemic uncertainty.

\subsubsection{Direct Uncertainty Quantification}

The overall idea of DUQ \citep{van2020uncertainty} is to model each class in feature space using a centroid that is learnable in the training phase, and during inference, use the distance to the closest centroid in feature space as a measure of uncertainty.

First DUQ defines a function $K_c$ which represents a radial basis function (RBF), replacing the standard Dense layer with a softmax activation. The input to this function is a trainable feature extractor network $f_\theta(x)$ output, which is a standard neural network architecture without the last layer (usually the task layers).

\begin{equation}
    K_c(f_\theta(x), e_c) = \exp \left(-\frac{n^{-1}||W_c f_\theta(x) - e_c||_2^2}{2\sigma^2} \right)
\end{equation}

Each class $c$ has a centroid $e_c$ in feature space, from where a distance is computed to $f_\theta(x)$, and passed through a exponential function to transform distances to pseudo-probabilities or kernel distances. $W_c$ is a per-class learnable weight matrix mapping from original feature space (produced by $f_\theta(x)$) to a reduced feature space, and has dimensions $m \times n$.

The value of $\sigma^2$ is a scaling factor, similar to the variance of a Gaussian distribution, but in feature space.

The loss function is binary cross-entropy applied to $K_c$ as a vector of pseudo-probabilities of length $c$, even in the multi-class setting, due to $K_c$ being a vector. Since the RBF layer interacts with the feature learning process during training, feature collapse is possible, and to prevent it, regularization is applied. DUQ in particular uses a gradient penalty that is added to the loss function.
\begin{equation}
    L_p(x) = \lambda \left(|| \nabla_x \sum K_c||_2^2 - 1\right)^2
\end{equation}
The gradient penalty aims to constrain the Lipschitz constant of layers to 1, which helps smoothness of $K_c$ and improves kernel distances. $\lambda$ is the standard regularization strength coefficient that should be tuned using cross-validation.

The hyper-parameters of DUQ are the feature space scaling $\sigma$ and dimension of the reduced feature space $n$. Class predictions can be made by taking the class with maximum $K_c$, implying closest distance to $e_c$ in feature space.
\begin{equation}
    \tilde{\mathbb{P}}(y = c \,| \, x) = \argmax_c K_c(f_\theta(x), e_c)
\end{equation}
Note the use of $\tilde{\mathbb{P}}$, as these are not probabilities, they can be called unnormalized probabilities as they do not sum to 1.0, and are more commonly called kernel distances.

\subsubsection{Deep Deterministic Uncertainty}

DDU \citep{mukhoti2023deep} follows from the basic design of DUQ but it has considerable differences. There is no use of a RBF output, but it uses the standard Dense layer with a softmax activation, trained using categorical cross-entropy. This basic setup is able to estimate aleatoric uncertainty.

To estimate epistemic uncertainty, DDU uses a Gaussian mixture to model feature density, in particular Gaussian Discriminant Analysis (GDA) which models each class with an individual Gaussian distribution. Given a feature extractor sub-network $f_\theta(x)$, the mean and covariance matrices are computed on the training set for each class:
\begin{equation}
    \mu_c = \text{mean}([f_\theta(x) \, | \, x \in \text{Train}_c])
\end{equation}
\begin{equation}
    \sigma^2_c = \text{cov}([f_\theta(x) \, | \, x \in \text{Train}_c])
\end{equation}

For new test sample $x$, epistemic uncertainty is computed as the negative log-density under the learned feature distribution using GDA (for all classes using $\mu_c$ and $\sigma^2_c$. A sample with low log-density indicates low epistemic uncertainty, being a close or inside the training set, and a sample with high log-density indicates high epistemic uncertainty, being far from the training set (implying a density of almost zero).

In order for the learned feature space to be distance aware, it must be regularized. Spectral normalization is commonly used, which helps control the Lipschitz constant of a layer. This constant can be estimated by the largest eigenvalue of the weight matrix, denoted as $ \sigma(w)$\marginline{Do not confuse $\sigma(w)$ with the standard deviation, which also is denoted with $\sigma$}, and is also known as the spectral norm.
\begin{equation}
    \tilde{W}(w) = \frac{w}{\sigma(w)}
\end{equation}
During training, the weight matrix of layers are divided by their spectral norm, which sets their Lipschitz constant to 1, smoothing training and preventing feature collapse, making the feature space learned by $f_\theta(x)$ distance-aware.

DDU is considerably easier to implement than DUQ, as only spectral normalization needs to be applied, and a density model learned from training set features. DDU is able to separately estimate aleatoric and epistemic uncertainty, while DDU cannot.

\subsection{Quantile Regression}
\begin{marginfigure}
    \begin{tikzpicture}[declare function={
            func(\x)= (x>=0) * ((1 - 0.4) * x)  +
            (x<0) * (-0.4 * \x);            
        }
        ]
        \begin{axis}[ymin=0, ymax=3, ylabel=$L(\tau)$, xmin=-3, xmax=3, xlabel=$\hat{y} - y$,width=0.9\textwidth]
            \addplot[blue, domain=-3:3, smooth]{func(x)};
        \end{axis}
    \end{tikzpicture}
    \caption{Plot of the Pinball loss for $\tau=0.4$}
\end{marginfigure}
A quantile is division of the range/domain of a probability distribution in fixed cumulative probability density steps. This is one way to represent a probability distribution, without making assumptions on specific distributions. This is called "distribution free" methods.
\begin{marginfigure}
	\begin{tikzpicture}[declare function={
			func(\x)= (x>=0) * ((1 - 0.7) * x)  +
			(x<0) * (-0.7 * \x);            
		}
		]
		\begin{axis}[ymin=0, ymax=3, ylabel=$L(\tau)$, xmin=-3, xmax=3, xlabel=$\hat{y} - y$,width=0.9\textwidth]
			\addplot[blue, domain=-3:3, smooth]{func(x)};
		\end{axis}
	\end{tikzpicture}
	\caption{Plot of the Pinball loss for $\tau=0.7$}
\end{marginfigure}
Quantile regression, as the name implies, performs regression of a specific quantile of the output distribution. This is made with a special loss function, the quantile loss:
\begin{equation}
    L(y, \hat{y}, \tau) = \begin{cases}
        (1 - \tau)(y - \hat{y} ) & \quad \text{if } \hat{y} > y,\\
        \tau(y - \hat{y}) 		& \quad \text{if } \hat{y} \leq y.
    \end{cases}
\end{equation}

A compact way to write the quantile loss is by using the indicator function $\mathbb{I}$ is $L(y, \hat{y}, \tau) = (\hat{y} - y)(\mathbb{I}[y \leq \hat{y}] - \tau)$.

The quantile loss has one parameter, the quantile $\tau \in [0, 1]$, this allows to perform regression of a single quantile $\tau$ of the output distribution.

To perform regression of multiple quantiles, then multiple output heads can be used, each with its own instance of the pinball loss with different values of $\tau$.

Quantile regression is very flexible, as the user can configure which quantiles to output, controlling computational complexity, and the quantile representation is distribution-free, there is no assumption on a particular distribution.

\section{Sampling-Based Methods}

Sampling-based methods are the ones that for each stochastic forward pass, they produce a sample of the output distribution that encodes uncertainty. They generally provide higher quality uncertainty estimates, but they are expensive to compute as multiple forward passes are required to reconstruct the output probability distribution and there is approximation error due to sampling.

All sampling methods use the predictive posterior distribution to make predictions (Eq \ref{bc:bayesian_pred_posterior}) or approximations when only samples of the predictive posterior are available.

\subsection[MC Dropout]{Monte Carlo Dropout}

Dropout is a well known technique aimed at regularizing neural networks. Activations $z$ from a layer are multiplied with a mask $m$, where $m \sim \Bernoulli{p}$ is a mask drawn from a Bernoulli distribution with probability $p$.
\begin{equation}
    z' = z \circleddot m
\end{equation}
The mask $m$ is a binary vector of length $n = |z|$, then the product $z \circleddot m$ has the effect of setting some activations to zero, while keeping other activations unchanged. This effect is applied during training with different randomly sampled masks, producing a stochastic effect that breaks correlations between neurons. 

Dropout, when enabled at inference time, has been show to produce approximations of the Bayesian posterior distribution \citep{gal2016dropout}. Each forward pass of a Model using Dropout enabled at inference time, will produce one sample of the approximate posterior, and through aggregating information from multiple samples, the approximate posterior distribution can be reconstructed.

The combination of Dropout at inference time is usually called Monte Carlo Dropout (or short, MC-Dropout).

\subsection[MC DropConnect]{Monte Carlo DropConnect}

DropConnect \citep{wan2013regularization} is a variation of Dropout, where a binary mask $m \sim \Bernoulli{p}$ is drawn from a Bernoulli distribution, but it is (component-wise) multiplied with the weights of a layer, instead of being applied to the activations. It also produces a regularizing effect.
\begin{align*}
    m &\sim \Bernoulli{p}\\
    w' &= m \circleddot w\\
    z' &= \text{activation}(w' x + b)
\end{align*}
The idea is to reduce the dependence of weights with each other, similarly to how Dropout does the same with activations. This requires implementing new custom instances of layers with the DropConnect effect applied to their internal weights, unlike Dropout which has a generic implementation that works with most layers.

DropConnect can also be enabled at inference time, and it has been shown to produce an approximate Bayesian posterior distribution \citep{mobiny2021dropconnect}. The combination of DropConnect at inference time is usually called Monte Carlo DropConnect (or short, MC-DropConnect).

\section[Bayesian NNs]{Bayesian Neural Networks}
This section expands Section \ref{sec:bayes_view_unc} with methods to learn the posterior distribution of the weights in a practical way, and enable the learning of Bayesian Neural Networks.

\subsection[VI Framework]{Variational Inference Framework}

\begin{marginfigure}
    \begin{tikzpicture}
        \begin{axis}[width=1.3\linewidth, domain=-6:10, xmajorticks=false, ymajorticks=false]
            \addplot+[mark=none, smooth, blue] {gauss(0.5, 1.0) + 0.5 * gauss(2.0, 2.0) + gauss(5.0, 1.2)};
            \addlegendentry{True}
            \addplot+[mark=none, smooth, red, dashed] {gauss(0.3, 1.0)};
            \addlegendentry{$q_\theta$}
        \end{axis}
    \end{tikzpicture}
    \caption{Visual representation of the approximating distribution $q_\theta$ and the true distribution. The core idea of variational inference is to find parameters $\theta$ that minimize the difference between these distributions, but the true distribution is unknown.}
\end{marginfigure}
The overall idea in the variational inference framework for BNNs is to approximate the posterior distribution of the weights $\condprob{w}{D}$ with a parametric distribution $\condprobvar{q_\theta}{w}{D}$. Here the distribution parameters are $\theta$, which implies that now the model has two kinds of parameters. First the approximate posterior distribution parameters $\theta$, and the weight distribution $\condprobvar{q_\theta}{w}{D}$, which can be sampled to produce weights from where a forward pass can be made.

In order to build the approximate posterior distribution, we must have a way to approximate it, and measure an error, same as we train standard machine learning models. A distance metric for distributions is the Kullback-Leibler divergence. Conceptually, we can propose to minimize the following loss:
\begin{equation}
    L(\theta) = \text{KL}(\condprobvar{q_\theta}{W}{D}, \condprob{W}{D})
\end{equation}
The problem with the previous loss is that we do not have a closed form or value for the posterior distribution of the weights $\condprob{W}{D}$. But with some mathematical tricks we can obtain a good approximation.
\begin{align*}
    \theta^* &= \argmin_\theta \text{KL}(\condprobvar{q_\theta}{W}{D}, \condprob{W}{D})\\
    &= \argmin_\theta \int_W 2 \condprobvar{q_\theta}{w}{D} \log \frac{\condprobvar{q_\theta}{w}{D}}{\prob{w} \condprob{D}{w}}dw\\
    &= \argmin_\theta \int_W \condprobvar{q_\theta}{w}{D} \log \frac{\condprobvar{q_\theta}{w}{D}}{\prob{w}} - \int_W \log \condprob{D}{w}dw\\
    &= \argmin_\theta \text{KL}(\condprobvar{q_\theta}{W}{D}, \prob{W}) - \expectover{\log \condprob{D}{W} }{\condprobvar{q_\theta}{W}{D}}
\end{align*}
\begin{marginfigure}
    \begin{tikzpicture}
        \begin{axis}[width=1.3\linewidth, domain=0:10, xmajorticks=false, ymajorticks=false]
            \addplot+[mark=none, red, dashed] {(x-5)^2.0 + 5*x};
            \addplot+[mark=none, blue] {x^1.2 + + 5*x + 120 + 10 * sin(120 * x)} node[below=0.5em, pos=0.9] {ELBO};
            \addplot+[mark=none, black] {x^1.05 + 300} node[below, pos=0.9] {$\log Z(\theta)$};
            
            \draw[black, <->] (2.0,19.0) -- (2,120) node[midway, right]{$-J(q)$};
            \draw[black, <->] (2.0,120.0) -- (2,302) node[midway, right]{$KL(q, p)$};
        \end{axis}
    \end{tikzpicture}
    \caption{Visual representation of relationships for the ELBO. Note how $J(q)$ is plotted as dashed (in red) because it is not an exact function since it is a lower bound.}
\end{marginfigure}
Here we replaced the definition of KL divergence, applied some rules of logarithms and sampling of the weights $w \sim \condprobvar{q_\theta}{W}{D}$.

But what is the trick to allow for this to happen? We replaced $\condprob{w}{D}$ with $\prob{w} \condprob{D}{w}$, this is an application of Bayes rule, but we missed the $\prob{D}$ term, so this produces \emph{unnormalized} probabilities.

Let's denote $\tilde{p}(x)$ our unnormalized probability, and $Z(\theta)$ the normalizing constant of bayes rule (the evidence). Then we can derive:
\begin{align*}
    J(q) &= \int_x q(x) \log \frac{q(x)}{\tilde{p}(x)}\\
    &= \int_x q(x) \log \frac{q(x)}{p(x)} - \log Z(\theta)\\
    &= \text{KL}(q, p) - \log Z(\theta)
\end{align*}
If we rearrange these terms to obtain the log-evidence, and noting that $\text{KL}(q, p) \geq 0$:
\begin{equation}
    \log Z(\theta) = \text{KL}(q, p) - J(q) \geq -J(q)
\end{equation}
This means that $-J(q)$ is a lower bound on the log-evidence $\log Z(\theta)$

These forms are called Evidence Lower BOund (ELBO)\marginnote{Evidence Lower Bound (ELBO)}, particularly for $-J(q)$. Note that there are multiple versions of the ELBO that are all equivalent.\\

The form above shows that the difference between $\log Z(\theta)$ and $-J(q)$ is $\text{KL}(q, p)$, so by maximizing the ELBO, the $\text{KL}(q, p)$ is minimized by squeezing it between $-J(q)$ and 
$\log Z(\theta)$.

Going back to our minimization problem, we obtained the following problem:
\begin{equation}
    \theta^* = \argmin_\theta \text{KL}(\condprobvar{q_\theta}{W}{D}, \prob{W}) - \expectover{\log \condprob{D}{W} }{\condprobvar{q_\theta}{W}{D}}
\end{equation}
This approximating formulation is possible due to multiple reasons:
\begin{itemize}
    \item By minimizing the KL between the approximating distribution $\condprobvar{q_\theta}{w}{D}$ and the prior distribution $\prob{w}$, we are maximizing the ELBO, and thus the log-evidence ($\log Z(\theta)$).
    \item By using unnormalized probabilities when applying bayes rule (leaving the evidence term out), the difference between the two formulations is exactly the log-evidence $\log Z(\theta)$.
\end{itemize}

In the end, by using an approximating distribution for the weights and the KL divergence between this distribution and the prior, we can be sure that learning is possible, even as the approximation quality is not known, but it is bounded through the ELBO.

We can wrap up the previous theory into a single loss function that can be used to learn a BNN using variational inference:
\begin{equation}
    L(\theta) = \underbrace{\text{KL}(\condprobvar{q_\theta}{W}{D}, \prob{W})}_{\text{KLD prior vs posterior}} - \underbrace{\expectover{\log \condprob{D}{W} }{\condprobvar{q_\theta}{W}{D}}}_{\text{NLL Loss}}
\end{equation}
The terms in this loss function can be interpreted as:\\

\begin{itemize}
    \item $\text{KL}(\condprobvar{q_\theta}{W}{D}, \prob{w})$ is the KL divergence between your approximate posterior of the weights $\condprobvar{q_\theta}{W}{D}$ and the prior $\prob{w}$. This term makes weights follow the prior distribution closely.
    \item $\expectover{-\log \condprob{D}{W}}{\condprobvar{q_\theta}{W}{D}}$ is the loss of your model (negative log-likelihood) but taken as expectation over the approximate posterior of the weights $\condprobvar{q_\theta}{W}{D}$.
\end{itemize}

Overall the loss is a combination of both terms, indicating that the trained model has to both have weight distributions that follow the prior distribution, while at the same time making correct predictions with these weight distributions. Thus training a Bayesian Neural Network using Variational Inference consists of setting up weights as probability distributions, and adding the KL divergence term between prior and weight distribution to the training loss of your model. The optimization process is performed using stochastic gradient descent, now on the approximate weight distribution parameters $\theta$ instead.

Note that the negative log-likehood term for the above loss has to be computed as an expectation over the approximate posterior $q_\theta$, which means that weights are sampled from $q_\theta$ and used to make forward passes and compute the loss $\log \condprob{D}{w}$.
\begin{equation}
    \expectover{\log \condprob{D}{w} }{\condprobvar{q_\theta}{w}{D}} = \int_W -\log \condprob{D}{w}dw \quad W \sim \condprobvar{q_\theta}{w}{D}
    \label{eq:variational_nll}
\end{equation}
Here we explicitly indicate that the set of weights for each forward pass and NLL computation is sampled from the posterior distribution of the weights, which is denoted by $w \sim \condprobvar{q_\theta}{w}{D}$.

A monte carlo approximation of Eq \ref{eq:variational_nll} is also possible:
\begin{equation}
    \expectover{\log \condprob{D}{w} }{\condprobvar{q_\theta}{w}{D}} \approx M^{-1} \sum^{M}_{M=1} -\log \condprob{D}{w} \quad w \sim \condprobvar{q_\theta}{w}{D}
\end{equation}

\begin{marginfigure}
    \subfloat[$\sigma_1 = 0.1, \sigma_2 = 1.5$]{
        \begin{tikzpicture}
            \begin{axis}[no markers, samples=100, domain=-5:5, width=1.3\linewidth, xtick=\empty, ytick=\empty, axis lines*=left, xlabel=$w$, ylabel=Density]
                \addplot [cyan!50!black] {gauss(0.0, 0.1) + gauss(0.0, 1.5)};
            \end{axis}
        \end{tikzpicture}
    }
    
    \subfloat[$\sigma_1 = 0.3, \sigma_2 = 1.5$]{
        \begin{tikzpicture}
            \begin{axis}[no markers, samples=100, domain=-5:5, width=1.3\linewidth, xtick=\empty, ytick=\empty, axis lines*=left, xlabel=$w$, ylabel=Density]
                \addplot [cyan!50!black] {gauss(0.0, 0.3) + gauss(0.0, 1.5)};
            \end{axis}
        \end{tikzpicture}
    }
    
    \subfloat[$\sigma_1 = 0.5, \sigma_2 = 1.5$]{
        \begin{tikzpicture}
            \begin{axis}[no markers, samples=100, domain=-5:5, width=1.3\linewidth, xtick=\empty, ytick=\empty, axis lines*=left, xlabel=$w$, ylabel=Density]
                \addplot [cyan!50!black] {gauss(0.0, 0.5) + gauss(0.0, 1.5)};
            \end{axis}
        \end{tikzpicture}
    }
    \caption{Prior distributions used by Bayes by Backprop, with varying values of $\sigma_1$}
    \label{methods:bbb_prior}
\end{marginfigure}
The term inside the sum is evaluated $M$ times, and each time a different set of weights $w$ randomly sampled from the approximate weight distribution $w \sim \condprobvar{q_\theta}{w}{D}$, weights which are used to make forward passes to compute the loss.

\subsection{Bayes by Backpropagation}
Bayes by Backpropagation (BBB) is a method proposed by \cite{blundell2015weight} to train Bayesian Neural Networks using Variational Inference. It uses Gaussian distributed weigthts $w \sim \Gaussian{\mu}{\sigma^2}$ together with a specific prior distribution. The parameters of a BNN using BBB are:
\begin{equation}
    \theta = [ \mu_i, \sigma_i^2 ]_i
\end{equation}
Each weight is represented by a mean $\mu$ and a variance $\sigma^2$, implying that the approximating distribution $\condprobvar{q_\theta}{w}{D} \sim \Gaussian{\mu}{\sigma^2}$ . This means that a BBB model has two times the number of parameters of the same model using standard neural network layers.

The prior used by BBB is a Gaussian mixture:
\begin{equation}
    \prob{W} = \Gaussian{0.0}{\sigma_1^2} + \Gaussian{0.0}{\sigma_2^2}
\end{equation}
By default this prior uses $\sigma_1 = 0.1$ and $\sigma_2 = 1.5$, but these values can be configured by the user. The different shapes of the prior distribution are shown in Figure \ref{methods:bbb_prior}.

\subsection{Flipout}

In the standard implementation of BBB, one major issue is training instabilities. Due to the model now being stochastic and training/predictions relying on sampling from the approximate posterior distribution of the weights, the loss function is also stochastic and can have significant variation during training.

Training a model now requires several times more epochs, specially if a single sample is used to compute the loss. Using multiple samples increases computational requirements considerably and can help stabilize the loss function. Overall BBB does not scale to large models, they simply do not converge and loss explodes (goes to infinity or NaN).

One additional problem is with the implementation of BBB, since we use batches within Stochastic Gradient Descent, the weight matrix and bias are sampled once and applied to a whole batch of data. So each batch gets different samples of the weight/bias, but samples in the same batch get the same weight/bias sample, which reduces weight sample diversity. This forces to increase the number of epochs during training, and it is also a limitation due to training using GPUs, where weights in a forward/backward pass are constants that do not change along a batch.

Flipout (\sidecite{wen2018flipout}) aims to reduce the variance of the predicted distributions and training process with a mathematical trick to reduce this correlation. Flipout improves upon Bayes by Backprop.

First we notice that since BBB uses a Gaussian distribution, the sampling process can be interpreted as a perturbation over the mean:
\begin{equation}
	w = \mu + \sigma z \text{ with } z \sim \text{Norm}(0, 1), \rightarrow w \sim \text{Norm}(\mu, \sigma)
\end{equation}
Which is also known as the Reparametrization trick. Then taking $\hat{\Delta W} = \sigma z$ now being a per-sample perturbation
\begin{equation}
	\Delta W_n = \Delta \hat{W} r_n s_n^T
\end{equation}
Where $r_n$ and $s_n$ are independent samples from a Rademacher distribution\marginnote{The Rademacher distribution is binary $-1$ or $1$ values with equal probability}, and $\Delta W_n$ is the perturbation added to the kernel/bias (the standard deviation). This ensures each sample in a batch receives different kernel/bias samples, which stabilizes training.

\cite{wen2018flipout} show that when using Flipout, the kernel/bias distributions is equal to the original weight distribution, but now each sample in a batch is decorrelated with respect to kernel/bias samples used in the layer.
\subsection{Selection of Priors}

One important component specific to Bayesian NNs is the selection of prior weight distributions.

The prior weight distribution is connected to random weight initialization methods. The difference is that a prior distribution combined with Variational Inference methods for BNNs will constrain the model weights to follow the prior distribution, while in a standard Neural Network, the weight initializer only sets the starting weight values, and the model can diverge to values outside of the weight initializer range.

\subsection{Network Architecture Details}
To implement Bayesian NNs, we need to implement custom versions of common layers\marginnote{By layers we mean Dense, RNN, LSTM, Convolutional, Attention, Self-Attention, etc} replacing point-wise weights with distribution weights, usually assuming a Gaussian distribution for the weights.

When building Bayesian Neural Network Architectures, they are not required to use all layers with weight distributions. The most simple Bayesian NN is one where the output layer uses a weight distribution, and all other previous layers are standard deterministic layers.

Selecting more or less Bayesian layers in a model will change its behavior, and apply a trade-off between amount of computation and quality of uncertainty. Less layers will result in a faster model, but not good uncertainty, while using many Bayesian layers, model will be slower, but uncertainty will be high quality.

There is also a trade-off related to instability during training, even when using Flipout, building large Bayesian NNs could lead to the loss exploding during training.

\section{Comparison of UQ Methods}
We have covered many methods to estimate uncertainty at the output of a machine learning model. How do these models compare and which one works best? Of course this does not have a general answer, as the selection of uncertainty methods is task and dataset specific, but we aim to provide some basic insights.

\sidecite{valdenegro2021exploring} evaluates several uncertainty estimation methods when trained variations of a training set, from small to large training sets, by using synthetic datasets that can easily be sampled to any desired training set size. The purpose of this experiment is to test epistemic uncertainty estimation as it is directly linked to the number of samples in the training set.

\begin{figure}[ht]
	\includegraphics[width=\linewidth]{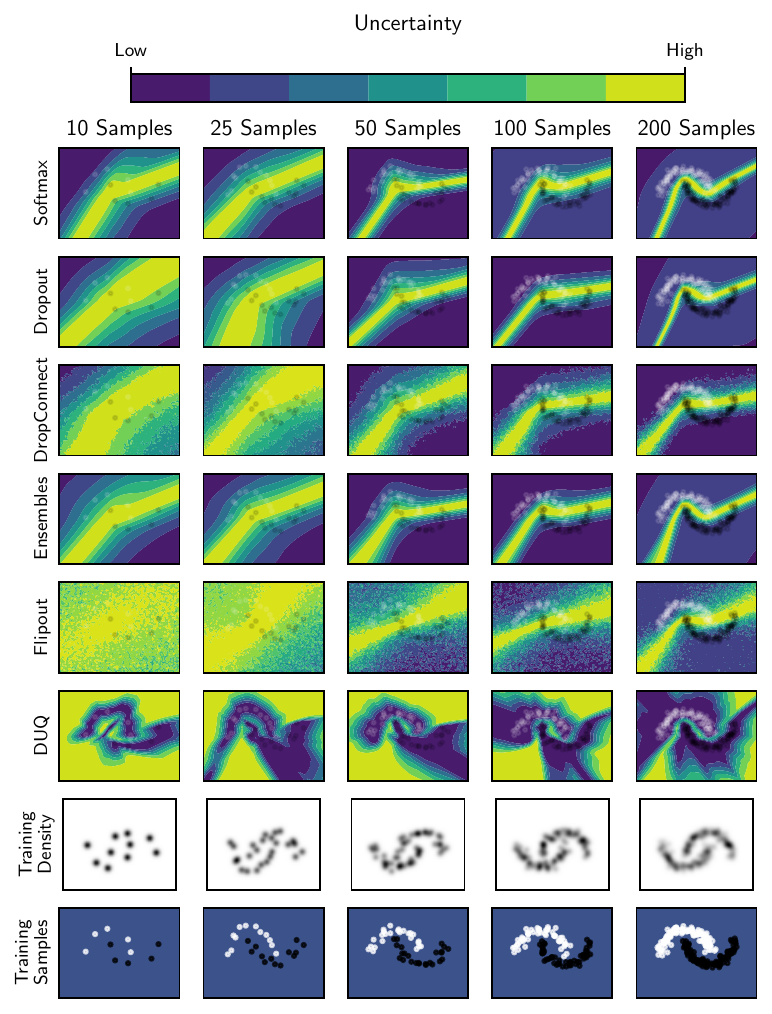}
	\caption[Comparison of classification uncertainty methods with varying training set size on the Two Moons dataset]{Comparison of classification uncertainty methods with varying training set size on the Two Moons dataset. Background color indicates uncertainty measured by Entropy. Softmax indicates a single network with softmax confidence, while Ensembles uses $M = 5$. Note how uncertainty decreases as the training set size increases.}
	\label{comp_classification_trainset}
\end{figure}

Figure \ref{comp_classification_trainset} makes a comparison in classification, using the two moons dataset\marginnote{Available at \url{https://scikit-learn.org/stable/modules/generated/sklearn.datasets.make_moons.html}}, which is a synthetic dataset of two half-moon shapes that can be generated at any training set size. There is no test set, and the aim is to show how classification models deal with the decision boundary uncertainty between the half-moons.

\begin{figure}[!htb]
	\includegraphics[width=\linewidth]{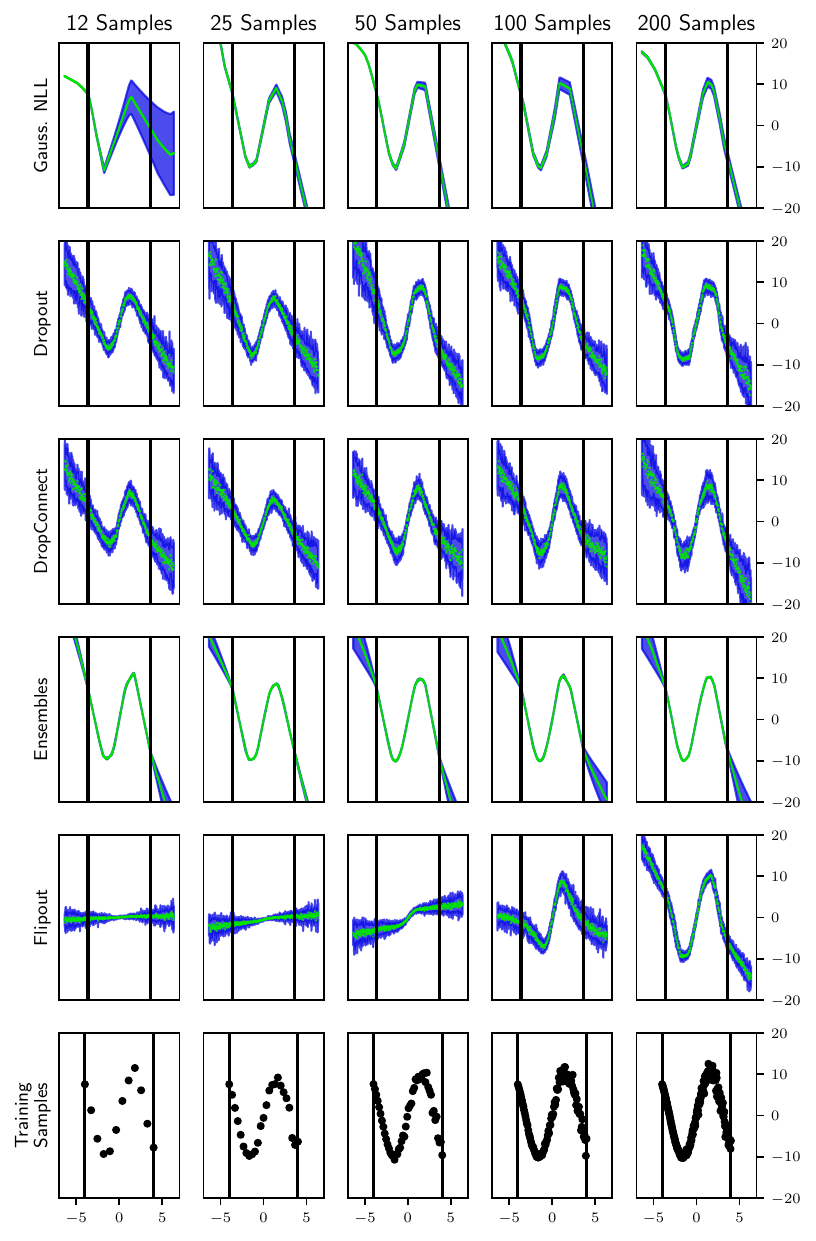}
	\caption[Comparison of regression uncertainty methods with varying training set size on a sinusoid dataset]{Comparison of regression uncertainty methods with varying training set size on a sinusoid dataset, with training set being sampled from the range $x \in [-4, 4]$, and values outside this range are out of distribution. Each plot shows {\color{green} mean} and {\color{blue} one standard deviation}. Some methods like Flipout do reduce their uncertainty with more training samples, and overall there is large variation in uncertainty across methods.}
	\label{comp_regression_trainset}
\end{figure}

Figure \ref{comp_regression_trainset} makes a similar comparison on a regression setting, varying the number of samples in the training set. A toy regression dataset for a function $f(x) = \sin(x) + \epsilon$ with $\epsilon \sim \mathcal{N}(0, \sigma(x))$ and $\sigma(x) = 0.15(1 + e^{-x})^{-1}$, producing a setting with Heteroscedastic data uncertainty.

One basic observation from both comparisons is that not all uncertainty estimation methods perform the same, there are significant differences in their predictions with uncertainty on both classification and regression. This is because different UQ methods make different assumptions on the way they measure uncertainty and their output distributions, and these adapt more or less to a given dataset.

The expectation of testing UQ methods across different training set sizes is to evaluate model uncertainty, and in most cases we see that smaller training sets do have higher uncertainty, but this does not always happen, for example MC-Dropout and MC-DropConnect seem to have the same uncertainty in the regression case (Figure \ref{comp_regression_trainset}), while it does reduce in the classification case.

Overall the best performing methods is Ensembles\marginnote{Ensembles is a very strong baseline}, with a clear reduction in uncertainty with larger training sets, which is also the case for Flipout, but the latter is much more uncertain in the small training set size in both cases. Overall an important observation is that uncertainty does depend both on the model and uncertainty estimation method being used. There is no ground truth value for model uncertainty, since it does depend on the model and its assumptions and uncertainty estimation formulation.

A disadvantage of Ensembles is the increased computational requirement\marginnote{Computational complexity of UQ}, as multiple models have to be trained but also used at inference time. Sampling methods and Bayesian NNs also require more computation as each sample corresponds to one forward pass, and many samples are required to accurately estimate the predictive posterior distribution. The only methods presented in this book that can do uncertainty estimation with a single neural network are DUQ and DDU.

\begin{table*}[!htb]
	\begin{tabular}{rllp{5.5cm}p{2.5cm}p{2cm}}
		\toprule
		& \textbf{R?} &\textbf{C?} & \textbf{Parameters} & \textbf{Values} &\textbf{Impl. Diff}\\
		\midrule
		MC-Dropout      & \cmark & \cmark  & Drop probability $p$ & $0.1 \leq p \leq 0.5$ & Easy\\
		MC-DropConnect  & \cmark & \cmark & Drop probability $p$ & $0.05 \leq p \leq 0.2$ & Medium\\
		Ensembles 		& \cmark & \cmark & Number of Ensemble Members $M$ & $M = 5$ & Easy\\
		Direct UQ 		& \xmark & \cmark &  Centroid dimensions $n$, length scale $\sigma$, gradient penalty strength $\lambda$ & $\sigma \in [0.05, 0.5]$, $\lambda$ is tunable, $n$ similar to feature dims & Medium\\
		\midrule 
		Bayes by Backprop   & \cmark & \cmark & Prior distribution parameters & N/A & Hard\\
		Flipout 			& \cmark & \cmark & Prior distribution parameters & N/A & Hard\\        
		\bottomrule
	\end{tabular}
	\caption[Summary of parameters and their typical values for different uncertainty estimation methods]{Summary of parameters and their typical values for different uncertainty estimation methods, their applicability for \textbf{C}lassification and \textbf{R}egression tasks, and implementation difficulty between Easy, Medium, and Hard.}
	\label{comp_parameters}
\end{table*}

We end this section with a conceptual comparison of uncertainty methods. Table \ref{comp_parameters} summarizes the UQ methods we introduced in this chapter, noting if they are applicable for classification or regression tasks, and sharing information about what parameters are required to tune and their common values. We also briefly describe the difficulty in implementing these methods, depending if they can be implemented with off-the-shelf framework components (Easy), if they require simple custom layers (Medium), or if they require complex custom layers (Hard) which in this case is only necessary for Bayesian NNs.

\section{Uncertainty Disentanglement}
\label{sec:disentangle}

Uncertainty disentanglement is to produce separate estimates of data and model uncertainty, that is, to disentangle predictive uncertainty into its data and model uncertainty components. That is, in broad terms:
\begin{equation}
	\text{Predictive Uncertainty} = \text{Model Uncertainty} + \text{Data Uncertainty}
\end{equation}
There are two different but very similar formulations to disentangle uncertainty, depending on the task of regression or classification. We cover the regression formulation first, from where the classification formulation derives in a clearer way.
\begin{figure}[ht]
	\includegraphics[width=\linewidth]{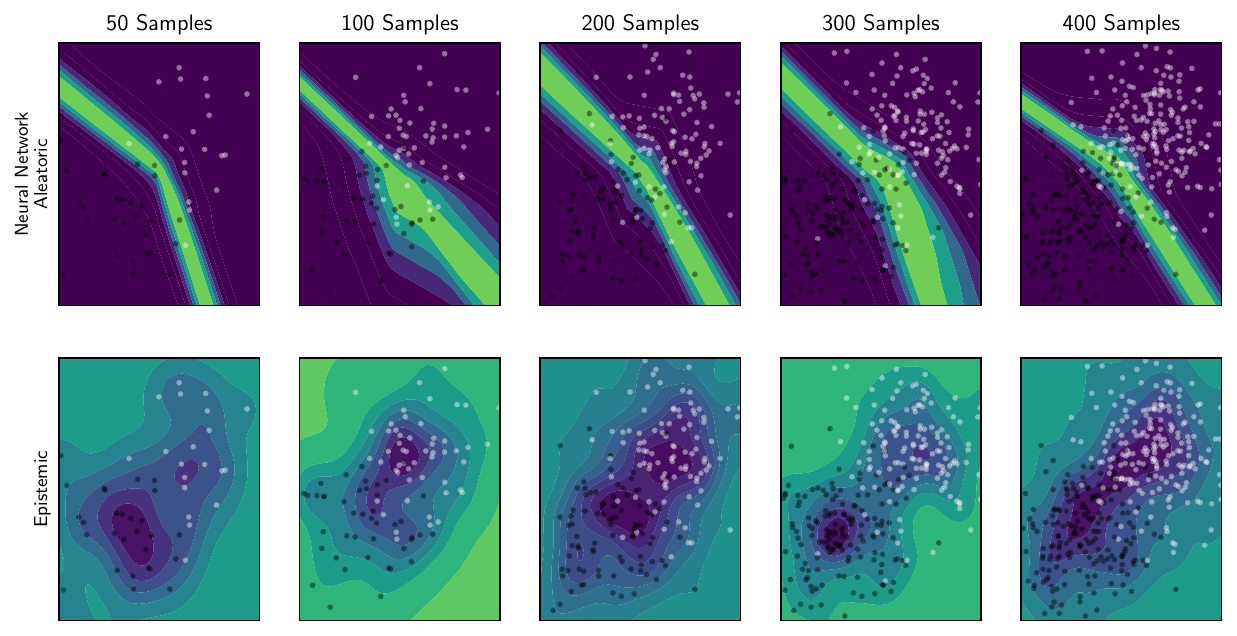}
	\caption[Conceptual example of uncertainty disentanglement on a classification task with a varying training set size.]{Conceptual example of uncertainty disentanglement on a classification task with a varying training set size. Data uncertainty is estimated with the cross-entropy loss and softmax output, while model uncertainty is estimated as Density estimation of intermediate model features. Data uncertainty is more or less constant, while model uncertainty decreases as more data points are included in the training set.\newline
    Low
    \begin{tikzpicture}
        \protect
        \node [shading = axis,rectangle, left color=blue, right color=green,shading angle=90, anchor=north, minimum width=0.6\linewidth, minimum height=0.2cm] (box) at (current page.north){};
    \end{tikzpicture} High
    }
\end{figure}
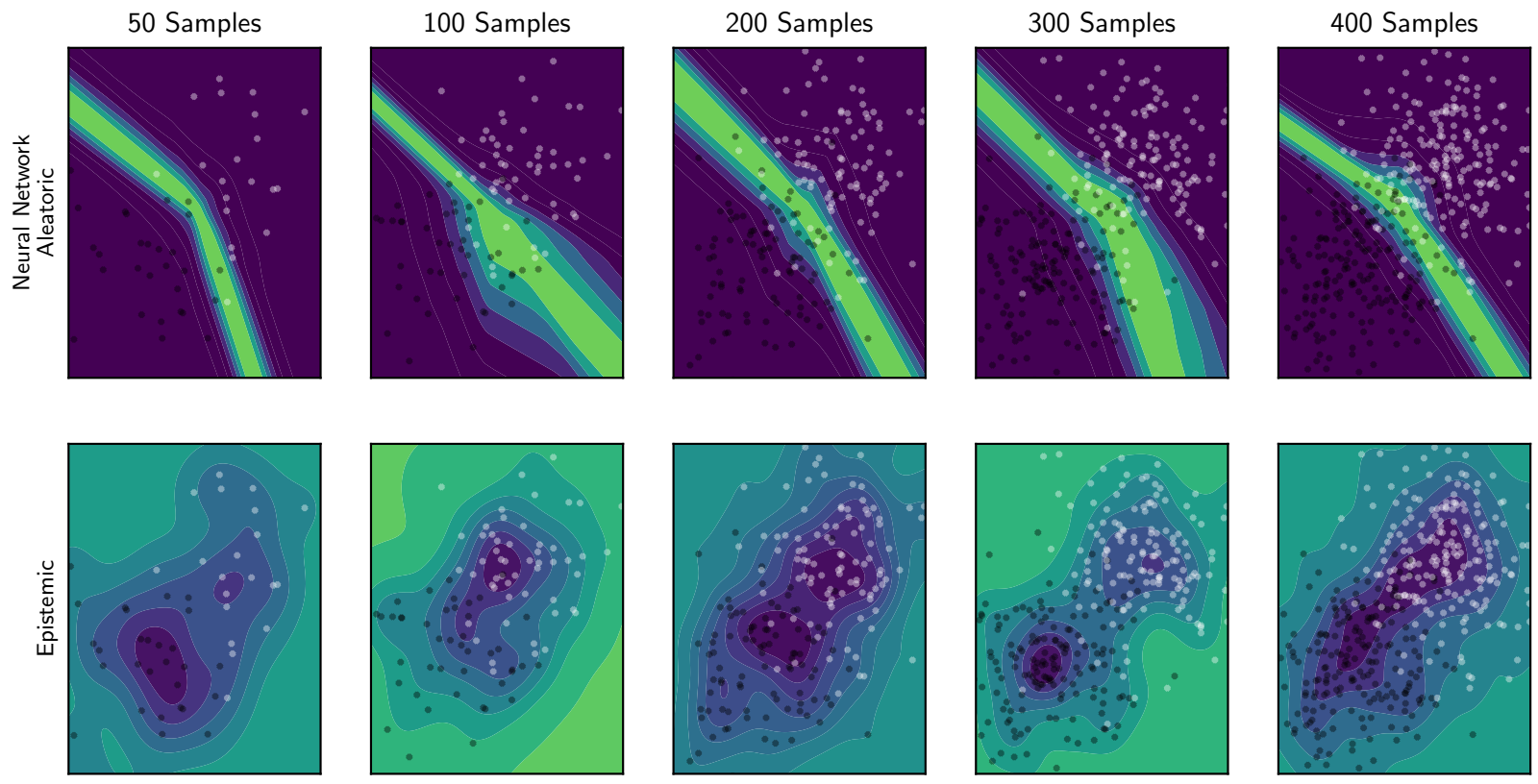

\subsection{Regression}
To perform uncertainty disentanglement in Regression, we need the following basic setup:
\begin{description}
	\item[Data Uncertainty] We can estimate Aleatoric Uncertainty using the Gaussian Negative Log-Likelihood and a Two-Head Model.
	\item[Model Uncertainty] Epistemic Uncertainty is produced by using a UQ method like MC-Dropout/DropConnect, Ensembles, Bayes by Backprop, Flipout, etc. This method should produce samples or use ensembling.
\end{description}
Consider the case of an ensemble for regression of a Gaussian distribution for each ensemble member. Equation \ref{eq:ensemble_reg_combination} computes the Gaussian mixture variance, which can also be applied with sampling methods. We take a deeper look at the variance $\sigma^2_*$, which can be decomposed into two important terms:

\begin{definitionbox}{Gaussian Mixture Uncertainty Disentangle}{gaussian_reg_disentangle}
	\begin{tabular}{ccccc}
		\addtolength{\tabcolsep}{-4pt}
		$\sigma^2_*(\mathbf{x})$ & $=$ & $\displaystyle M^{-1} \sum_i \sigma^2_i(\mathbf{x})$ & $+$ & $\displaystyle M^{-1} \sum_i \mu^2_i(\mathbf{x})- \mu^2_*(\mathbf{x})$\\[1ex]
		& $=$ &  $\mathbb{E}_i [\sigma_i^2(\mathbf{x})]$ & $+$ & $\mathbb{E}_i [\mu_i^2(\mathbf{x})] - \mathbb{E}_i [\mu_i(\mathbf{x})]^2$\\[1ex]
		& $=$ & $\underbrace{\mathbb{E}_i [\sigma_i^2(\mathbf{x})]}_{\text{Data Uncertainty}}$ & $+$ & $\underbrace{\text{Var}_i[\mu_i(\mathbf{x})]}_{\text{Model Uncertainty}}$
	\end{tabular}
	
	This is easy to interpret, Data uncertainty is the mean of the per-sample/ensemble variances, and Model uncertainty is the variance of the per-sample/ensemble means.
\end{definitionbox}

In this short derivation we use two properties, first is that the expectation is linear, and second is an alternate computation of the variance as the expectation of the square minus the square of the mean.

The intuition of each terms are given below:
\begin{description}
	\item[Data Uncertainty] Aleatoric Uncertainty is given by the mean of per-sample/ensemble variances, which can be interpreted as combining all variances into a single variance.
	\item[Model Uncertainty] Epistemic Uncertainty is given by the variance of per-sample/ensemble means. This can be interpreted as the variation or \emph{disagreement} between models/samples, which is by definition epistemic uncertainty.\\
	
	If all means are the same, the model(s) is very confident. If all means are different and varied, the model(s) disagree and this signals epistemic uncertainty.
\end{description}

While we have made this disentanglement formulation using ensembles, the exact same formulation can be used with sampling-based methods, where we consider each sample to be a different ensemble member.

Note that this disentanglement formulation assumes that each sample/ensemble predicts a Gaussian distribution, forming a Gaussian mixture but combined into a single Gaussian using Eq \ref{eq:ensemble_reg_combination}, from where the predicted variance can be disentangled into model and data uncertainty.

\subsection{Classification}
To perform uncertainty disentanglement in Regression, we need the following basic setup:
\begin{marginfigure}
	\scalebox{0.8}{
		\begin{tikzpicture}[style={align=center, minimum height=0.5cm, minimum width = 1.8cm}]
			\node[] (dummy) {};
			\node[rounded corners, draw, above=2.0em of dummy, align=center] (sampSoftmax) {Sampling\\Softmax};
			\node[above=1.0em of sampSoftmax] (probs) {{Probabilities}};
			
			\node[rounded corners, draw, left=0.1em of dummy, align=center] (objClassifier) {Logit\\Mean $\mu(\mathbf{x})$};			
			\node[rounded corners, draw, right=0.1em of dummy, align=center] (objDetector) {Logit\\Var. $\sigma^2(\mathbf{x})$};
			
			\node[below=2.0em of dummy] (fc) {{Feature Vector}};
			
			\draw[-latex] (fc) -- (objDetector);
			\draw[-latex] (fc) -- (objClassifier);
			\draw[-latex] (objClassifier) -- (sampSoftmax);
			\draw[-latex] (objDetector) -- (sampSoftmax);
			\draw[-latex] (sampSoftmax) -- (probs);
		\end{tikzpicture}
	}
	\caption[Computational graph for the sampling softmax function]{Computational graph for the sampling softmax function, showing the two fully connected layers that produce logit mean and variance, and how they relate to the final probabilities through the sampling softmax function.}
\end{marginfigure}
\begin{description}
    \item[Data Uncertainty] We can estimate Aleatoric Uncertainty using the standard cross-entropy but the output logits of the model should have uncertainty, and passed through the Sampling Softmax function, which takes Gaussian distributed logits and passed this distribution through the softmax function.
    \item[Model Uncertainty] Epistemic Uncertainty is produced by using a UQ method like MC-Dropout/DropConnect, Ensembles, Bayes by Backprop, Flipout, etc. This method should produce samples or use ensembling.
\end{description}
This setup is different than the one we use in Regression, because there is no direct way to perform uncertainty disentanglement in Classification due to the probabilities that models output, then we need to take a deviation by using logits with uncertainty, which is more similar to how uncertainty disentanglement works in Regression.
\begin{definitionbox}{Sampling Softmax}{sampling_softmax}
	How to pass your Gaussian distributed logits with mean $\mu(\mathbf{x})$ and variance $\sigma^2(\mathbf{x})$, through the softmax function? The only known way is to use sampling:
	\begin{align}
		\hat{\mathbf{z}}_j &\sim \mathcal{N}(\mu(\mathbf{x}), \sigma^2(\mathbf{x}))\label{eq:ssoftmax1} \\
		\condprob{\mathbf{y}}{\mathbf{x}} &= N^{-1} \sum_j \text{softmax}(\hat{\mathbf{z}}_j) \qquad j \in [1, N] \label{eq:ssoftmax2}
	\end{align}    
	Here we generate $N$ samples of the logit Gaussian distribution $\mathcal{N}(\mu(\mathbf{x}), \sigma^2(\mathbf{x}))$, which is denoted by $\hat{\mathbf{z}}_j$, apply the standard softmax activation to each sample, and then take the sample mean to obtain the output $\condprob{\mathbf{y}}{\mathbf{x}}$.\\
	
	The number of samples $N$ defines the approximation quality to the true softmax function (which we do not know).
\end{definitionbox}
Then the disentanglement assuming Gaussian distributed logits is made with the following formulation:
\begin{definitionbox}{Gaussian Logits Disentanglement}{gau_logits_disentangle}
	Then at inference time, we again assume that the uncertainty method uses sampling through forwarding passes or ensembling a model on the $i$ axis (with $i \in [1, M]$). Then, data $\sigma^2_{\text{Ale}}$ and model $\sigma^2_{\text{Epi}}$ uncertainty logits can be computed as:
	\begin{align}
		\sigma^2_{\text{Data}}(\mathbf{x}) &= \mathbb{E}_i [\sigma_i^2(\mathbf{x})]\\
		\sigma^2_{\text{Model}}(\mathbf{x}) &= \text{Var}_i [\mu_i(\mathbf{x})].
	\end{align} 
	Note that these are logits, not probabilities. These steps are equivalent to the disentangling for regression, where the input to the Sampling softmax function is a kind of regression (with uncertainty) of the logits.
	
	Passing each corresponding logit through a softmax function can produce probabilities, from where entropy is a possible metric to obtain a scalar uncertainty measure:
	\begin{align*}
		p_{\text{Ale}}(y | \mathbf{x}) &= \text{sampling\_softmax}(\mu(\mathbf{x}), \sigma^2_{\text{Ale}}(\mathbf{x}))\\
		H_{\text{Ale}}(y | \mathbf{x}) &= \text{entropy}(p_{Ale}(y | \mathbf{x}))\\    
		p_{\text{Epi}}(y | \mathbf{x}) &= \text{sampling\_softmax}(\mu(\mathbf{x}), \sigma^2_{\text{Epi}}(\mathbf{x}))\\
		H_{\text{Epi}}(y | \mathbf{x}) &= \text{entropy}(p_{Epi}(y | \mathbf{x}))
	\end{align*}
\end{definitionbox}

Another formulation for uncertainty disentanglement in Classification is based on Information Theory, in particular on the Mutual Information $\mathbb{I}[X; Y]$ which measures the mutual dependence between two random variables $X$ and $X$, and can be defined with respect to the entropy $\mathbb{H}$ and conditional entropy\marginnote{The conditional entropy is defined as $\mathbb{H}(X \, |\, Y) = \int_X \int_Y p(x, y) \log \frac{p(x, y)}{p(x)}dxdy$} $\mathbb{H}(\cdot \, | \cdot )$ as:
\begin{equation}
	{I}[X; Y] = \mathbb{H}(X) - \mathbb{H}(X \, |\, Y) = \mathbb{H}(Y) - \mathbb{H}(Y \, |\, X)
\end{equation}
When the two variables $X$ and $Y$ are independent, then the mutual information is zero, meaning that knowing one variable does not give information about the other variable.

Then for the purposes of uncertainty disentanglement, we would like to study the relationship between the output variable $Y$ and the weights $\theta$, which is the mutual information $\mathbb{I}[Y; \theta]$, this term can be approximated using the prediction entropy between two terms:
\begin{definitionbox}{Information Theoretic Disentanglement}{inf_theory_disentangle}
	Considering an uncertainty estimation with samples/ensembles, the Entropy of the mean probability $\expectover{\mathbb{H}[\condprob{y}{x, \theta}]}{\theta}$ is considered to represent the total uncertainty, and the Expected Entropy $\mathbb{H}[\expectover{\condprob{y}{x, \theta}}{\theta}]$ is considered to represent data uncertainty.\\
	
	Both of these terms take expectations over the weights $\expectover{\cdot}{\theta}$ which in both cases can be approximated by taking the expectation over ensemble members or samples from multiple forward passes. Then these terms relate to the mutual entropy $\mathbb{I}[Y; \theta]$ as:	
	\begin{equation}
		\underbrace{\mathbb{I}[Y; \theta]}_\text{Model Uncertainty}  \approx \underbrace{\mathbb{H}[\expectover{\condprob{y}{x, \theta}}{\theta}]}_\text{Total Uncertainty} - \underbrace{\expectover{\mathbb{H}[\condprob{y}{x, \theta}]}{\theta}}_\text{Data Uncertainty}
	\end{equation}
	
	Note that this is an approximation (denoted by $\approx$), and in practice it is treated as an equality, to compute model uncertainty with the mutual information as the difference between Entropy of mean probability and Expected Entropy.
\end{definitionbox}

\begin{table}[!b]
    \begin{tabular}{rlllll}
        \toprule
            			& \textbf{Data U} & \textbf{Model U} & \textbf{D+M U?} & \textbf{Disentangle?}\\
        \midrule
        \textbf{UQ Methods} \\
        MC-Dropout      & \xmark & \cmark  & \cmark & \cmark \\
        MC-DropCon.  & \xmark & \cmark  & \cmark & \cmark \\
        Ensembles 		& \xmark & \cmark  & \cmark & \cmark \\
        Direct UQ 		& \xmark & \cmark  & \xmark & \xmark \\
        \midrule
        \textbf{Bayesian NNs} \\
        BBBackprop  	& \xmark & \cmark & \cmark & \cmark\\
        Flipout 			& \xmark & \cmark & \cmark & \cmark\\
        \midrule
        Samp. Softmax & \cmark & \xmark & \cmark & \cmark\\
        Samp. Sigmoid & \cmark & \xmark & \cmark & \cmark\\
        \midrule
        \textbf{UQ Losses} \\
        Gaussian NLL 	& \cmark & \xmark & \cmark & \cmark \\
        Laplace NLL 	& \cmark & \xmark & \cmark & \cmark \\
        \bottomrule
    \end{tabular}
    \caption{Summary of uncertainty quantification methods and their capabilities relative to estimating data and model uncertainty, and if they allow for a disentanglement formulation.}
    \label{comp_uncert_disentangle}
\end{table}

Table \ref{comp_uncert_disentangle} summarizes methods relating to its ability to capture data or model uncertainty, if they can be combined to produce predictive uncertainty, and if there is a disentanglement formulation for them.

\begin{figure}[!tb]
	\includegraphics[width=\linewidth]{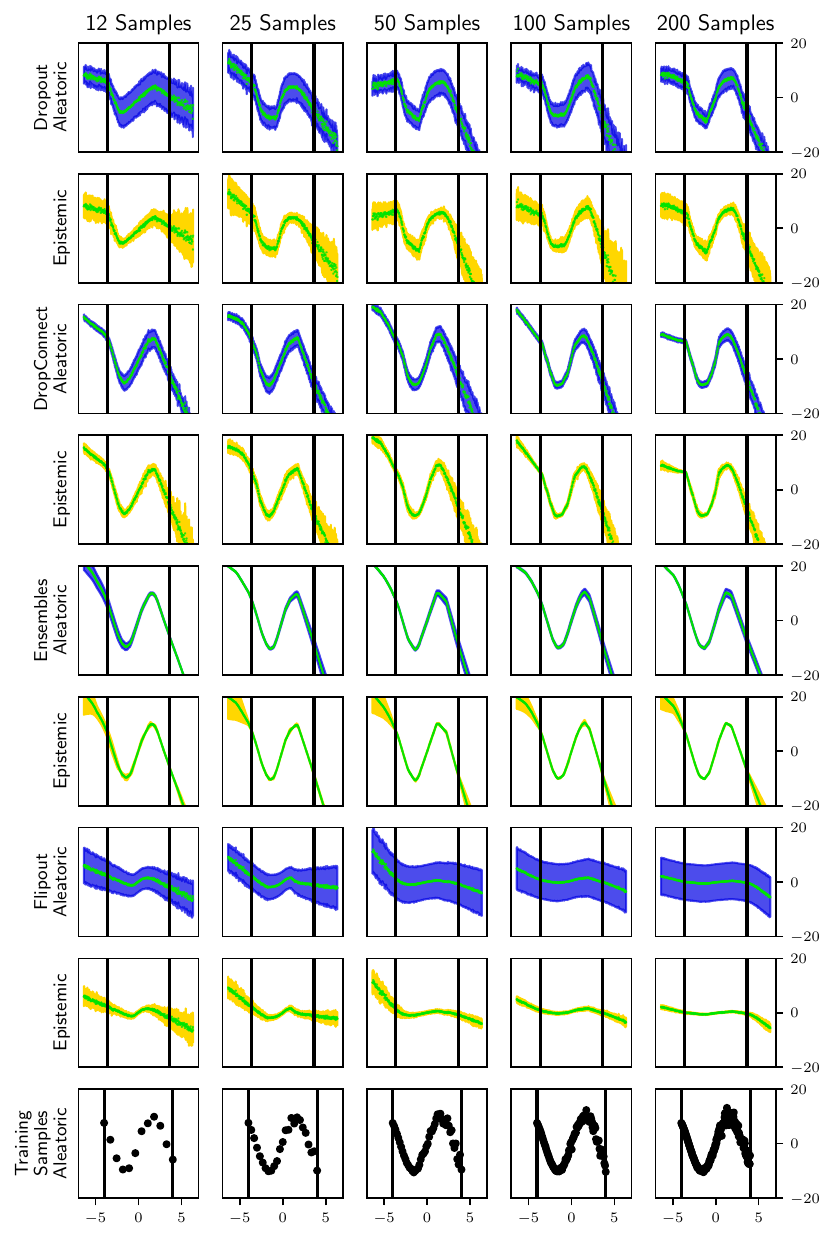}
	\caption[Comparison of regression uncertainty disentanglement methods with varying training set size on a sinusoid dataset]{Comparison of regression uncertainty disentanglement methods with varying training set size on a sinusoid dataset, with training set being sampled from the range $x \in [-4, 4]$, and values outside this range are out of distribution. Each plot shows {\color{green} mean}, {\color{blue} data uncertainty std}, and {\color{gold} model uncertainty std}. Some methods do show that data uncertainty is constant across training set sizes, like Ensembles, while model uncertainty does reduce with larger training set sizes.}
	\label{comp_regression_disentangle_trainset}
\end{figure}

Figure \ref{comp_regression_disentangle_trainset} makes a similar comparison using the regression disentanglement formulation over a simple sinusoid. Note how in most methods, data uncertainty is more or less constant (after training with enough samples, usually from 50 samples), while model uncertainty does change and more or less reduce with more training data. Both types of uncertainty can be unstable with small training sets.

These results do show that disentangling data and model uncertainty is definitely not a trivial problem, specially in the classification setting, and some uncertainty quantification methods like Dropout have both high data and model uncertainty. There is a common problem of data uncertainty underestimation that definitely is shown in these results.

Finally, it should be noted that data uncertainty\marginnote{Data uncertainty invalid in OOD settting} is only valid inside the training set, since outside the training set model uncertainty will increase, and there is no expectation of generalization for data uncertainty outside the training set, since there is no training data by definition, and this uncertainty is learned from data itself.

%% file: sections/evaluation.tex
\chapter{Evaluation of Uncertainty}
\label{ch:evaluation}
\margintoc

\section{Introduction}

So far we have covered many ways to build models with uncertainty estimation, but there is an important open question. Since our models now output probability distributions, we cannot use standard metrics for evaluation, we need new kinds of metrics to evaluate our probability distribution predictions, specially to measure quality of uncertainty.

In this chapter we will cover metrics, loss functions, and evaluation methodologies for uncertainty estimates.

\section{Metrics and Losses for Uncertainty}
As uncertainty quantification makes additional prediction components, like a mean $\mu$ and standard deviation $\sigma$ for regression, or per-class probabilities $\condprob{y = c}{x}$ for classification, this additional information means that standard loss functions and metrics cannot be used, and new metrics that consider prediction uncertainty have to be considered.

This section enumerates several popular loss functions and metrics that explicitly evaluate predictions with uncertainty.
\subsection{Distribution Predictions}
The most general case is predicting a probability distribution, and there are loss functions and metrics for this kind of predictions, the most common one being the KL-divergence.
\begin{definitionbox}{Kullback-Leibler Divergence}{kl_divergence}
    The KL-divergence is a distance measure between probability distributions $P$ and $Q$, defined for continuous distributions as:
    \begin{equation}
        L(p, q) = \int_{-\infty}^{\infty} p(x) \log \left( \frac{p(x)}{q(x)} \right) dx
    \end{equation}
    With $p$ and $q$ being the corresponding probability density functions. And for discrete distributions:
    \begin{equation}
        L(y, \hat{y}) = \sum_i y_i \log \left( \frac{y_i}{\hat{y}_i} \right)
    \end{equation}
\end{definitionbox}

\subsection{Classification}
\begin{definitionbox}{Cross-Entropy Loss}{cross_entropy}
    The standard cross-entropy loss function does consider prediction uncertainty for multi-class classification, with prediction probabilities $\hat{y}^c_i = \condprob{y = c}{x}$ being a vector, and $c$ being the class index and $i$ the dataset element index.
    \begin{equation}
        L(\mathbf{y}, \hat{\mathbf{y}}) =  -\sum_i \sum_c y^c_i \log(\hat{y}^c_i)
    \end{equation}
    Ground truth labels $y^c_i$ are probability vectors, each element being the probability of that class. Usually integer labels are one-hot encoded to produce a $(0/1)$ probability vector, or they can be full probability vectors with $\sum_c y^c = 1$.\\
    
    A simplification of the cross-entropy loss is available for binary classification problems, where $\hat{y}_i = \condprob{y = 1}{x}$ is the positive class probability and $i$ the dataset element index.
    \begin{equation}
        L(y, \hat{y}) =  -\sum_i y_i \log(\hat{y}_i) + (1 - y_i) \log(1 - \hat{y}_i)
    \end{equation}
\end{definitionbox}

\begin{definitionbox}{Entropy}{entropy}
    Entropy is a metric that measures the information content in a probability distribution. For classification, we use a discrete probability distribution that is conditioned by each class $\hat{y}^c = \condprob{x}{c}$. Entropy is defined as:
    \begin{equation}
        \mathbb{H}(\hat{y}) = -\sum_c \hat{y}^c \log \hat{y}^c
    \end{equation}
    Where $c \in [0, C - 1]$ are class indices. Entropy only operates on a prediction $\hat{y}$, and does not need any ground truth labels. It is useful to quantify uncertainty in a classification prediction, as entropy maximizes when the predicted distribution is uniform, and it is zero for a peaky distribution, indicating certainty for a particular class.    
\end{definitionbox}

\begin{marginfigure}
    \begin{tikzpicture}        
        \begin{axis}[width=\linewidth, domain=0.0000000001:1, xlabel={$\condprob{y = 1}{x}$}, ylabel={Entropy (Bits)}]
            \addplot+[mark=none, smooth] {-x * ln(x) / ln(2) - (1-x) * ln(1-x) / ln(2)};
        \end{axis}
    \end{tikzpicture}
    \caption{Plot of entropy for a binary classification example.}
    \label{fig:entropy_binary}
\end{marginfigure}

The entropy is important as it is directly related to uncertainty. The units of entropy are called bits if one uses the base-two logarithm. Figure \ref{fig:entropy_binary} displays how entropy behaves for the binary classification case, clearly showing how maximum entropy is achieved when the predicted probability is $0.5$, exactly at the borderline between both classes.

Some important properties for entropy: The Uniform distribution is the one that maximizes entropy, as its results are harder to predict. For a fixed mean and variance, the Gaussian distribution is the one that maximizes the entropy.

\begin{definitionbox}{Brier Score}{brier_score}
    The brier score is the mean squared error but applied to output probabilities $\hat{y}^c = \condprob{c}{x}$ and compared to ground truth probabilities $y_c$:
    \begin{equation}
        \text{Brier}(y, \hat{y}) = C^{-1} \sum_c (y^c - \hat{y}^c)^2
    \end{equation}
    With $c$ being class indices as mentioned before.
\end{definitionbox}

The brier score measures how close are the predicted and true probabilities, in an interpretable and intuitive way.

Classification losses and metrics are mostly based on per-class probabilities. In particular Entropy and Brier Score are used as way to summarize the predicted distribution into a single number, with Entropy not requiring labels, and the Brier Score requiring true probability labels.

\subsection{Regression}
\begin{definitionbox}{Gaussian Negative Log-Likelihood Loss}{gaussian_nll}
    The standard mean squared error does not consider prediction uncertainty, as it only supervises the prediction mean. An alternative is the Gaussian NLL that we have previously derived, which does consider both prediction mean and standard deviation.
    \begin{equation}
        L(y, \hat{\mu}, \hat{\sigma}^2) = \sum_i \left[\log \hat{\sigma}^2_i + \left(\frac{\hat{\mu}_i - y}{\hat{\sigma}_i}\right)^2 \right]
    \end{equation}
    Where $\hat{\mu}$ is the prediction mean, $\hat{\sigma}$ is the prediction standard deviation, and $y$ is the (continuous) ground truth label, which supervises the mean. The Gaussian NLL can also be used as a metric to evaluate regression predictions with uncertainty under the Gaussian assumption.
\end{definitionbox}

Accuracy is a common metric used in classification, but it does not work with regression tasks due to only equality between prediction and true label indicating correctness, but the Coverage metric extends this concept to regression tasks with uncertainty, in particular when predicting intervals.

\begin{definitionbox}{Coverage}{def:coverage}
    Given a set of intervals $[\hat{y}^l, \hat{y}^h_i]_i$ (with upper and lower bounds), then coverage is defined as:    
    \begin{equation}
        \text{Cov}(y, \hat{y}) = N^{-1} \sum \mathrm{1}[\hat{y}^l_i \leq y \leq \hat{y}^h_i]
    \end{equation}
    Where $\hat{y}^l_i$ is the i-th predicted lower bound, and $\hat{y}^h_i$ is the corresponding upper bound, of the interval.
\end{definitionbox}

Coverage is conceptually similar to accuracy (for classification), as an interval can contain the correct label $y$ or not, producing a 0-1 score than is averaged across a set of predictions, which can be interpreted as an interval accuracy.

Entropy can also be extended to continuous settings, this is called differential entropy.

\begin{definitionbox}{Differential Entropy}{def:diff_entropy}
    Given a continuous random variable $X$ with domain $\Omega$ and PDF $f(x)$, then differential entropy is computed as:
    \begin{equation}
        \mathbb{H}(X) = -\int_\Omega f(x) \log f(x) dx
    \end{equation}    
\end{definitionbox}

For example the entropy of a Gaussian distribution is $\frac{1}{2} \log(2 \pi \sigma^2)+ \frac{1}{2}$. As you can see, the entropy only depends on the variance $\sigma^2$ of the distribution, which shows the relationship between uncertainty and (differential) entropy.

Table \ref{table:metrics_losses_unc_summary} makes a summary of the metrics we presented in this section and which uncertainty representation they require.

\begin{table*}[!b]
    \begin{tabular}{p{5cm}lp{7cm}}
        \toprule
         					& \textbf{Loss or Metric} & \textbf{Uncertainty\newline Representation}\\
        \midrule
        \textbf{Probability Distribution} \\
        KL-divergence		& L & Probability density function.\\
        \midrule
        \textbf{Classification} \\
        Categorical Cross-Entropy & L & Probability vector.\\
        Binary Cross-Entropy & L & Probability of positive class.\\
        Entropy 		& M & Probability vector.\\
        Brier Score 	& L & Probability vector.\\
        \midrule
        \textbf{Regression} \\
        Gaussian NLL 	& L & Gaussian mean and std.\\
        Coverage 		& M & Confidence interval.\\
        Differential Entropy & M & Probability density function.\\
        \bottomrule
    \end{tabular}
    \caption[Summary of metrics and losses to evaluate predictions with uncertainty and which uncertainty representation they require]{Summary of metrics and losses to evaluate predictions with uncertainty and which uncertainty representation they require. The second column indicates if it is a (M)etric or a (L)oss Function. Note that losses can always be used as metrics, but the opposite is not true.}
    \label{table:metrics_losses_unc_summary}
\end{table*}

\section{Proper Scoring Rules}

An important question in machine learning is how to select loss functions and metrics for evaluation. Not every loss or metric for a task measures the same. Is there some theory to discriminate between good and bad metrics? This is the concept of a proper scoring rule.

This concept comes from the statistics literature, and is about designing loss functions and metrics to behave properly, under some definition of proper. In simple words, it basically means that some loss functions and metrics are properly designed to evaluate probability distribution outputs, and others are not and can provide misleading conclusions.

First we define what a scoring rule is:

\begin{definitionbox}{Scoring Rule}{scoring_rule}
    A scoring rule is a function $S(\hat{p}, (y, x))$ that evaluates the quality (higher is better) of a predicted probability distribution $\condprobvar{\hat{p}}{y}{x}$ relative to an event $y | x \sim \condprobvar{q}{y}{x}$, with $\condprobvar{q}{y}{x}$ being the true distribution that generates $(y, x)$ values.\\
    
    The expected score for predicted $\hat{p}$ and true $q$ probability distributions can be built from the previous formulation by taking the integral over samples from $q$.
    \begin{equation}
        S(\hat{p}, q) = \expectover{S(\hat{p}, y)}{y \sim q} = \int_{\Omega_q} q(y, x) S(P, (y, x)) dydx
    \end{equation}
    
\end{definitionbox}

This definition is a bit complex, but easy to understand. Scoring rules evaluate (or score) probability distribution based on a predicted distribution $\hat{q}$ and a true distribution $q$. The first part in Definition \ref{def:scoring_rule} defines a scoring rule based on samples, which is an abstract definition. In statistics and machine learning we use expected (or average) scores, and the second part in Definition \ref{def:scoring_rule} is the expected scoring rule, which is what we will use in the coming pages. In machine learning we often use expected scoring rules implicitly, for example when we average a metric or loss function over a dataset.

\begin{definitionbox}{Proper Scoring Rule}{prop_scoring_rule}
    A proper scoring rule is one where $\forall \hat{p}, q$:
    \begin{equation}
        S(\hat{p}, q) \leq S(q, q)
    \end{equation}
    With equality only holding if $\condprobvar{\hat{p}}{y}{x} = \condprobvar{q}{y}{x} \quad \forall \hat{p}, q$.    
\end{definitionbox}
\marginnote{Proper Scoring Rule}

This is easy to interpret, a proper scoring rule obtains its best value \textbf{only} by predicting the true distribution $q$. Predicting any other distribution $\hat{p}$ further away from the true distribution $q$ leads to a worse score value.

Scoring rules and functions can be negatively oriented if smaller value indicates a better prediction, or positively oriented if bigger values indicate a better prediction. Definition \ref{def:prop_scoring_rule} uses a positively oriented formulation, and the negatively oriented one can be obtained by flipping the inequality.

Any scoring rule can be made into a loss function, by taking $L(\theta) = -S(p_\theta, q)$. This means that any insights we make on scoring rules also apply to the corresponding loss function, but flipping the direction of being best.
Most metrics and loss functions are proper, first we show some examples of them.
\begin{examplebox}{Common Proper Scoring Rules}{proper}
    \begin{description}[\compact\setlabelphantom{Brier Score}]
    	\item[Log-likelihoods] Log-likelihoods based on the logarithm of a joint probability distribution are proper scoring rules, such as the Gaussian NLL and the cross-entropy losses.
    	\item[KL-Divergence] The Kullback-Leibler divergence is also a proper scoring rule, as it is specifically only zero when $p = q$, and positive values if not.
    	\item[Brier Score] The brier score can only be zero if a model predicts the true probabilities, which makes it a proper scoring rule.
    \end{description}
\end{examplebox}
And in particular the Accuracy metric and mean squared error loss are not proper scoring rules.
\begin{examplebox}{Accuracy and Mean Squared Error are not Proper}{acc_improper}
    There are two simple examples of metrics that are not proper scoring rules: Accuracy and Squared Errors. Both suffer from the same problem, they do not evaluate distribution outputs, only the class with the highest probability for accuracy, and the mean prediction for the squared error.\\
    
    For accuracy, this implies that probabilities far from the class boundary ($1/C$ for $C$ classes) do not matter and do not change the accuracy value, and it is also possible to score 100\% accuracy by not predicting the true probability distribution, as only the most likely class is considered.\\
    
    For the mean squared error, it only evaluates the mean and not the standard deviation of the predicted distribution.
\end{examplebox}

Additionally, metrics that are built using the geometry mean are often not proper.

\begin{examplebox}{Metrics using Geometric Means are not Proper}{acc_improper}
    Some metrics are build based on geometric means of several metric components $m_i(y, \hat{y})$, often representing individual prediction qualities, that are aggregated via the geometric mean:
    \begin{equation}
        M(y, \hat{y}) = \sqrt[^p]{\prod_i m_i(y, \hat{y})}
    \end{equation}
    The problem with metric components, is that if any component $m_i(y, \hat{y})$ becomes zero, the whole metric value is zero, wrong signalling that the best metric value has been reached. In terms of proper scoring rules, the best value is not reached by predicting the true distribution, thus metrics using geometric means are often not proper scoring rules.
\end{examplebox}

\section{Calibration}

We have described that the concept of confidence should indicate how likely the prediction is correct, but how can we measure this? How can we conceptualize and build metrics that evaluate the quality and trustworthiness of confidence.

We can set some expectations on how a confidence should behave. If we make predictions over a dataset, and group all predictions made with 10\% confidence, then we would expect these predictions should have around 10\% accuracy. If we repeat this grouping with predictions made with 30\% confidence, they should have accuracy around 30\%. We can generalize this thinking that grouping predictions with $\alpha \%$ confidence, they should have around $\alpha \%$ accuracy.

When we group predictions according to confidence, we are building estimates of the confidences predicted by the model, which is then compared to an empirical frequency of correct predictions, which is accuracy.
\begin{equation}
    \overUnderArrow[\parbox{4cm}{Probability predicted by your Model}]{}{\text{\textbf{Confidence}}} = \overUnderArrow[\parbox{4.5cm}{Observed frequency of correct predictions}]{}{\text{\textbf{Accuracy}}}
    \label{eq:conf_vs_accuracy}
\end{equation}
This definition works for both classification and regression, but there are some minor differences that we will detail later in this chapter.

The problem of calibration is that accuracy and confidence do not always match, then accuracy could be lower than confidence, or higher. This defines two classes of calibration errors, called Overconfidence as shown in Definition \ref{def:overconfidence}, where predicted confidence is higher than observed accuracy, and Underconfidence, shown in Definition \ref{def:underconfidence}, where predicted confidence is lower than observed accuracy.

When predicted confidence does not match with observed accuracy, this is called miscalibration or calibration error. There are two main types of calibration errors:

\begin{definitionbox}{Overconfidence}{overconfidence}
	Regions where overall confidence is higher than accuracy indicate an overconfident model, so confidences should be lower than they are.\\
	
	This is the worst case as a high confidence gives a false sense of security.
\end{definitionbox}

\begin{definitionbox}{Underconfidence}{underconfidence}
	Regions where overall confidence is lower than accuracy indicate that the model is underconfident, which means that the confidence should actually be higher than it is, in order to match accuracy.\\
	
	This is the second-to-worst case as low confidences are misleading but not as dangerous as false high confidences.
\end{definitionbox}

Note that over/underconfidence is a function of confidence, and it is possible for predicted confidences to have different behavior depending on the confidence value, which means a model can be both overconfident and underconfident, just for different confidence values.

\subsection{Calibration Plots}

It is possible to observe calibration errors by building a calibration plot, which shows the relationship between predicted confidence and observed accuracy in a graphical form. This plot is also called a reliability plot.

\begin{figure}[t]
    \begin{generalbox}{Classification Calibration Plot}
        \begin{algorithmic}
            \STATE Take predictions of a model with confidence.
            \STATE Defice number of bins $M = |B|$
            \STATE Define confidence bins $B_i$ with $0 \leq i \leq M$.
            \FOR{each bin $B_i$}
                \STATE $\text{conf}(B_i) \gets \text{predictions with confidence inside } B_i$
                \STATE $\text{acc}(B_i) \gets \text{accuracy of predictions inside } B_i$
                \STATE Add average $(\text{conf}(B_i), \text{acc}(B_i))$ to plot.
            \ENDFOR
            \STATE Display plot.        
        \end{algorithmic}
    \end{generalbox}
    \caption{Algorithm to build a classification calibration plot. The number of bins $M$ has to be manually tuned.}
    \label{eval:calib_plot_algorithm}
\end{figure}    

The algorithm to build a calibration plot \marginnote{Classification Calibration Plot} for a classification setting is shown in Figure \ref{eval:calib_plot_algorithm}. This uses binning as a way to group predictions by confidence. Each bin $B_i$ has lower and upper bounds, and they are non-overlapping, similar how bins work in a histogram. Usually the confidence space is divided into $M$ equally spaced bins ($B_i - B_{i-1} = |B|$), for which $M$ becomes a hyper-parameter to tune.

To select the number of bins $M = |B|$, \marginnote{Setting Number of Bins $|B|$} there is a trade-off to be made. A low number of bins will oversmooth the plot, due to bins containing many samples but not representing the overall confidence distribution, but a high number of bins will produce a spiky look (undersmoothing), due to many bins being empty. A value has to be found between these extremes, usually using heuristics (like $|B| = \sqrt{N}$, ceiling of the square root of number of data points), but it should be noted that there is no best value for $|B|$ and multiple values might produce very similar calibration plots. For more details consult Section \ref{appendix:numbins}.

The diagonal of a calibration plot represents perfect calibration, where predicted confidence equals observed accuracy (see Eq \ref{eq:conf_vs_accuracy}), and visually any calibration error will be visible by being off the diagonal.

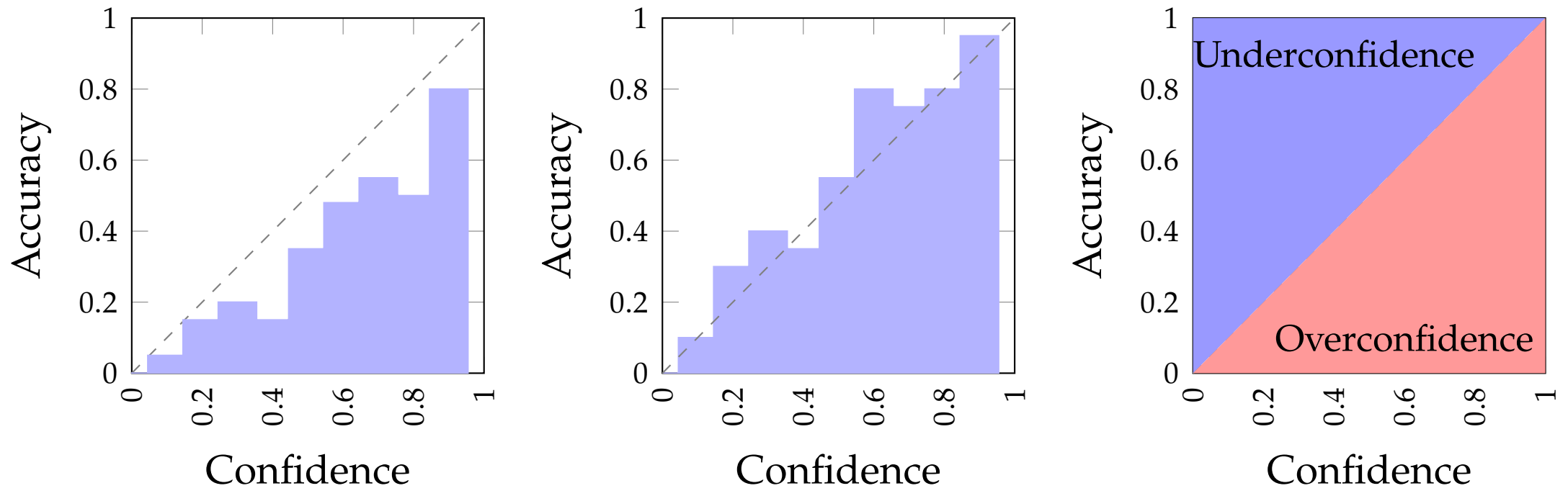
\begin{figure*}[t]
    \centering
    \subfloat[Overconfidence]{
        \centering
        \begin{tikzpicture}
            \begin{axis}[height = 0.20 \textheight, width = 0.45\textwidth, xlabel={Confidence}, ylabel={Accuracy}, xmin = 0.0, xmax = 1.0, ymin = 0.0, ymax = 1.0, ymajorgrids=false, xmajorgrids=false, grid style=dashed, legend pos = north east, legend style={font=\scriptsize}, x tick label style={font=\scriptsize, rotate=90}, y tick label style={font=\scriptsize}]
                
                \addplot+[mark = none, gray, dashed] coordinates { (0, 0) (1, 1)};
                \addplot+[mark = none, blue!30,ybar,fill] coordinates { (0.0,0.0) (0.1, 0.05) (0.2, 0.15) (0.3, 0.2) (0.4, 0.15) (0.5, 0.35) (0.6, 0.48) (0.7, 0.55) (0.8, 0.5) (0.9, 0.8)};
                
            \end{axis}		
        \end{tikzpicture}
    }    
    \subfloat[Underconfidence]{
        \centering
        \begin{tikzpicture}
            \begin{axis}[height = 0.20 \textheight, width = 0.45\textwidth, xlabel={Confidence}, ylabel={Accuracy}, xmin = 0.0, xmax = 1.0, ymin = 0.0, ymax = 1.0, ymajorgrids=false, xmajorgrids=false, grid style=dashed, legend pos = north east, legend style={font=\scriptsize}, x tick label style={font=\scriptsize, rotate=90}, y tick label style={font=\scriptsize}]                
                
                \addplot+[mark = none, blue!30,ybar,fill] coordinates { (0.0,0.0) (0.1, 0.1) (0.2, 0.3) (0.3, 0.4) (0.4, 0.35) (0.5, 0.55) (0.6, 0.8) (0.7, 0.75) (0.8, 0.8) (0.9, 0.95)};
                \addplot+[mark = none, gray, dashed] coordinates { (0, 0) (1, 1)};
                
            \end{axis}		
        \end{tikzpicture}
    }
    \subfloat[Reference]{
    	\begin{tikzpicture}
    		\begin{axis}[height = 0.20 \textheight, width = 0.45\textwidth, xlabel={Confidence}, ylabel={Accuracy}, xmin = 0.0, xmax = 1.0, ymin = 0.0, ymax = 1.0, ymajorgrids=false, xmajorgrids=false, grid style=dashed, legend pos = north east, legend style={font=\scriptsize}, x tick label style={font=\scriptsize, rotate=90}, y tick label style={font=\scriptsize}]
    			
    			\addplot+[mark = none, gray, dashed] coordinates { (0, 0) (1, 1)};S
    			\addplot[draw=none, fill, blue!40, domain=0:1] { x} \drawge;
    			\addplot[draw=none, fill, red!40, domain=0:1] { x} \drawle;
    			
    			\node[] at (axis cs: 0.4,0.9) {\small Underconfidence};
    			\node[] at (axis cs: 0.6,0.1) {\small Overconfidence};
    			
    		\end{axis}		
    	\end{tikzpicture}
    }
    \caption{Examples of calibration plots, including the regions in a calibration plot signaling underconfidence and overconfidence.}
    \label{eval:example_calib_plots}
\end{figure*}

Two example calibration plots are shown in Figure \ref{eval:example_calib_plots}, visually displaying overconfidence and underconfidence. This figure shows the regions for underconfidence and overconfidence, as future reference for the reader.

The previous settings have implicitly assumed a classification task \marginnote{Regression Calibration Plot}, but a calibration plot can also be made for a regression task. Since there is no accuracy metric for regression, we must use confidence intervals. This is shown in Figure \ref{eval:calib_plot_regression_algorithm}. The logic is similar to a classification calibration plot, but instead of grouping confidences in equally spaced bins, the confidence range $[0, 1]$ is divided into $M$ equally sized samples, and then each confidence is used to build a confidence interval with the given confidence, from where it is expected that coverage would match the interval's confidence. The plot is made the same way by plotting confidence and coverage.

One disadvantage of this regression calibration plot formulation is that distributional assumption on predictions, because to build a confidence interval with a given confidence, a distribution must be assumed, often a Gaussian, but other distributions can be used depending on the data and application.

\begin{figure}
    \begin{generalbox}{Regression Calibration Plot}
        \begin{algorithmic}
            \STATE Take predictions of a model with confidence $\hat{y} = [ ( \mu, \sigma^2)_i]$.
            \STATE Define a set of confidences $S_\alpha = [0.1, 0.2, ..., 1.0]$ by dividing the $[0, 1]$ range into $M$ equally spaced values.
            \FOR{each confidence $\alpha \in S_\alpha$}
            \STATE $\hat{I}_\alpha \gets [ \text{CI}(\alpha, \mu, \sigma) \, | \, \forall (\mu, \sigma) \in \hat{y}]$  \algcomment{Build per-prediction CIs}
            \STATE $\text{acc} \gets \text{coverage}(y, \hat{I}_\alpha)$ \algcomment{Compute CI coverage vs true labels}
            \STATE Add $(\alpha, \text{acc})$ to plot.
            \ENDFOR
            \STATE Display plot.        
        \end{algorithmic}
    \end{generalbox}
    \caption{Algorithm to build a calibration plot for regression, based on confidence intervals. The number of confidence values $M$ has to be manually tuned.}
    \label{eval:calib_plot_regression_algorithm}
\end{figure}

While calibration plots are very useful \marginnote{Calibration Error Metrics} for visualization of calibration errors, we often need quantitative metrics to assess the degree of miscalibration produced by a model's confidence. For this several calibration error metrics have been developed. The most basic one is the calibration error:

\begin{definitionbox}{Calibration Error}{calib_error}
    This is the standard metric to measure miscalibration. It is affected by variations in the number of samples in each bin. It is just the absolute error between accuracy and confidence.
    \begin{equation}
        \text{CE} = \sum_i |\text{acc}(B_i) - \text{conf}(B_i)|
    \end{equation}
\end{definitionbox}

A more common metric is the expected calibration error (ECE) \citep{guo2017calibration}, preferred because the standard calibration error is biased due to each bin having different number of elements. The ECE compensates by weighting each absolute error by the number of elements in the bin as a fraction of the total number of elements.

\begin{definitionbox}{Expected Calibration Error}{exp_calib_error}
    \begin{equation}
        \text{ECE} = \sum_i N^{-1} |B_i|  \left|\text{acc}(B_i) - \text{conf}(B_i)\right|
    \end{equation}
    In order to compensate for the varying number of elements in each bin $B_i$, ECE weights the error produced by each bin by the proportion of samples in that bin with respect of the total of samples.\\
    
    This produces a metric that is much more stable and less prone to outliers. This is usually the metric that is commonly reported in scientific publications.
\end{definitionbox}

Finally, for applications that are risk averse, the maximum calibration error might be important, as we often want to minimize the maximum miscalibration in the model, to prevent large errors from being hidden by the expectation used in previous metrics.

\begin{definitionbox}{Maximum Calibration Error}{max_calib_error}
	\begin{equation}
		\text{MCE} = \max_i \left|\text{acc}(B_i) - \text{conf}(B_i)\right|
	\end{equation}
	For risk averse applications, the MCE computes the maximum level of miscalibration, so appropriate risk can be estimated and addressed by the application designed. Variations in miscalibration in each bin can hide in the overall mean or expectation.
\end{definitionbox}

As mentioned before, for scientific publications the most common metrics is the expected calibration error, and should be used as a default choice when performing model comparison or hyper-parameter tuning in terms of calibration error.

\section{Evaluating Model and Data Uncertainties}

Previous sections in this chapter have covered methods to evaluate the predictive uncertainty from a model, but data and model uncertainties do have specific expectations that can be used for evaluation. First we describe expectations for model uncertainty. Two main properties in particular are important.
\begin{marginfigure}
	\begin{tikzpicture}        
		\begin{axis}[width=\linewidth, domain=0.0:1, xlabel={Training Set Size}, ylabel={Model Uncertainty}, xmajorticks=false, ymajorticks=false]
			\addplot+[mark=none, smooth] {exp(-5.0 * x)};
		\end{axis}
	\end{tikzpicture}
	\caption{Expected behavior of epistemic uncertainty when varying the training set size.}
	\label{fig:epi_expectations}
\end{marginfigure}
The interpretation of model uncertainty is:
\begin{equation}
	\condprob{\hat{y}}{x} = \prob{\hat{y} \text{ is correct}}
\end{equation}
This is, the (epistemic) probability or confidence output by the model, is the probability that the prediction is correct.

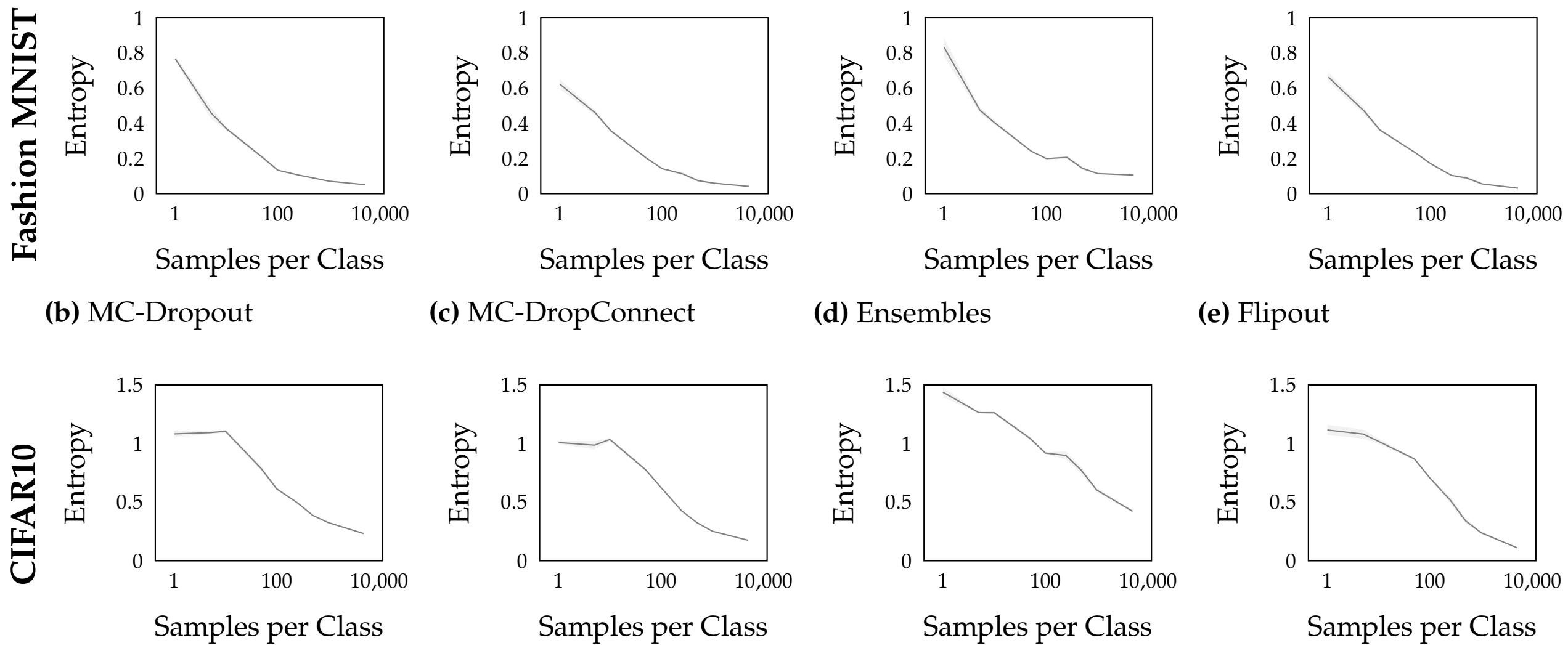
\begin{figure*}
	\subfloat{
		\rotatebox{90}{\hspace*{0.5em} \textbf{Fashion MNIST}}
	}
	\subfloat[MC-Dropout]{
		\begin{tikzpicture}
			\begin{axis}[width = 0.25\linewidth, xlabel={Samples per Class}, ylabel={Entropy}, ymajorgrids=false, xmajorgrids=false, grid style=dashed, tick label style={font=\tiny}, xmode=log, log ticks with fixed point, ymin=0.0, ymax=1.0, ytick style={draw=none}, xtick style={draw=none}, label style={font=\footnotesize}]
				
				\plotwitherrorareas{data/sampcomp/entropy-vs-SPC-fashion_mnist-results-miniVGG-dropout-combined.csv}{gray}{DE}{mean_mean_entropy}{std_mean_entropy}
				\legend{}
			\end{axis}		            
		\end{tikzpicture}
	}
	\subfloat[MC-DropConnect]{
		\begin{tikzpicture}
			\begin{axis}[width = 0.25\linewidth, xlabel={Samples per Class}, ylabel={Entropy}, ymajorgrids=false, xmajorgrids=false, grid style=dashed, tick label style={font=\tiny}, xmode=log, log ticks with fixed point, ymin=0.0, ymax=1.0, ytick style={draw=none}, xtick style={draw=none}, label style={font=\footnotesize}]
				
				\plotwitherrorareas{data/sampcomp/entropy-vs-SPC-fashion_mnist-results-miniVGG-dropconnect-combined.csv}{gray}{DE}{mean_mean_entropy}{std_mean_entropy}
				\legend{}
			\end{axis}		            
		\end{tikzpicture}
	}
	\subfloat[Ensembles]{
		\begin{tikzpicture}
			\begin{axis}[width = 0.25\linewidth, xlabel={Samples per Class}, ylabel={Entropy}, ymajorgrids=false, xmajorgrids=false, grid style=dashed, tick label style={font=\tiny}, xmode=log, log ticks with fixed point, ymin=0.0, ymax=1.0, ytick style={draw=none}, xtick style={draw=none}, label style={font=\footnotesize}]
				
				\plotwitherrorareas{data/sampcomp/entropy-vs-SPC-fashion_mnist-results-miniVGG-deepensemble-combined.csv}{gray}{DE}{mean_mean_entropy}{std_mean_entropy}
				\legend{}
			\end{axis}		            
		\end{tikzpicture}
	}
	\subfloat[Flipout]{
		\begin{tikzpicture}
			\begin{axis}[width = 0.25\linewidth, xlabel={Samples per Class}, ylabel={Entropy}, ymajorgrids=false, xmajorgrids=false, grid style=dashed, tick label style={font=\tiny}, xmode=log, log ticks with fixed point, ymin=0.0, ymax=1.0, ytick style={draw=none}, xtick style={draw=none}, label style={font=\footnotesize}]
				
				\plotwitherrorareas{data/sampcomp/entropy-vs-SPC-fashion_mnist-results-miniVGG-flipout-combined.csv}{gray}{DE}{mean_mean_entropy}{std_mean_entropy}
				\legend{}
			\end{axis}		            
		\end{tikzpicture}
	}
	
	\subfloat{
		\rotatebox{90}{\hspace*{2em}\textbf{CIFAR10}}
	}
	\subfloat[MC-Dropout]{
		\begin{tikzpicture}
			\begin{axis}[width = 0.25\linewidth, xlabel={Samples per Class}, ylabel={Entropy}, ymajorgrids=false, xmajorgrids=false, grid style=dashed, tick label style={font=\tiny}, xmode=log, log ticks with fixed point, ymin=0.0, ymax=1.5, ytick style={draw=none}, xtick style={draw=none}, label style={font=\footnotesize}]
				
				\plotwitherrorareas{data/sampcomp/entropy-vs-SPC-cifar10-results-miniVGG-dropout-combined.csv}{gray}{DE}{mean_mean_entropy}{std_mean_entropy}
				\legend{}
			\end{axis}		            
		\end{tikzpicture}
	}
	\subfloat[MC-DropConnect]{
		\begin{tikzpicture}
			\begin{axis}[width = 0.25\linewidth, xlabel={Samples per Class}, ylabel={Entropy}, ymajorgrids=false, xmajorgrids=false, grid style=dashed, tick label style={font=\tiny}, xmode=log, log ticks with fixed point, ymin=0.0, ymax=1.5, ytick style={draw=none}, xtick style={draw=none}, label style={font=\footnotesize}]
				
				\plotwitherrorareas{data/sampcomp/entropy-vs-SPC-cifar10-results-miniVGG-dropconnect-combined.csv}{gray}{DE}{mean_mean_entropy}{std_mean_entropy}
				\legend{}
			\end{axis}		            
		\end{tikzpicture}
	}
	\subfloat[Ensembles]{
		\begin{tikzpicture}
			\begin{axis}[width = 0.25\linewidth, xlabel={Samples per Class}, ylabel={Entropy}, ymajorgrids=false, xmajorgrids=false, grid style=dashed, tick label style={font=\tiny}, xmode=log, log ticks with fixed point, ymin=0.0, ymax=1.5, ytick style={draw=none}, xtick style={draw=none}, label style={font=\footnotesize}]
				
				\plotwitherrorareas{data/sampcomp/entropy-vs-SPC-cifar10-results-miniVGG-deepensemble-combined.csv}{gray}{DE}{mean_mean_entropy}{std_mean_entropy}
				\legend{}
			\end{axis}		            
		\end{tikzpicture}
	}
	\subfloat[Flipout]{
		\begin{tikzpicture}
			\begin{axis}[width = 0.25\linewidth, xlabel={Samples per Class}, ylabel={Entropy}, ymajorgrids=false, xmajorgrids=false, grid style=dashed, tick label style={font=\tiny}, xmode=log, log ticks with fixed point, ymin=0.0, ymax=1.5, ytick style={draw=none}, xtick style={draw=none}, label style={font=\footnotesize}]
				
				\plotwitherrorareas{data/sampcomp/entropy-vs-SPC-cifar10-results-miniVGG-flipout-combined.csv}{gray}{DE}{mean_mean_entropy}{std_mean_entropy}
				\legend{}
			\end{axis}		            
		\end{tikzpicture}
	}
	\caption[Effect of training set size on predictive uncertainty over Fashion-MNIST and CIFAR10 datasets]{Effect of training set size on predictive uncertainty over Fashion-MNIST and CIFAR10 datasets. Uncertainty measured as predictive entropy overall reduces with a larger training set. Results from \cite{valdenegro2021exploring}.}
    \label{entropy_vs_spc}
\end{figure*}

The second important property, is that model uncertainty scales with size of the training set. Lower training set sizes will increase model uncertainty, and this uncertainty will decrease if the training set increases in size. This is shown conceptually in Figure \ref{fig:epi_expectations} and in a real setting in Figure \ref{entropy_vs_spc}.  This is because a model that has learned from a small training set will have gaps in its knowledge that are reflected in model uncertainty, and these gaps can only be "patched" by increasing knowledge during training, which is accomplished by learning from a larger training set.

\begin{definitionbox}{Expectations from Model Uncertainty}{exp_epistemic}
	\begin{description}[\compact\setlabelphantom{Training Set Size}]
		\item[Interpretation] Epistemic uncertainty concerns the model's belief on its prediction, so it should be interpreted as the probability of the prediction being correct.
		\item[Training Set Size] Epistemic uncertainty should be higher for a given task if the training set becomes smaller, for example via downsampling the training set \citep{hullermeier2021aleatoric}.
		\item[Outside Training Set] Epistemic uncertainty should be higher for samples that are outside the training set, compared to epistemic uncertainty inside the training set.
		\item[Inside Training Set] Epistemic uncertainty should be low, close to zero, inside the training set, since the model has learned everything from its training set.		
		\item[No Ground Truth Labels] Epistemic uncertainty has no ground truth labels, as it is completely defined by the model and not by data itself.
		\item[Reproducibility] If a model is trained multiple times, with different (random) initial weights but on the same training data, then epistemic uncertainty will change, as weight values determine the model predictions.
	\end{description}
\end{definitionbox}

Expectations for data uncertainty are very different than model uncertainty. Data uncertainty is uncertainty in the data itself, mostly represented as noise in labels or in input features, but the most common definition of data uncertainty is noise in labels or incorrect labels.

Then the only interpretation of data uncertainty is noise in labels. Data uncertainty does not change with size variations of the training set, but unlike model uncertainty, data uncertainty can have ground truth labels, such as the variance of the noise distribution.

\section{Plots for Uncertainty Evaluation}
In previous sections we have described calibration plots as one of the main tools to evaluate your uncertainty estimates, for classification and regression settings. But there are more options if some constraints are loosened, which might reveal interesting properties of uncertainty produced by your model.

\subsection{Correct vs Incorrect Predictions}
The aim of model uncertainty is to be useful to the human user as a likelihood for the prediction to be correct. It is natural to expect that model confidence will be able to separate correct vs incorrect predictions. Correct predictions are expected to have high confidence, while incorrect predictions should have low confidence.

\begin{figure}
    \subfloat[Easy]{
        \begin{tikzpicture}        
            \begin{axis}[width=0.47\linewidth, domain=-10:15, tick label style={rotate=30}, ymajorticks=false, xmajorticks=false, xlabel={Uncertainty}]
                \addplot+[mark=none, smooth, fill=blue!50] {gauss(-3.0, 1.0)};
                \addplot+[mark=none, smooth, fill=red!50] {gauss(9.0, 1.5)};
            \end{axis}
        \end{tikzpicture}
    }
    \subfloat[Medium]{
        \begin{tikzpicture}        
            \begin{axis}[width=0.47\linewidth, domain=-10:15, tick label style={rotate=30}, ymajorticks=false, xmajorticks=false, xlabel={Uncertainty}]
                \addplot+[mark=none, smooth, fill=blue!50] {gauss(0.0, 1.0)};
                \addplot+[mark=none, smooth, fill=red!50] {gauss(6.0, 1.5)};
            \end{axis}
        \end{tikzpicture}
    }
    \subfloat[Hard]{
        \begin{tikzpicture}        
            \begin{axis}[width=0.47\linewidth, domain=-10:15, tick label style={rotate=30}, ymajorticks=false, xmajorticks=false, xlabel={Uncertainty}]
                \addplot+[mark=none, smooth, fill=blue!50] {gauss(3.0, 1.0)};
                \addplot+[mark=none, smooth, fill=red!50] {gauss(6.0, 1.5)};
            \end{axis}
        \end{tikzpicture}
    }
    \caption[Uncertainty expectations for correct vs incorrect predictions by uncertainty distribution]{Uncertainty expectations for {\color{blue} correct} vs {\color{red} incorrect} predictions by uncertainty distribution. Correct predictions should have low uncertainty, while incorrect predictions should have high uncertainty. Three difficulty settings are shown.}
    \label{fig:correct_vs_incorrect_unc}
\end{figure}
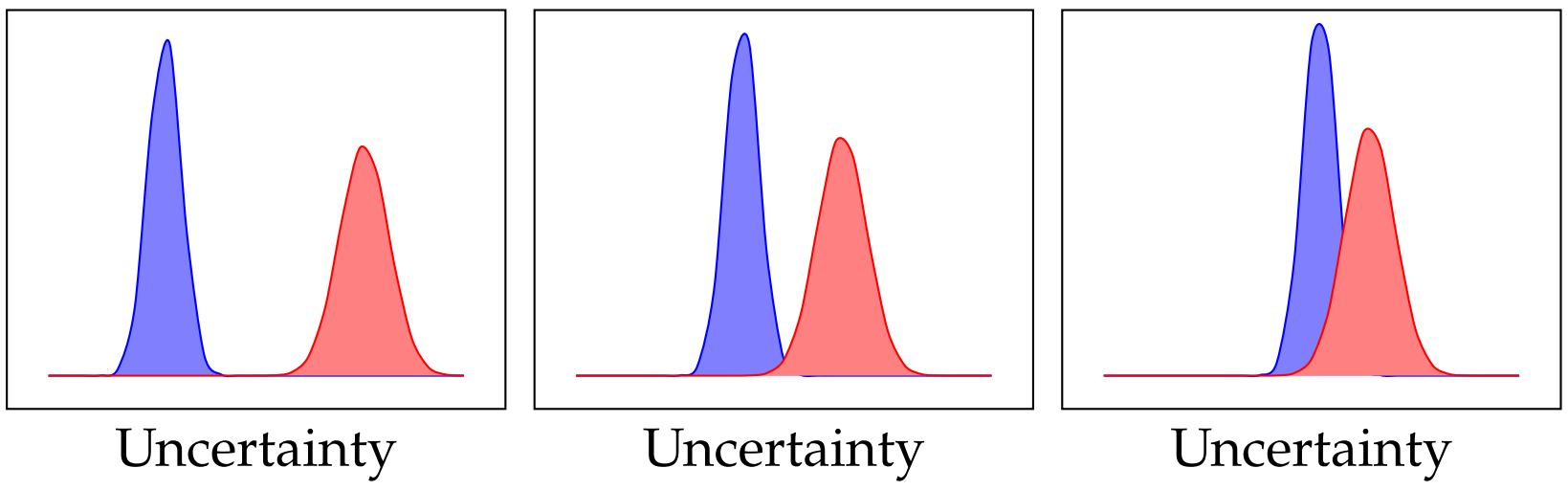

To evaluate the expected behavior of uncertainty or confidence, make predictions over a dataset, and using the ground truth labels, determine which predictions are correct and which are incorrect, together with the model confidence. A density plot can be made by separately plotting confidences for correct and incorrect predictins, as shown in Figure \ref{fig:correct_vs_incorrect_unc}.

\begin{definitionbox}{Expectations from Data Uncertainty}{exp_aleatoric}
    \begin{description}[\compact\setlabelphantom{Always Present}]
    	\item[Interpretation] Aleatoric uncertainty is noise in the data, mostly in labels, like incorrect or noisy labels, but can also be noise in input features.
    	\item[Ground Truth Labels] Aleatoric uncertainty can have ground truth labels, in the form of label uncertainty, but this requires a special setup, such as multiple measurements from where variance or probability can be estimated, or crowdsourcing if labels are made by humans. In all cases the aleatoric uncertainty label would be the label probability distribution for classification, or the standard deviation of the label noise for regression.
    	\item[Always Present] Aleatoric uncertainty is never exactly zero, as all datasets have a level of incorrect or noisy labels, so aleatoric uncertainty is always present in a dataset, particularly in its labels.
    	\item[Reproducibility] If a model is trained multiple times, with different (random) initial weights but on the same training data, then aleatoric uncertainty should be the same up to numerical error.
    \end{description}
\end{definitionbox}

The degree of separation between the correct and incorrect confidence distributions defines the quality of uncertainty. Ideally correct and incorrect confidences would have a large separation, but due to approximations and (feature) distribution shift, the two confidence distributions can have significant overlap, indicating that confidence/uncertainty does not separate correct from incorrect predictions.

Metrics such as the area under the ROC curve can be used to quantify the quality of uncertainty in this setting, this problem is called misclassification prediction.

\subsection{Error vs Confidence Threshold Plots}

Another way to visualize the relationship between error and confidence is by considering another expectation of uncertainty. If we threshold uncertainty or confidence values, we expect that as we remove predictions with low confidence, the remaining predictions will be more accurate, and have higher accuracy (classification) or lower error (regression).

For classification, a confidence value such as the maximum probability $\max_c \condprob{\hat{y}^c}{x}$, is tested against a threshold $\tau$. All predictions with confidence passing this threshold are kept, and accuracy is computed for these predictions $\hat{Y}_\tau$:
\begin{equation}
	\hat{Y}_\tau = \{\hat{y}_i \,|\, \max_c \condprob{\hat{y}_i^c}{x} \geq \tau \}
\end{equation}
Then accuracy is computed for $\hat{Y}_\tau$ using the corresponding ground truth labels, and $(\tau, \text{accuracy}(Y, \hat{Y}_\tau))$ values are plotted. This is shown in Figure \ref{fig:conf_error_plot_threshold}.
\begin{marginfigure}
	\begin{tikzpicture}
		\begin{axis}[height = 0.18 \textheight, width = \columnwidth, xlabel={Confidence ($\tau$)}, ylabel={Acc (\%) for $y_e > \tau$}, xmin = 0.0, xmax = 1.0, ymin= 70, ymax=100, ymajorgrids=false, xmajorgrids=false, grid style=dashed, legend pos = north east, legend style={font=\scriptsize}, tick label style={font=\scriptsize}]
			
			\addplot+[mark = none, red] table[x  = conf, y  = acc, col sep = semicolon] {data/svhn-accuracy-confidence-deepensembles-num-ens-5_cnn_svhn.csv};			
			\addplot+[mark = none, green] table[x  = conf, y  = acc, col sep = semicolon] {data/svhn-accuracy-confidence-sub-deepensembles_1_num-ens-5_cnn_svhn.csv};			
			\addplot+[mark = none, blue] table[x  = conf, y  = acc, col sep = semicolon] {data/svhn-accuracy-confidence-sub-deepensembles_2_num-ens-5_cnn_svhn.csv};        			
			\addplot+[mark = none, black] table[x  = conf, y  = acc, col sep = semicolon] {data/svhn-accuracy-confidence-sub-deepensembles_3_num-ens-5_cnn_svhn.csv};        
		\end{axis}		
	\end{tikzpicture}
	\caption{Example of a error vs confidence threshold plot for a classification setting, showing accuracy instead of error. Each curve represents a different model.}
	\label{fig:conf_error_plot_threshold}
\end{marginfigure}

For regression, the prediction standard deviation is instead put tested a threshold $\tau$, and a linear error metric (i.e. mean absolute error) is used to compute an error for remaining predictions. A plot is made for values $(\tau, \text{error}(Y, \hat{Y}_\tau))$.
\begin{equation}
	\hat{Y}_\tau = \{\hat{y}_i \,|\, \sigma^i \geq \tau \quad \forall (\mu_i, \sigma_i^2) \in \hat{Y}\}
\end{equation}

The interpretation of this plot is that as the confidence threshold $\tau$ increases, predictions should be more accurate and/or have lower error. This is useful to compare different models or uncertainty quantification methods in the same task, as shown in Figure \ref{fig:conf_error_plot_threshold}, the model producing the highest curve (accuracy) or lowest curve (error) is the best performing one, but there is no direct baseline to compare like a calibration plot. Often a synthetic baseline is made by comparing to a model producing a random uncertainty estimate, which should produce a flat curve.

\subsection{Error vs Confidence Scatterplots}
\begin{marginfigure}
	\begin{tikzpicture}
		\begin{axis}[height = 0.18 \textheight, width = \columnwidth, xlabel={Std Deviation}, ylabel={Absolute Error}, xmin = 0.0, xmax = 1.0, ymin= 0, ymax=40, ymajorgrids=false, xmajorgrids=false, grid style=dashed, legend pos = north east, legend style={font=\scriptsize}, tick label style={font=\scriptsize}]
			
			\addplot+[scatter, only marks, mark size=0.9pt, point meta=false] table[x = conf, y = error, col sep = semicolon] {data/conf-vs-error-ensembles-toyregression.csv};
		\end{axis}		
	\end{tikzpicture}
	\caption{Example of a error vs confidence scatterplot for a regression setting, using standard deviation as uncertainty.}
	\label{fig:conf_error_scatterplot}
\end{marginfigure}
The final plot we will cover is a simplification of the previous plot. Instead of thresholding confidence or uncertainty, it is possible to directly plot error vs confidence as a scatterplot.

A error vs confidence scatterplot is built by making predictions over a dataset, and then for each sample, compute an error metric that is in the same units as confidence or uncertainty, like mean absolute error, root mean squared error, or classification error, and then plot $(\text{confidence}, \text{error})$ in a scatterplot, as shown in Figure \ref{fig:conf_error_scatterplot}.

Error vs confidence scatterplots can often be messy but still show a clear pattern. Analyzing Figure \ref{fig:conf_error_scatterplot}, it is clear that there are two uncertainty patterns, but in both cases the standard deviation does have a relationship to predict error, just not in the same scale.

Some small details for this kind of plot, in regression settings sometimes it is best to normalize the standard deviation if multiple models will be compared, and standard deviation and error should be in the same (linear) scale, comparing standard deviation vs mean squared error would be a mistake since the squared error is in square units compared to the standard deviation.

Error vs confidence scatterplots can also be summarized with some metrics, examples can be the pearson or spearman correlation coefficients, to show if there is a relationship between confidence and error, but this relationship does not have to be linear.

\section{Summary}

This chapter has presented many loss functions and metrics that can be used to evaluate your uncertainty estimates. An important message is that standard metrics do not evaluate quality of uncertainty, and specialized metrics and plots have to be used.

Proper scoring rules ensure that a scoring rule correctly evaluates distribution outputs, and should be preferred to other scoring rules, specially when evaluating uncertainty estimates. Calibration is the concept where confidence can be used as a proxy for accuracy or error, and it can be evaluated using calibration plots and different calibration error metrics.

We have described the (different) expectations for model and data uncertainty, from where evaluation methodologies can be developed, depending on application, task, and uncertainty type.

Finally we have shown multiple graphical ways to visualize the behavior of uncertainty, allowing for a thorough evaluation of uncertainty estimates.

%% file: sections/calibration.tex
j\chapter{Calibration}
\label{ch:calibration}
\margintoc

We have covered methods to perform uncertainty estimation in machine learning models, but unfortunately there is no guarantee that predicted confidence will be a faithful representation of true accuracy. This problem is called miscalibration, and this chapter will first explore the different concepts of calibration, and then show the most common methods for improving calibration.

\section{Definition}
The general concept of calibration is that prediction confidence should be proportional to accuracy.

\begin{equation}
	\overUnderArrow[\parbox{3.5cm}{Probability predicted by your Model}]{}{\text{Confidence}} = \overUnderArrow[\parbox{4cm}{Observed frequency of correct predictions}]{}{\text{Accuracy}}
\end{equation}
This simple equality should hold for all confidence levels. In simple words, this means that the empirical probabilities are equal to the predicted probabilities or confidences by the model.

Additionally, the interpretation of a predicted confidence score should be that is represents the probability of a prediction being correct.

Calibration is a property of a model that can only be measured over a whole dataset, due to requiring a measure of accuracy, which can only be computed over a set of predictions. This is also noted by \cite{song2021classifier}:

\begin{quotation}
	Determining the degree to which a forecaster is well-calibrated cannot be done on a per-forecast basis, but rather requires looking at a sufficiently large and diverse set of forecasts.
    \begin{flushright}
        -- \cite{song2021classifier}
    \end{flushright}
\end{quotation}
This means calibration can only be measured on a dataset, not directly in each prediction. It is fundamentally a frequentist measure.

We now make concrete definitions of calibration for classification and regression tasks.

\begin{definitionbox}{Calibration in Classification}{calib_classification}
	We can now formally define what calibration of probabilities for classification means. The following should hold for all confidence levels $\alpha \in [0, 1]$:
	\begin{equation}
		\condprob{y_\alpha}{x} = |\hat{y}_\alpha|^{-1} \sum_{i \in \hat{y}_\alpha} \mathrm{1}[\hat{y}_i = y_i]
	\end{equation}
	Where $\hat{y}_\alpha$ is the set of all predictions $\hat{y}$ with confidence $\alpha$. This is usually where binning is required to group confidences around $\alpha$ since it is a continuous value.	
\end{definitionbox}

\begin{definitionbox}{Calibration in Regression}{calib_regression}
	For regression, we use confidence intervals:
	\begin{equation}
		\prob{l_\alpha \leq y \leq u_\alpha} = \alpha
	\end{equation}
	Where $[l_\alpha, u_\alpha]$ are the bounds of a confidence interval, which is a function of the confidence level $\alpha$, and predicted mean $\hat{\mu}_i$ and variances $\hat{\sigma}^2_i$.\\
	In terms of observed frequencies, with $\hat{y}_\alpha$ is the set of all predictions $\hat{y}$ around confidence interval with confidence $\alpha$:
	\begin{equation}
		|\hat{y}_\alpha|^{-1} \sum_{i \in \hat{y}_\alpha} \mathrm{1}[\hat{y}^l_{\alpha, i} \leq y \leq \hat{y}^h_{\alpha, i}] = \alpha
	\end{equation}
	Again this should hold for all confidence levels $\alpha \in [0, 1]$. This can be interpreted as the empirical $\alpha$-confidence interval should contain the predicted mean with probability $\alpha$.
\end{definitionbox}

There are other conceptual interpretations of calibration, for example calibration in general means using a known scale.

For probabilities, you might be producing numbers in $[0, 1]$, but these are not probabilities, only when you can give them a probabilistic interpretation, and that they represent the probability of a prediction being correct, you can use these as probabilities and confidences.

A model might make random confidence predictions, which are not calibrated. We have extensively discussed the safety needs for calibrated probabilities, specially when using predictions to make decisions over humans.

Calibration is not a unique objective, the model still needs to make good predictions, it can be considered to be an additional objective, with the first objective being supervised by the common cross-entropy or squared error losses that guide the model to make correct predictions.

\begin{definitionbox}{Degenerate Calibration Cases}{calib_degen}
    There are cases where calibration might seem to be perfect and yet probabilities are not being an indication of a prediction's correctness.
    
    \textbf{Regression}. If we predict the dataset mean, and a very large (or infinite) variance, this will always be calibrated (have zero calibration error), since it will always cover the true values. But now the confidence interval bounds do not have any meaningful information.
    
    \textbf{Classification}. If a model outputs constant confidence equal to the model's accuracy, then it will also be perfectly calibrated but now the predicted confidence is not informative of prediction accuracy.    
\end{definitionbox}

Basically, there are ways to \emph{cheat} calibration, so calibration by itself is not the only goal during model training.

An important and often asked question now comes up: why are models and predictions miscalibrated? Why models after training, are not naturally adapted to provide a confidence value that works as a proxy for accuracy?

The answer is very simple, models are often trained to maximize accuracy, not to reason about their confidence. The following definitions show how this works in classification and regression settings.

\begin{definitionbox}{Miscalibration Sources in Classification}{miscalib_classification}
	This happens for multiple reasons:
	
	\begin{description}
		\item[Loss] The cross-entropy loss with one-hot encoded labels, usually makes the predicted probabilities to be maximized (towards $1.0$), without considering that they can be overconfident due to incorrect predictions.
		\item[Multi-Class] Depending on the multi-class formulation (for example, SVMs with one-vs-all or all-vs-all), which break the assumptions used to evaluate calibration, and generally are hacks to transform a binary model into supporting multiple classes.
		\item[Not Probabilities] Some machine learning models like SVMs do not output confidences, but abstract measures of confidence, which are generally not calibrated and are not probabilities.
	\end{description}
\end{definitionbox}

\begin{definitionbox}{Miscalibration Sources in Regression}{miscalib_regression}
	\begin{description}
		\item[Loss] The Gaussian negative log-likelihood supervises the variance outputs, and tries to cover data points according to the estimated noise, estimating only aleatoric uncertainty. Generally this estimation is pretty good, but once epistemic uncertainty is added, then the combined variances are most likely miscalibrated.
		\item[Confidence Levels] The Gaussian NLL loss models a Gaussian, but it is not aware that we use particular confidence intervals to evaluate calibration, depending on aleatoric uncertainty, the true label might be outside of the expected confidence interval.
	\end{description}
\end{definitionbox}

\begin{figure}
	\subfloat[Neural Network, ECE = 0.18]{
		\centering
		\begin{tikzpicture}
			\begin{axis}[height = 0.20 \textheight, width = 0.45\textwidth, xlabel={Confidence}, ylabel={Accuracy}, xmin = 0.0, xmax = 1.0, ymin = 0.0, ymax = 1.0, ymajorgrids=false, xmajorgrids=false, grid style=dashed, legend pos = north east, legend style={font=\scriptsize}, x tick label style={font=\scriptsize, rotate=90}, y tick label style={font=\scriptsize}]
				
				\addplot+[mark = none, gray, dashed] coordinates { (0, 0) (1, 1)};
				\addplot+[mark = none, blue!30,ybar,fill] coordinates { (0.0,0.0) (0.1, 0.1) (0.2, 0.2) (0.3, 0.3) (0.4, 0.95) (0.5, 0.2) (0.6, 0.5) (0.7, 0.58) (0.8, 0.6) (0.9, 0.67) (1.0, 0.93)};
				
			\end{axis}		
		\end{tikzpicture}
	}    
	\subfloat[BNN with MC-Dropout, ECE = 0.11]{
		\centering
		\begin{tikzpicture}
			\begin{axis}[height = 0.20 \textheight, width = 0.45\textwidth, xlabel={Confidence}, ylabel={Accuracy}, xmin = 0.0, xmax = 1.0, ymin = 0.0, ymax = 1.0, ymajorgrids=false, xmajorgrids=false, grid style=dashed, legend pos = north east, legend style={font=\scriptsize}, x tick label style={font=\scriptsize, rotate=90}, y tick label style={font=\scriptsize}]                
				
				\addplot+[mark = none, blue!30,ybar,fill] coordinates { (0.0,0.0) (0.1, 0.1) (0.2, 0.2) (0.3, 0.6) (0.4, 0.2) (0.5, 0.5) (0.6, 0.8) (0.7, 0.75) (0.8, 0.85) (0.9, 0.95) (1.0, 0.98)};
				\addplot+[mark = none, gray, dashed] coordinates { (0, 0) (1, 1)};
				
			\end{axis}		
		\end{tikzpicture}
	}
	\caption[Comparison of calibration between a standard neural network and a BNN using MC-Dropout on MNIST]{Comparison of calibration between a standard neural network and a BNN using MC-Dropout, both trained on MNIST, showing that by using uncertainty quantification methods, calibration error usually reduces, around 40\% in this case, but it does not automatically decrease to zero.}
	\label{uq_calib}
\end{figure}
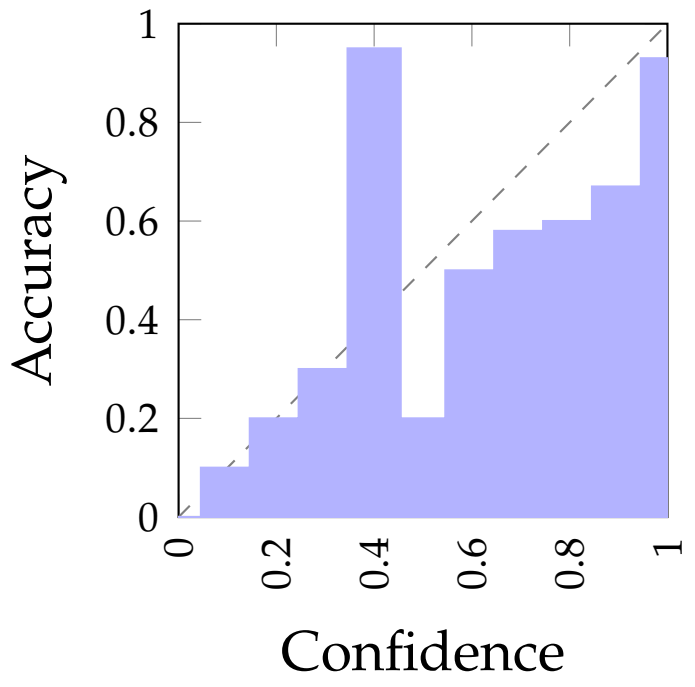
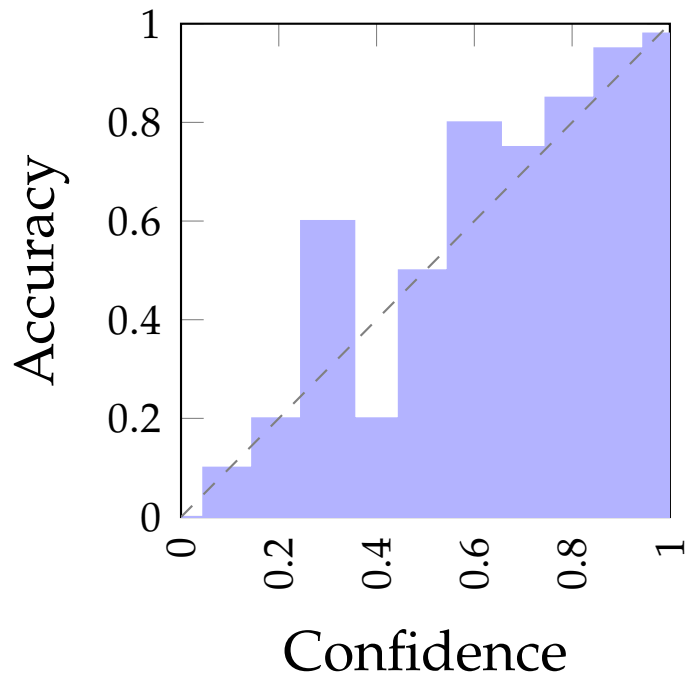

Finally, the use of uncertainty quantification methods also naturally improves calibration, as shown in Figure \ref{uq_calib}, where using MC-Dropout reduces the calibration error by around 40\% compared to the same neural network without Dropout.

\section{Types of Calibration}
\label{calib:sec_types}

So far we have described the overall concept of calibration. But there are multiple ways to formulate calibration, depending on how a confidence is defined, particularly for classification.

\begin{definitionbox}{Confidence Calibration}{conf_calibration}
	The class with largest probability is evaluated for calibration.
	\begin{equation}
		\sigma = \max_{i} p_i
	\end{equation}
    With $i = 0, 1, ..., C$ with $C$ being the number of classes.
\end{definitionbox}

\begin{definitionbox}{Classwise Calibration}{classwise_calibration}
	Each class needs to be individually calibrated, in a one-vs-all or one-vs-rest setting.\\
	
	This means that calibration is evaluated as binary classification for each class, with one class versus the rest.\\
	
	For this purpose a calibration error can be computed and evaluated for each class (independent of each other).
\end{definitionbox}

\begin{definitionbox}{Multi-class Calibration}{multiclass_calibration}
	All classes are jointly calibrated, and calibration error is evaluated for all classes at the same time.\\
	
	For example, this would be combining a mean calibration error for all classes, and optimizing it jointly.
\end{definitionbox}

These definitions can be ordered by strength of calibration, with Multi-class Calibration being the strongest form.

\begin{equation*}
	\text{Confidence} < \text{Classwise} < \text{Multiclass}
\end{equation*}

This strength categorization can also be considered in terms of difficulty, as calibrating the maximum probability (confidence) is easier than jointly calibrating all classes (multiclass).

\section{Calibration and Sources of Uncertainty}

There is a relationship between the concept of calibration and the two sources of uncertainty. Calibration could be evaluated separately for model and data uncertainty. This section shows the relationship between sources of uncertainty and calibration, which gives intuitions on how to improve calibration.

In general model uncertainty is better expected to be calibrated, since it relates to the model and how it makes a prediction, so it should gauge the confidence level of the prediction being correct directly.

Data uncertainty has an effect of modelling the label noise, which might produce incorrect predictions (due to noise), or increase the size of confidence intervals, but does not directly interact with the prediction being correct, this is exclusively expected from model uncertainty.

There is a well known way to decompose any proper scoring rule into some interesting terms, let's start with the overall decomposition, shown in \cite{kull2015novel}:
\begin{equation*}
	\overUnderArrow[\parbox{2.5cm}{Expected Scoring Rule}]{}{\expect{d(\hat{Y}, Y)}} = \overUnderArrow[\parbox{2.8cm}{Expected Calibration Loss}]{}{\expect{d(\hat{Y}, C)}} + \overUnderArrow[\parbox{2.8cm}{Expected Refinement Loss}]{}{\expect{d(C, Y)}}
\end{equation*}
\marginnote{Here $d$ is a divergence associated to the scoring rule. For the cross-entropy loss, $d$ is the KL-divergence, and for the Brier score it is the mean squared error.}

Where  $Y$ are the ground truth probabilities, $\hat{Y}$ are the predicted probabilities of your model, and $C$ are the calibrated probabilities, which are perfect empirical probabilities that would produce zero calibration error, in a way the best probabilities that a model could produce. Note that in general we do not know $C$.

The calibration loss $\expect{d(\hat{Y}, C)}$ measures how well the predicted probabilities $\hat{Y}$ approximate the calibrated probabilities $C$, while the refinement loss measures how well the calibrated $C$ and ground truth probabilities $Y$ match.

Refinement loss is also called \emph{sharpness}, and is basically a measure of how wide the predicted distribution is. A sharp prediction is precise, and confidence indicates that the model is very confident, while a blunt prediction has a large variance and low confidence.

How do we connect this formulation with a Bayesian point of view? If we replace $C = Q$, where $Q$ are the true posterior probabilities, which are model independent and can be estimated as $Q_i = \expectcond{Y_i}{X}$. These can be interpreted as the true proportion of class $i$ among instances with feature values $X$, then:
\begin{equation*}
	\overUnderArrow[\parbox{2.5cm}{Expected Scoring Rule}]{}{\expect{d(\hat{Y}, Y)}} = \overUnderArrow[\parbox{2.8cm}{Expected Epistemic Loss}]{}{\expect{d(\hat{Y}, Q)}} + \overUnderArrow[\parbox{2.8cm}{Expected Aleatoric Loss}]{}{\expect{d(Q, Y)}}
\end{equation*}
These terms now bear a clear relationship with model and data uncertainty:

\textbf{Epistemic Loss}. The term $\expect{d(\hat{Y}, Q)}$ measures the differences between predicted probabilities $\hat{Y}$ and true posterior probabilities $Q$, and this term is reducible since it depends on predicted probabilities made by a model, a different model could make better probabilities that have optimal alignment with the true posterior probabilities $Q$. This is then a form of model or epistemic uncertainty.

\textbf{Aleatoric Loss}. The term $\expect{d(Q, Y)}$ is irreducible since it measures the alignment between the ground truth probabilities $Y$ and the true posterior probabilities $Q$, both which are defined by the data, so this term represents data or aleatoric uncertainty.

\begin{definitionbox}{Decomposition of Proper Scoring Rules}{decomp_proper_scoring}
    \textbf{Model Uncertainty}. Loss or error produced by the model, miscalibration is part of epistemic uncertainty.
    
    \textbf{Data Uncertainty}. Loss or error produced by the data, including the difference between calibrated probabilities $C$ and ground truth probabilities $Y$, meaning that ground truth probabilities might prevent a model from being perfectly calibrated, this is a form of incorrect labels.\\
    
    The decomposition of a proper scoring rule shows how data and model uncertainty are related, and how high data uncertainty affects the calibration of a model, but also that calibration is a property of model uncertainty.
    
    High data uncertainty will make it hard for the model to have a sharp (confident) prediction, while low data uncertainty makes it easy to learn a confident/sharp predictions and this leads to better calibrated models.
\end{definitionbox}

Here we have made a direct connection between uncertainty decomposition and calibration, from this formulation it is clear that calibration derives from a form of epistemic loss, which is the model uncertainty, while data uncertainty derives from the difference between the true predicted posterior distribution $Q$ and the labels $Y$. Do not forget that this decomposition only applies to proper scoring rules, which is the majority of loss functions that we use.

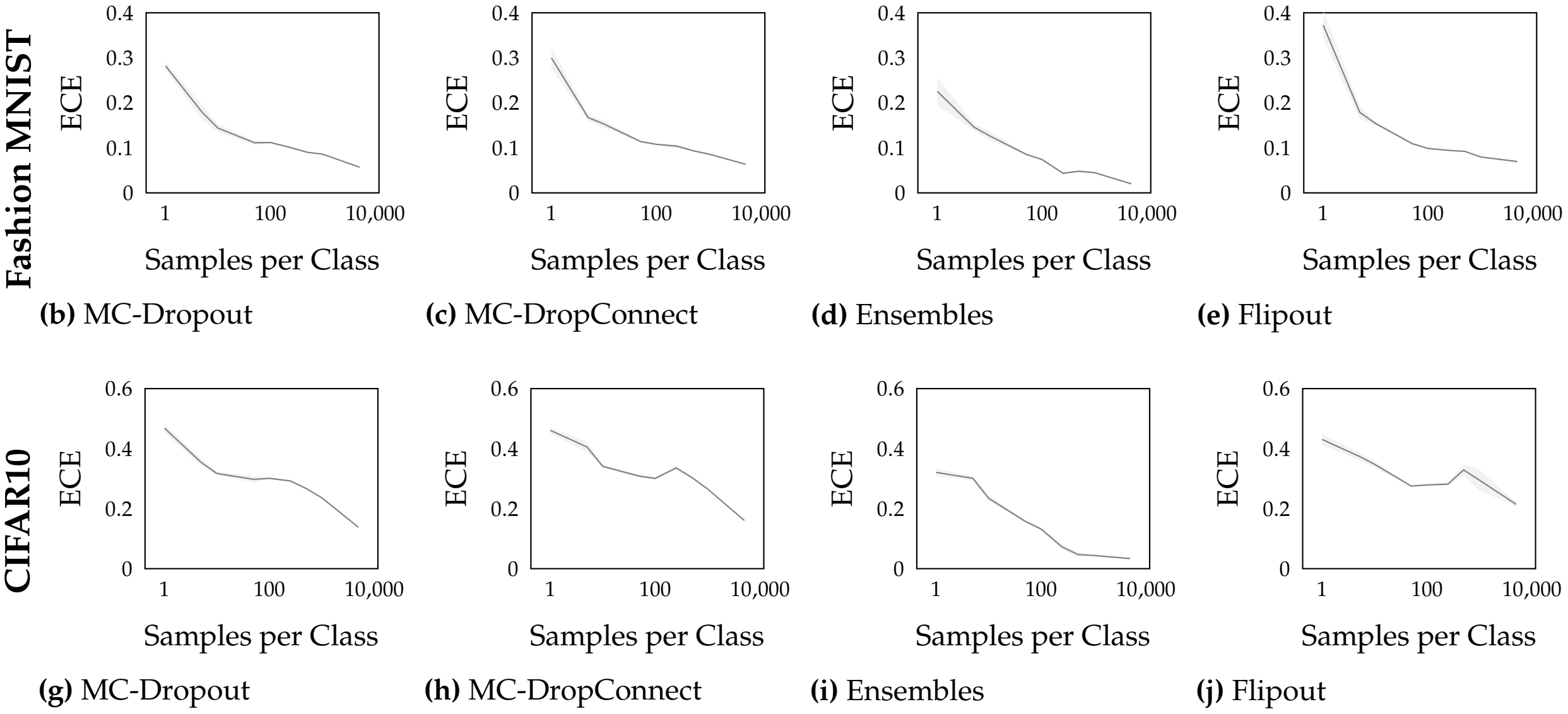
\begin{figure*}
	\subfloat{
		\rotatebox{90}{\textbf{Fashion MNIST}}
	}
	\subfloat[MC-Dropout]{
		\begin{tikzpicture}
			\begin{axis}[width = 0.25\linewidth, xlabel={Samples per Class}, ylabel={ECE}, ymajorgrids=false, xmajorgrids=false, grid style=dashed, tick label style={font=\tiny}, xmode=log, log ticks with fixed point, ymin=0.0, ymax=0.4, ytick style={draw=none}, xtick style={draw=none}, label style={font=\footnotesize}]
				
				\plotwitherrorareas{data/sampcomp/entropy-vs-SPC-fashion_mnist-results-miniVGG-dropout-combined.csv}{gray}{DE}{mean_ece}{std_ece}
				\legend{}
			\end{axis}		            
		\end{tikzpicture}
	}
	\subfloat[MC-DropConnect]{
		\begin{tikzpicture}
			\begin{axis}[width = 0.25\linewidth, xlabel={Samples per Class}, ylabel={ECE}, ymajorgrids=false, xmajorgrids=false, grid style=dashed, tick label style={font=\tiny}, xmode=log, log ticks with fixed point, ymin=0.0, ymax=0.4, ytick style={draw=none}, xtick style={draw=none}, label style={font=\footnotesize}]
				
				\plotwitherrorareas{data/sampcomp/entropy-vs-SPC-fashion_mnist-results-miniVGG-dropconnect-combined.csv}{gray}{DE}{mean_ece}{std_ece}
				\legend{}
			\end{axis}		            
		\end{tikzpicture}
	}
	\subfloat[Ensembles]{
		\begin{tikzpicture}
			\begin{axis}[width = 0.25\linewidth, xlabel={Samples per Class}, ylabel={ECE}, ymajorgrids=false, xmajorgrids=false, grid style=dashed, tick label style={font=\tiny}, xmode=log, log ticks with fixed point, ymin=0.0, ymax=0.4, ytick style={draw=none}, xtick style={draw=none}, label style={font=\footnotesize}]
				
				\plotwitherrorareas{data/sampcomp/entropy-vs-SPC-fashion_mnist-results-miniVGG-deepensemble-combined.csv}{gray}{DE}{mean_ece}{std_ece}
				\legend{}
			\end{axis}		            
		\end{tikzpicture}
	}
	\subfloat[Flipout]{
		\begin{tikzpicture}
			\begin{axis}[width = 0.25\linewidth, xlabel={Samples per Class}, ylabel={ECE}, ymajorgrids=false, xmajorgrids=false, grid style=dashed, tick label style={font=\tiny}, xmode=log, log ticks with fixed point, ymin=0.0, ymax=0.4, ytick style={draw=none}, xtick style={draw=none}, label style={font=\footnotesize}]
				
				\plotwitherrorareas{data/sampcomp/entropy-vs-SPC-fashion_mnist-results-miniVGG-flipout-combined.csv}{gray}{DE}{mean_ece}{std_ece}
				\legend{}
			\end{axis}		            
		\end{tikzpicture}
	}
	
	\subfloat{
		\rotatebox{90}{\hspace*{2em}\textbf{CIFAR10}}
	}
	\subfloat[MC-Dropout]{
		\begin{tikzpicture}
			\begin{axis}[width = 0.25\linewidth, xlabel={Samples per Class}, ylabel={ECE}, ymajorgrids=false, xmajorgrids=false, grid style=dashed, tick label style={font=\tiny}, xmode=log, log ticks with fixed point, ymin=0.0, ymax=0.6, ytick style={draw=none}, xtick style={draw=none}, label style={font=\footnotesize}]
				
				\plotwitherrorareas{data/sampcomp/entropy-vs-SPC-cifar10-results-miniVGG-dropout-combined.csv}{gray}{DE}{mean_ece}{std_ece}
				\legend{}
			\end{axis}		            
		\end{tikzpicture}
	}
	\subfloat[MC-DropConnect]{
		\begin{tikzpicture}
			\begin{axis}[width = 0.25\linewidth, xlabel={Samples per Class}, ylabel={ECE}, ymajorgrids=false, xmajorgrids=false, grid style=dashed, tick label style={font=\tiny}, xmode=log, log ticks with fixed point, ymin=0.0, ymax=0.6, ytick style={draw=none}, xtick style={draw=none}, label style={font=\footnotesize}]
				
				\plotwitherrorareas{data/sampcomp/entropy-vs-SPC-cifar10-results-miniVGG-dropconnect-combined.csv}{gray}{DE}{mean_ece}{std_ece}
				\legend{}
			\end{axis}		            
		\end{tikzpicture}
	}
	\subfloat[Ensembles]{
		\begin{tikzpicture}
			\begin{axis}[width = 0.25\linewidth, xlabel={Samples per Class}, ylabel={ECE}, ymajorgrids=false, xmajorgrids=false, grid style=dashed, tick label style={font=\tiny}, xmode=log, log ticks with fixed point, ymin=0.0, ymax=0.6, ytick style={draw=none}, xtick style={draw=none}, label style={font=\footnotesize}]
				
				\plotwitherrorareas{data/sampcomp/entropy-vs-SPC-cifar10-results-miniVGG-deepensemble-combined.csv}{gray}{DE}{mean_ece}{std_ece}
				\legend{}
			\end{axis}		            
		\end{tikzpicture}
	}
	\subfloat[Flipout]{
		\begin{tikzpicture}
			\begin{axis}[width = 0.25\linewidth, xlabel={Samples per Class}, ylabel={ECE}, ymajorgrids=false, xmajorgrids=false, grid style=dashed, tick label style={font=\tiny}, xmode=log, log ticks with fixed point, ymin=0.0, ymax=0.6, ytick style={draw=none}, xtick style={draw=none}, label style={font=\footnotesize}]
				
				\plotwitherrorareas{data/sampcomp/entropy-vs-SPC-cifar10-results-miniVGG-flipout-combined.csv}{gray}{DE}{mean_ece}{std_ece}
				\legend{}
			\end{axis}		            
		\end{tikzpicture}
	}
	\caption{Effect of number of training samples per class on test expected calibration error (ECE) over Fashion-MNIST and CIFAR10 datasets. Note how each uncertainty method has different patterns due to varying model uncertainty. Results from \cite{valdenegro2021exploring}.}
	\label{ece_vs_spc}
\end{figure*}

Figure \ref{ece_vs_spc} shows how calibration improves, by decreasing ECE, with increasing size of the training set, which also demonstrates how model uncertainty and calibration are related. Calibration error decreases as model uncertainty decreases.

\section{Methods for Calibration}

We have discussed methods to evaluate calibration, but now the question is, how can we improve the calibration of a trained model? Or train a better calibrated model?

For this there are two major types of methods:
\begin{description}
	\item[Post-Hoc] Modify or recompute predicted probabilities to improve their calibration.
	\item[A priori] Improve model structure or training to produce a better calibrated model.
\end{description}

\begin{definitionbox}{Calibration Maps}{calib_maps}
	A calibration map is a function $g: \mathbb{S}_\mathbb{Y} \rightarrow \mathbb{P}_\mathbb{Y}$, where the mapping is applied to the confidences produced by a model, the output is a new (better) calibrated probability, and the mapping $g$ is learned in a way that improves calibration of the predicted confidences.
	\begin{itemize}
		\item $\mathbb{S}$ is the uncalibrated confidence or probability space, this is produced by your model.
		\item $\mathbb{P}$ is the calibrated probability space on $\mathbb{Y}$, given by $\mathbb{P}_\mathbb{Y} = \{[s_0, s_1, ..., s_{C-1}] \, | \, s_i \geq 0, \sum_{i=0}^{C - 1} s_i = 1]\}$.
		\item $\mathbb{Y}$ is the categorical space for $C$ classes, given by $\mathbb{Y} = \{0, 1, ..., C - 1\}$
	\end{itemize}
\end{definitionbox}

\subsection{Platt Scaling}

Platt Scaling was developed to produce calibrated probabilities for Support Vector Machines (\sidecite{platt1999probabilistic}), but it can be used for any model. It uses the following calibration map:
\begin{equation}
	g(x) = \frac{1}{1 + \text{exp}(A x + B)}
\end{equation}
The parameters $A$ and $B$ are learned with gradient descent, using a proper scoring rule like the cross-entropy. The labels $y$ can be softened by:
\begin{equation}
	y_{+} = \frac{N_{+} + 1}{N_{+} + 2} \qquad y_{-} = \frac{1}{N_{-} + 2}
\end{equation}
Where the $_{+}$ indicates positive class, and $_{-}$ the negative class, and $N$ indicates the corresponding number of samples. Training is can be done using the original training set, or train on the validation set of the original model and then evaluate on a test set.

Note that this formulation is for binary classification, and Platt scaling can be extended to multiple classes or other forms of calibration described in Sec \ref{calib:sec_types}, just by applying the appropriate model like multi-class logistic regression for multiple classes and an appropriate loss function.

\subsection{Empirical Binning}

When we estimate any calibration error metric, we also need to compute bins $\text{conf}(B_i)$ (confidences) and $\text{acc}(B_i)$ (empirical accuracies), which are used to group predictions according to similar confidence. An observation is that the information in the calibration plot can also be used to improve calibration, as shown in \sidecite{zadrozny2001learning}.

If we use the following calibration map:
\begin{equation}
	g(x) = \text{acc}(B_i) \qquad \text{if } x \in B_i.
\end{equation}
Basically this method uses a piecewise constant function that for a given input confidences, it searches for the bin that contains that confidence value, and then it outputs the mean accuracy of that bin. This would definitely improve calibration of the model, because for the mean confidences present in the calibration plot, using their corresponding accuracy would mean zero calibration error for that confidence value, but now the possible confidences are discretized, not all possible confidence values can be produced.

This a non-parametric calibration method, as no parameters are estimated and the data in the calibration plot is used directly to improve calibration. While this method is expected to reduce calibration error, it will rarely reach a calibration error of zero.

\subsection{Isotonic Calibration}

Isotonic regression is a method for regression of monotonic (increasing) functions, and generally we can assume that calibration maps are monotonic or at least non-decreasing. This concept can be applied to calibration, as shown in \sidecite{zadrozny2002transforming}.

One disadvantage of empirical binning is that bins are often equally spaced and of constant size. A simple idea could also improve the effectiveness of the calibration method, what if we could adjust the bin edges?

In Isotonic calibration, we use the same setup as empirical binning, but now the bin edges $B_i = [l_i, u_i]$ with $l_i < u_i$, are learned from the data instead of being fixed values that are usually set from a decided number of bins $N$. The bin sizes do not have to be equal.

This allows bins to adapt and better fit the true calibration map. This also a non-parametric calibration method. 

\subsection{Temperature Scaling}

Another way to improve calibration is to fiddle with the logits, which are the inputs to softmax or sigmoid functions that produce output probabilities.

\pgfmathdeclarefunction{tempSoftmax}{1}{%
	\pgfmathparse{1 / (1 + exp(-x/#1))}%
}

The softmax activation function also has a version that uses a temperature $T$ to scale the logits:
\begin{equation}
	\text{softmax}(\mathbf{x}, T) = \left[\frac{\text{exp}(\mathbf{x}_i / T)}{\sum_j \text{exp}(\mathbf{x}_j / T)}\right]_i \quad \forall i \in [0, C - 1]
\end{equation}
\begin{marginfigure}
	\begin{tikzpicture}				
		\begin{axis}[every axis plot post/.append style={mark=none,domain=-9:9,samples=50,smooth}, 
			axis x line*=bottom, axis y line*=left, enlargelimits=upper,
			ylabel={Probability}, xlabel={Logits}, width=\linewidth, legend to name=named, legend columns=2] 
			\addplot+[red] {tempSoftmax(0.1)};
			\addlegendentry{$T = 0.1$}
			\addplot+[green] {tempSoftmax(0.5)};
			\addlegendentry{$T = 0.5$}
			\addplot+[blue] {tempSoftmax(1.0)};
			\addlegendentry{$T = 1.0$}
			\addplot+[black] {tempSoftmax(2.0)};
			\addlegendentry{$T = 2.0$}
			\addplot+[gray] {tempSoftmax(4.0)};
			\addlegendentry{$T = 4.0$}
		\end{axis}				
	\end{tikzpicture}
	\begin{center}
		\ref{named}
	\end{center}
	\caption{Plot of the Sigmoid Activation with Temperature, as an easy way to visualize how the Softmax with Temperature behaves (both have the same behavior).}
	\label{softmax_temp}
\end{marginfigure}
The logits $x_i$ are divided by a temperature factor $T$. This has the effect of softening the output distribution for increasing $T$, and can improve calibration as the labels become softer instead of closer to one-hot encoded (also called hard labels). This effect is shown in Figure \ref{softmax_temp}.

A model can be re-trained to optimize the $T$ parameter, using another loss such as the expected calibration error (with fixed
bins), in order to improve calibration. This would require to use non-gradient methods since the ECE is not differentiable. Empirical confidences can also be used as targets, or fitting to a previously computed calibration plot.

This can be done globally for all classes, with a common $T$ , or to have a $T_i$ specific for each class:
\begin{equation}
	\text{softmax}(\mathbf{x}, \mathbf{T}) = \left[\frac{\text{exp}(\mathbf{x}_i / T_i)}{\sum_j \text{exp}(\mathbf{x}_j / T_i)}\right]_i \quad \forall i \in [0, C - 1]
\end{equation}
Where now $\mathbf{T}$ is a vector of length $C$ that is optimized to improve model calibration.

One interpretation of per-class temperatures $T_i$ is class imbalance, as each $T_i$ will scale to counteract class imbalances and adjust confidences accordingly.

\subsection{Soft vs Hard Labels}

Classification models are often trained on one-hot encoded probabilities, meaning that target probabilities only have values of zero or one\sidenote{For example $[0, 1]$ and $[1, 0]$ for binary classification.}, which is not a good representation of probabilities that lead to calibrated models. We have previously mentioned that this is one reason why standard models are badly calibrated, since the model learns only to produce zero or one output probability values. These kind of target probabilities are called hard labels.

But there is an alternative, soft labels, meaning that we expect labels to not be produced from a one-hot encoding, and instead be softer\marginnote{For example, in a 3-class setting, a one-hot encoded label could be $[1, 0, 0]$, and a soft label could be $[0.8, 0.1, 0.1]$.}, meaning that probability mass is distributed over all classes, but still the highest probability is assigned to the correct class.

It has been shown, for example by \sidecite{maher2021instance}, that soft labels improve model calibration, due to the model being trained to produce probabilities values in the full range $[0, 1]$ instead of just seeing 0's and 1's during training, which reduces overconfidence and improves overall calibration.

One major question is how to produce soft labels? One alternative is a method called label smoothing (\sidecite{szegedy2016rethinking}), where for each label instance in the training set, a new probability vector $q$ is computed according to:
\begin{equation}
	q(i) = (1 - \epsilon) \mathrm{1}[i = k] + \frac{\epsilon}{C}
\end{equation}
Where $C$ is the number of classes, $k$ is the index of the correct class for that label instance, and $\epsilon$ is the amount of probability mass to distribute among the non-correct classes, with a typical value $\epsilon = 0.1$ This method conceptually distributes a small part ($\epsilon$) of the probability mass to other classes.

Another option for soft labels is to use multiple annotations, and infer the soft class distribution from those annotations. This method is more expensive as it requires multiple annotations but it is more profitable, specially if human annotators are used, then class probabilities encode human ambiguity and a model has the chance to learn this ambiguity. An example of this scheme is the FER+ dataset for facial emotion recognition (\sidecite{barsoumICMI2016}).

\subsection{Mixup Training}
One problem with label smoothing is that the label are changed to be soft, but the features are not changed. Another method to produce soft labels is mixup training, which does produce soft labels and makes changes to the features.

The concept of Mixup training (\sidecite{thulasidasan2019mixup}) is to train not only on training samples, but also in their neighborhoods. It is basically a data augmentation method, where intermediate samples $(\tilde{x}, \tilde{y})$ are produced by linear interpolation of neighboring instances in feature space:
\begin{align}
	\tilde{x} &= \lambda x_i + (1 - \lambda) x_j\\
	\tilde{y} &= \lambda y_i + (1 - \lambda) y_j
\end{align}
Where $x_i, x_j$ are pairs of training samples (features, images, etc) with one randomly selected and the other being a neighbor, and $y_i, y_j$ are their corresponding one-hot encoded labels. $\lambda$ is the mixing factor which is sampled from $\lambda \sim \text{Beta}(\alpha, \alpha)$ , where $\alpha \in [0, \infty]$ is a hyper-parameter, usually set to $[0.1, 0.4]$.

Since mixup training does linear interpolation of both features and labels, and it considers if instances are neighboring in feature space, it has an effect of label smoothing but that is instance-dependent. Experimental results show that mixup training improves calibration and task performance, but with increased training cost due to the data augmentation effect.

\section{Calibration for Regression}

So far all calibration methods we have covered are implicitly formulated for classification. But there is a simple transformation that allows to apply these methods for regression.

Since regression does not directly use probabilities, using a standard deviation $\sigma$ to represent uncertainty, instead a calibration map can be applied to the quantiles $\tau$ of the predicted distribution, which are also in the range $\tau \in [0, 1]$.

A more complex approach, depending on the method, is to apply a calibration map to the standard deviation directly, taking into consideration that the standard deviation is positive but unbounded (no maximum value), and the calibration map design should take this into consideration.

\section{Challenges}

The biggest challenge for any machine learning method, including calibration, is that of generalization, which is also the problem when making predictions, generalization is the biggest problem in machine learning.

Generally speaking, that a model/method is well calibrated in one dataset, does not mean that it will be well calibrated in other datasets, particularly a test set, or datasets very far from the training distribution.

There are no guarantees of adequate performance in the test set. This is clearly exemplified in recent research, where \sidecite{ovadia2019can} evaluated different uncertainty estimation methods across input data corruptions in the test set. It is expected that input feature corruptions act as a form of dataset shift or out of distribution data, and then calibration performance should stay constant, while accuracy will decrease due to the model failing to generalize on corrupted data.

\begin{figure}
	\begin{tikzpicture}				
		\begin{axis}[axis x line*=bottom, axis y line*=left, ylabel={Expected Calibration Error}, xlabel={Corruption Severity Level}, 	width=\linewidth, height=0.3\textheight, legend columns=2, xtick={1,2,3,4,5}, ytick={0.0, 0.1, 0.2, 0.3, 0.4}, legend pos=north west] 
			\addplot[green, only marks, error bars/.cd, y dir=both, y explicit] coordinates { (1, 0.03) +- (0, 0.02) (2, 0.02) +- (0, 0.03) (3, 0.04) +- (0, 0.05) (4, 0.05) +- (0, 0.06) (5, 0.09) +- (0, 0.08) };
			\addlegendentry{Ensemble}
			
			\addplot[red, only marks, error bars/.cd, y dir=both, y explicit] coordinates { (1.1, 0.03) +- (0, 0.05) (2.1, 0.05) +- (0, 0.1) (3.1, 0.09) +- (0, 0.15) (4.1, 0.12) +- (0, 0.25) (5.1, 0.18) +- (0, 0.28) };
			\addlegendentry{Temp Scaling}
			
			\addplot[blue, only marks, error bars/.cd, y dir=both, y explicit] coordinates { (1.2, 0.02) +- (0, 0.015) (2.2, 0.035) +- (0, 0.05) (3.2, 0.08) +- (0, 0.08) (4.2, 0.11) +- (0, 0.3) (5.2, 0.15) +- (0, 0.32) };
			\addlegendentry{Dropout}
		\end{axis}				
	\end{tikzpicture}
	\caption[Reproduction of experimental results from \cite{ovadia2019can}]{Reproduction of experimental results from \cite{ovadia2019can}, where a test set is corrupted with varying severity (1-5). These results show how ECE degrades with increasing corruption intensity, showing that uncertainty estimation can be unreliable in a heavily out of distribution setting. Note how ensembles has the smallest increase in ECE.}
	\label{ece_ovadia}
\end{figure}
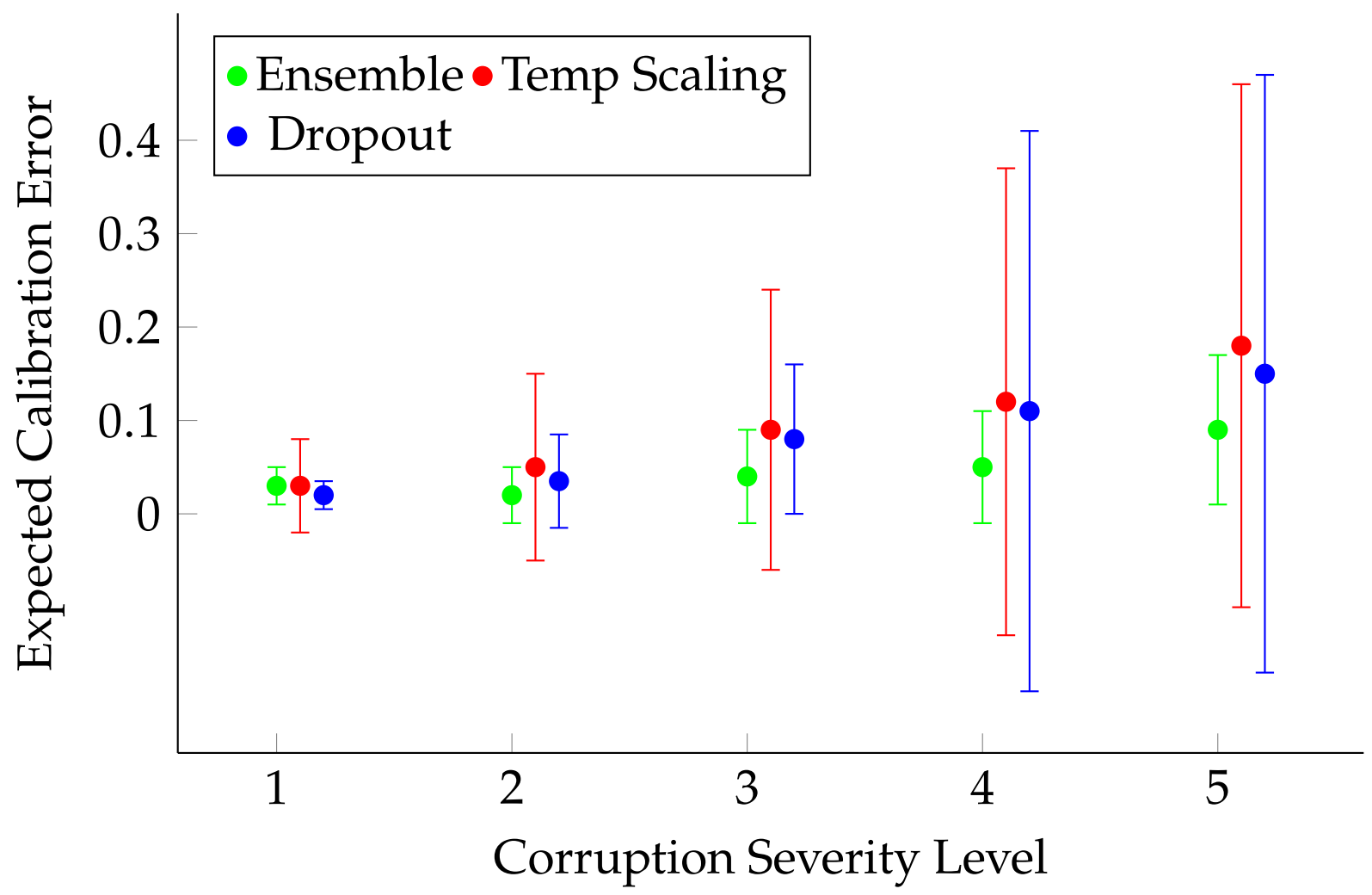

But Figure \ref{ece_ovadia} shows that our expectations are not fulfilled experimentally, on several datasets and over several uncertainty estimation methods, ECE actually increases, and the increase is larger with more severe corruptions. These results clearly indicate that uncertainty estimation methods fail in severe dataset shift and out of distribution settings and then confidence no longer works as a proxy for prediction accuracy.

%% file: sections/ood-detection.tex
\chapter{Out of Distribution Detection} 
\label{ch:ood_detect}
\margintoc

\section{Introduction}

In this book we have extensively explored that confidence should be proportional to error or be a probability of correct classification, in order for a human user to decide if a prediction is useful or not. These are fundamental problems for the use of machine learning models in human applications.

A big problem are model limitations. If a model is given an input that was not trained on, it will answer incorrectly, giving always what is popularly known a "hallucinations". This is particularly more true for large models, specially as we do not know their training sets or the contents of them. A model giving a useful calibrated confidence is able to say "I do not know" instead of providing a incorrect answer.

For applications in Medicine, Autonomous Driving, and other applications involving humans, knowing the limitations of machine learning models is crucial.

The question is then, how to detect when a model is given an input that are very different from its training set? Without knowing the contents or size of the training set, this has to be done by the model itself, the same way a human will say "I do not know" when asked a question it cannot answer.
This is the concept of Out of Distribution Detection (OOD Detection) that we will explore in this chapter.

Note that in the machine learning literature, there is also the concept of Out of Distribution generalization, which aims to build a robust machine learning model that can generalize in (some) out of distribution settings, but fundamentally these are different approaches to a similar problem, detection and generalization are not the same problem and the approaches are very different.

\subsection{Model Limitations}

We start with concepts on what are model limits or limitations. Every machine learning model, independent of the size of its training set, has limitations that often are not known a priori. These limitations are always present, due to multiple reasons, like gaps in the training set, limited size of the training set, or simply incorrectly labeled data points.

\begin{figure}[t]
	\centering
		\begin{tikzpicture}
			\begin{axis}[height = 0.20 \textheight, width = \textwidth, xlabel={}, ylabel={}, xmin = -6.28, xmax = 6.28, grid style=dashed, legend pos = north east, legend style={font=\scriptsize}, tick label style={font=\scriptsize}, ymin =-2.0 , ymax=2.0, trig format plots=rad, xtick=\empty, ytick=\empty]
				
				\draw[fill=gray!25,opacity=0.5]  (-3.1415, -2.0) -- (3.1415, -2.0) -- (3.1415, 2.0) -- (-3.1415, 2.0) -- cycle ;
				
				\draw[solid, black] (0.0, 1.25) node[above] {Training Set};
				
				\addplot[name path=upper, draw=none] table[x = x, y expr = \thisrow{pred_mu} + \thisrow{pred_sigma}, col sep = semicolon] {data/toy-regression-15ensembles-disentangled.csv};
				\addplot[name path=lower, draw=none] table[x = x, y expr = \thisrow{pred_mu} - \thisrow{pred_sigma}, col sep = semicolon] {data/toy-regression-15ensembles-disentangled.csv};
				
				\addplot[mark=none, black] table[x = x, y = pred_mu, col sep = semicolon] {data/toy-regression-15ensembles-disentangled.csv};
				
				\addplot [fill=darkgray!50, area legend] fill between[of=upper and lower]; 
				
				\addplot[mark=none, red, domain=-6:6, smooth] {sin(x)};
				
				\draw[dashed, gray] (3.1415, -2) -- (3.1415, 2);
				\draw[dashed, gray] (-3.1415, -2) -- (-3.1415, 2);
				
				\draw[to-to, solid, black] (-3.1415, -1.7) -- (3.1415, -1.7) node[midway, above] {Interpolation};
				\draw[to-to, solid, black] (-6.2, -1.7) -- (-3.1415, -1.7) node[midway, above] {Extrapolation};
				\draw[to-to, solid, black] (3.1415, -1.7) -- (6.2, -1.7) node[midway, above] {Extrapolation};
			\end{axis}		
		\end{tikzpicture}
		\caption[Conceptual display of model interpolation and extrapolation with uncertainty for a simple regression setting.]{Conceptual display of model interpolation and extrapolation with uncertainty for a simple regression setting. The  shaded area represents the training set, while the other two areas represent data where the model has not been trained. The black curve is the mean prediction, and the red curve is the true value.}
		\label{fig:interp_extrap_uncertainty}
\end{figure}
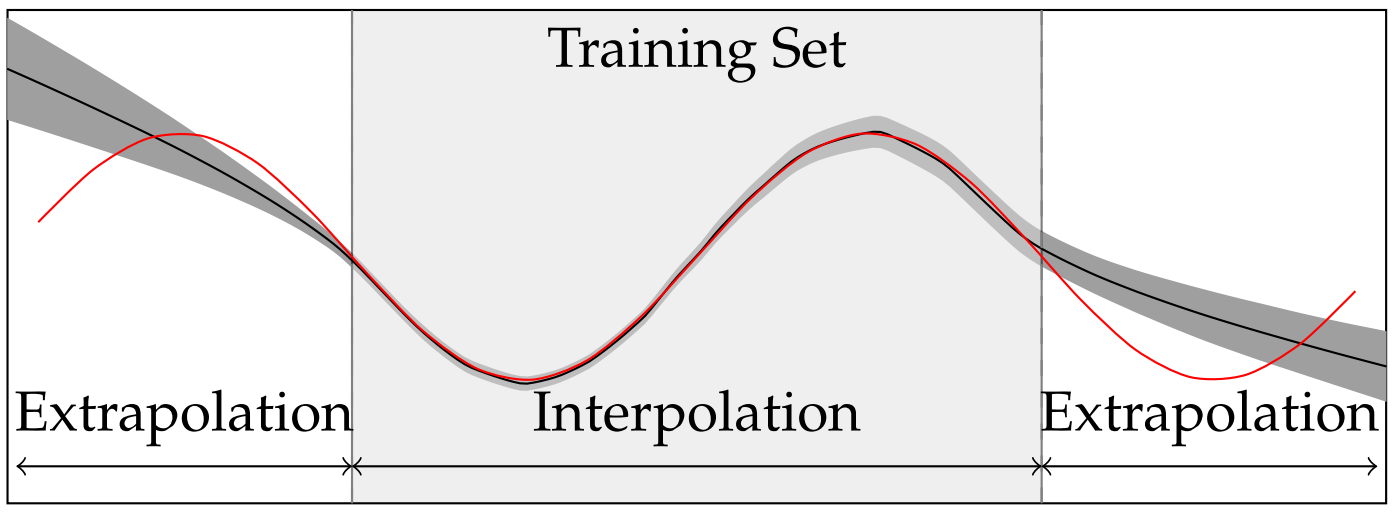

Figure \ref{fig:interp_extrap_uncertainty} presents a simple regression setting with uncertainty, with the training set being the shaded area. In this region the model performs very well with input values in this area, but outside the shaded area, the model has not been trained, so it will never make correct predictions. This is signaled by the predictive uncertainty, as the red and black curves diverge and there is increasing degrees of uncertainty.

The effect inside the training set is called interpolation, as the model is interpolating between values it has seen during its training, while outside of the training set, the effect is called extrapolation, as the model can only interpolate its knowledge far from where it has been trained, effectively extrapolating, and this case is the most interesting one as it is the desired model ability to generalize, but very far from the training set the model will never make correct predictions as it is stretching its knowledge.

\begin{definitionbox}{Model Interpolation and Extrapolation}{interpolation_extrapolation}
	When a model is given an input that is similar to what it learned from the training set, then the model is \textbf{interpolating}, and generally this works very well using neural networks.\\
	
	When a model is given input that is dissimilar or outside of its training set, then it \textbf{extrapolates}, basically guessing, and this is the situation where failure is almost guaranteed, and needs to be detected at inference time.
\end{definitionbox}

The whole concept of out of distribution detection corresponds to detecting when the model is interpolating (input is inside the training set) versus when it is extrapolating (input outside the training set), and how far the model's limits are being tested when given an input, corresponding to measuring a conceptual distance from the training set limits to the input sample.

\begin{definitionbox}{Common Model Limitations}{model_limits}
	\begin{description}[\compact\setlabelphantom{Taskfo}]
		\item[Training Set Size] This limits the model through diversity in data points used for training. The real world is very complex and datasets only approximate this complexity, but smaller training sets are more limited than large ones.
		\item[Gaps in Training Set] Training sets can be more limited than just number of samples, they can have gaps, meaning missing interclass or intraclass diversity, missing particular kinds of samples, or be biased towards certain populations.
		\item[Sample Difficulty] Not all samples in the training set are equally difficult, and there can be samples that are consistently incorrectly predicted by a model, even in the training set.
		\item[Task] A model is trained for a specific tasks or sets of tasks, and it can only perform well on those tasks. A model will not make correct predictions for a task that it was not explicitly trained on.
		\item[Assumptions] Models have specific structures and they make assumptions on data, including how the model was trained, degrees of under or overfitting, etc.
	\end{description}
	The overall idea is that (model) uncertainty signals these limitations.
\end{definitionbox}

\subsection{Model vs World Knowledge}

To train machine learning models , we use data as a proxy for the real world, but this creates a mismatch between the contents of the training set, what the models learns, and the real world. It is possible to assess knowledge between the model and the world, but first we define some assumptions. The most common assumption made when training a machine learning model is the closed set assumption.

\begin{definitionbox}{Closed Set Assumption}{closed_set_assumpt}
	The training, validation, and test set contains a pre-defined set of classes (classification) or a distribution of values (regression), and no other distribution or classes exist. For example for model trained on cats and dogs, the model (implicitly) makes the assumption that no other classes exist in the world.
\end{definitionbox}

If a model is trained without the closed set assumption, it is said to follow the open set assumption, also known as open world. This setting is considerably more difficult than the more common closet set assumption, as the model has to correctly predict the known classes, but also predict when a data point is of an unknown class. The set of unknown classes is unbounded.

\begin{definitionbox}{Open Set Assumption}{open_set_assumpt}
	Also known as open world assumption or setting, this is a model that makes the following two assumptions:
	\begin{description}
		\item[Known Classes] There are classes that the model was trained on and are known to the model.
		\item[Unknown Classes] There are more classes that the model is not trained on, but still exist and can appear at inference time. Ideally, the model can recognize these classes as "unknown".
	\end{description}			
\end{definitionbox}

These concepts can be generalized in matrix form as:

\begin{table}[h]
	\begin{tabular}{rp{0.35\linewidth}p{0.35\linewidth}}
		\toprule
		Model/World     & Known 			& Unknown\\
		\midrule
		Known 			& \textbf{Known Knowns}.		& \textbf{Known Unknowns}.\\
						& Closed set assumption. 		& Open world assumption. \\
		Unknown			& \textbf{Unknown Knowns}. 		& \textbf{Unknown Unknowns}. \\
						& Model is unaware of its true world knowledge. & Model is unaware of its lack of knowledge about the world.\\
		\bottomrule
	\end{tabular}
	\caption{Model vs World Awareness matrix.}
\end{table}

And in more detail:

\textbf{Known Knowns}. Basic and only knowledge in the closed set assumption.

\textbf{Known Unknowns}. The model is aware that there is knowledge that is unknown, for example,
there might be more classes than the ones used in the training set. This is the basic open set recognition setting. These can also be seen as risks that the model is aware of.

\textbf{Unknown Knowns}. The model is unaware of some knowledge it actually knows. This relates more to the designer of the model than the model itself, for example, Transformers have properties that were there, but we were not aware of them.

\textbf{Unknown Unknowns}. This is the most difficult setting, as there is knowledge that is unknown,
and the model is completely unaware of its lack of knowledge. The open set recognition setting partially covers this case, as there are possibly infinite classes, and it is not possible to enumerate them all, thisr elates to the frame problem.\marginnote{The frame problem is the problem to enumerate all effects of an action in an environment, which is often not possible.}

\clearpage
\section{Distribution Shifts}
Related to the topic of out of distribution detection, is the concept of distribution shift, where the test set distribution shifts relative to the training set distribution, and this happens commonly in real-world use of machine learning models. There are three forms of distribution shift.

\begin{marginfigure}
    \subfloat[Train Set]{
    	\parbox{\linewidth}{
    		\centering
        \includegraphics[width=0.4\linewidth]{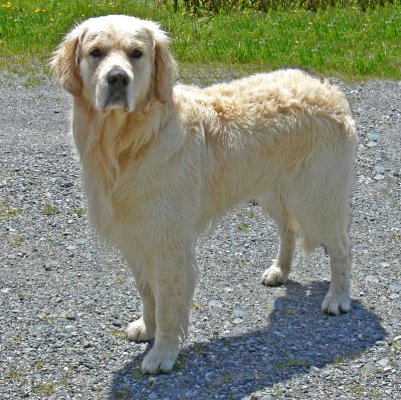}        
        \includegraphics[width=0.4\linewidth]{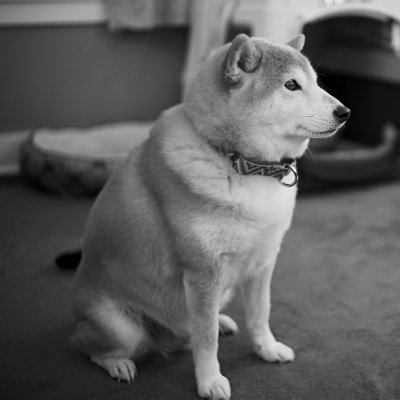}\\
        \vspace*{0.5em}
        \includegraphics[width=0.4\linewidth]{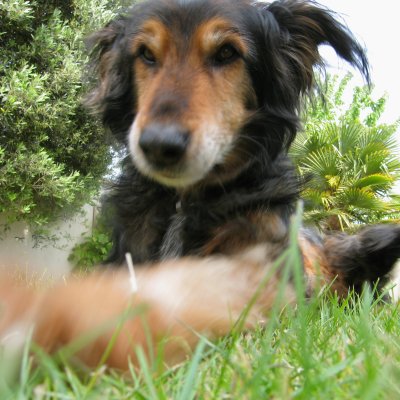}
    	}
    }
    
    \subfloat[Test Set]{
    	\parbox{\linewidth}{
    		\centering
        \includegraphics[width=0.4\linewidth]{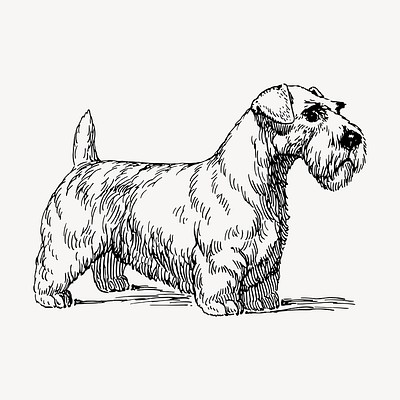}        
        \includegraphics[width=0.4\linewidth]{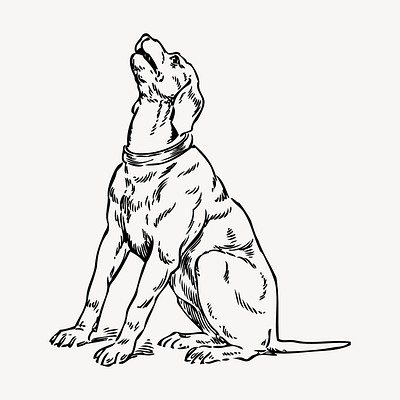}\\
        \vspace*{0.5em}
        \includegraphics[width=0.4\linewidth]{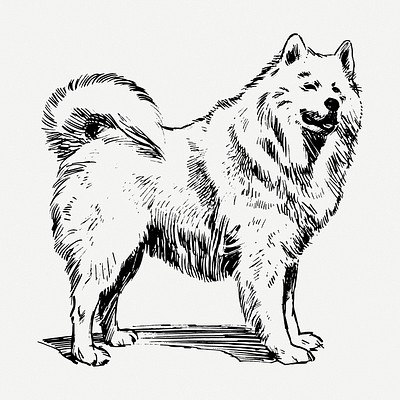}
    	}
    }
    \caption[Example of feature shift with natural vs drawings of dogs.]{Example of feature shift, where the training and test sets both have images of dogs, but the training set contains natural dog photos, while the test set contains dog drawings. It is expected that a model trained on this training set would fail to make correct predictions on this test set. Images under CC-zero license/public domain, sourced from \href{Openverse}{https://openverse.org}}
    \label{ood_feature_shift_dogs}
\end{marginfigure}

\begin{definitionbox}{Feature Shift}{feature_shift}
    This is also known as covariate shift, where covariate is another word for feature used in statistics.\\
    
    Feature shift is when the input feature distribution $X$ changes, for example, new semantic classes might be present, but the model is not aware of these classes. Or currently trained classes are the same, but are visually very different from training set samples for the same class. This example is shown in Figure \ref{ood_feature_shift_dogs}.\\
    
    Basically, the feature distribution changes, but the label distribution is the same, and the model has not been retrained.
\end{definitionbox}

\begin{definitionbox}{Label Shift}{label_shift}
	The feature distribution $X$ is the same, but the label distribution $Y$ changes.\\
    	
	This can be due to a change in the domain that generates data and labels.	
	For example, in a medical setting, we know that disease generates symptoms, and symptoms are measured to infer disease.But these causal relationships are not the same across the world, or not constant due to advances in medical science.\\
    
    For example if you had a cough and cold-like symptoms, at this current time, it would not be a big problem, but if you had the same symptoms during the COVID-19 pandemic, then your first thought is that you could be infected with a Coronavirus.\\
    
    Similarly if you get bitten by a mosquito in your summer vacation around Europe, you would not worry too much, but if you get bitten by a mosquito in some African regions, you could get worried as it can imply a Malaria infection.
\end{definitionbox}

\begin{definitionbox}{Concept Shift}{concept_shift}
	Another possible way for feature/labels to shift is via concept, where the meaning of labels might change over time or over geographical regions.\\
	
	A very good example is the name for concepts, like soda, soft drink, pop, or coke, mean the same over different regions.\\
	
	Brand names can also differ around the world, Aldi is the same in The Netherlands, but in Germany there is Aldi Nord and Aldi Süd. In Chile there is a supermarket chain called Jumbo, but is different from the Jumbo in The Netherlands.
\end{definitionbox}
The big issue with distribution shifts is that models are not prepared for the test distribution to shift, and output uncertainty estimation should reflect the shift with a higher uncertainty.

Distribution shifts are very important since they happens in reality when deploying any machine learning model. You train a model on a dataset, but this dataset is static and fixed in time, while the world is dynamic and always changing. The model performance will not be static, it will continuously drop over time and at some point predictions will become basically random.
\begin{figure}[ht]    
    \begin{tabular}{p{0.135\textwidth}p{0.135\textwidth}p{0.135\textwidth}p{0.135\textwidth}p{0.135\textwidth}p{0.135\textwidth}}
        \textbf{Original}& \textbf{Severi. 1} & \textbf{Severi. 2} & \textbf{Severi. 3} & \textbf{Severi. 4} & \textbf{Severi. 5}\\
        \includegraphics[width=\linewidth]{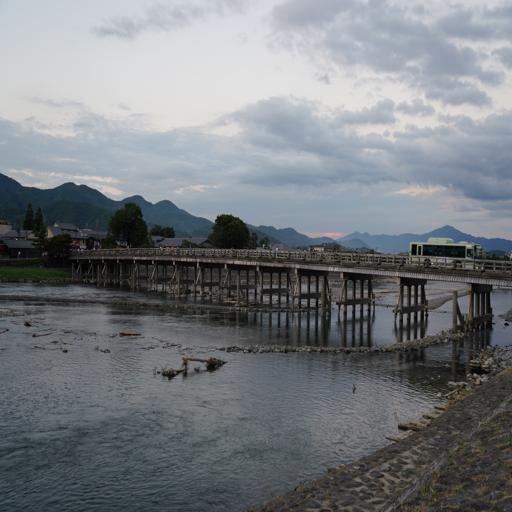} & \includegraphics[width=\linewidth]{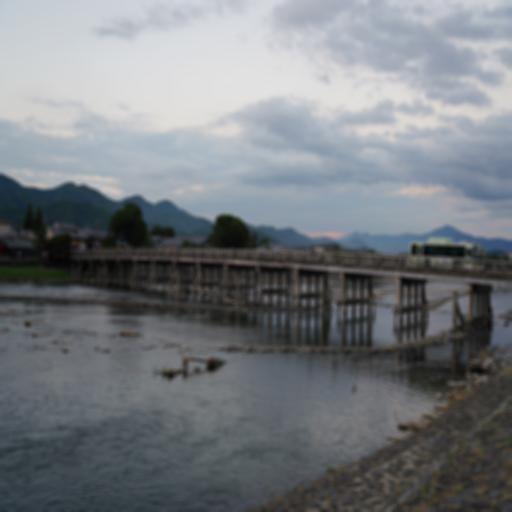} & \includegraphics[width=\linewidth]{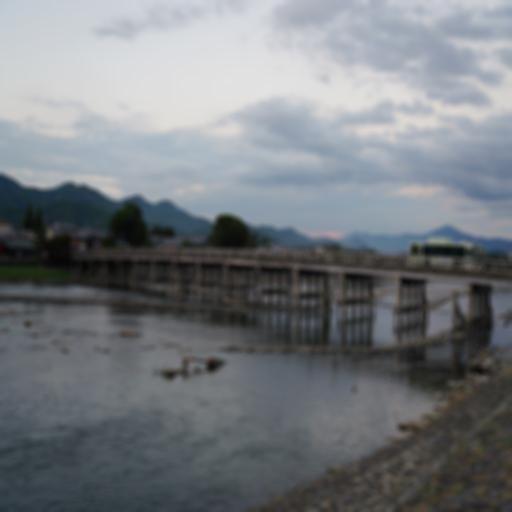} & \includegraphics[width=\linewidth]{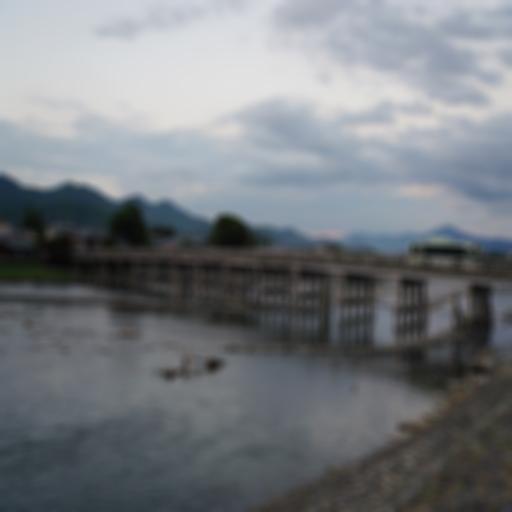} & \includegraphics[width=\linewidth]{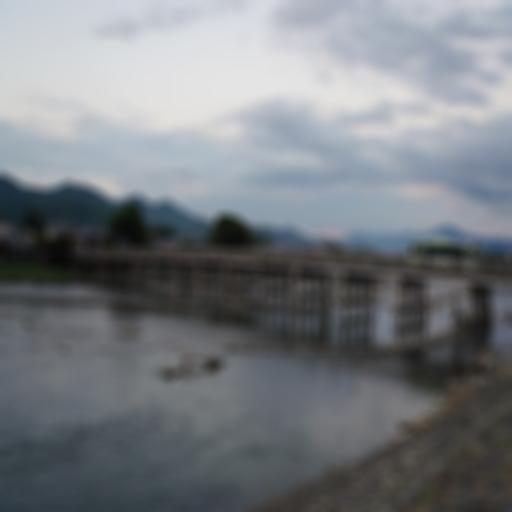} & \includegraphics[width=\linewidth]{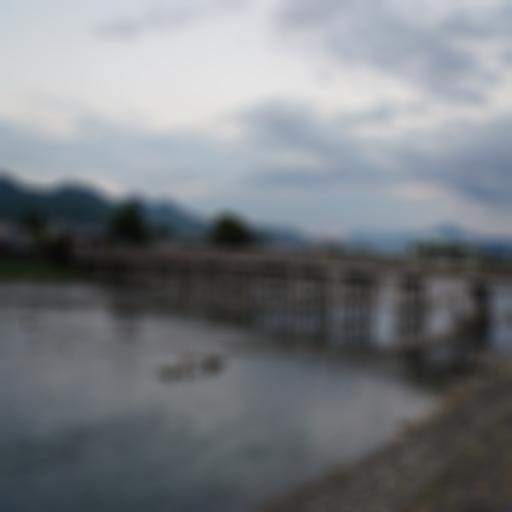}\\
        \footnotesize{Togetsukyo Bridge in Arashiyama}. & \footnotesize{Kyoto, Japan}  & \footnotesize{Kyoto, Japan} & \footnotesize{Kyoto, Japan} & \footnotesize{Lake Como in Italy} & \footnotesize{Lake} \\
        \textbf{Conf:} 90\% & \textbf{Conf:} 80\% & \textbf{Conf:} 75\% & \textbf{Conf:} 70\% & \textbf{Conf:} 75\% & \textbf{Conf:} 60\%\\
    \end{tabular}
    \caption[Example of feature distribution shift via image corruption from \cite{borszukovszki2025know}]{Example of feature distribution shift via image corruption from \cite{borszukovszki2025know}. The Vision-Language Model Claude is prompted to answer a question while the Input image has Defocus Blur Corruption. \textbf{Question}: Where was this photo taken? \textbf{Correct Answer}: Japan, Kyoto, Arashiyama Area, the Bridge is named Togetsu-kyo Bridge. It is clear how answers and confidence degrade with increasing corruption severity, and confidence does not signal answer correctness.}
    \label{fig:claude_answers_corrupted}
\end{figure}

\section{Out of Distribution Task Definition}

Out of distribution detection is the process of rejecting input samples with labels that do not overlap with the training set (label shift), or feature values that are outside the training set.

But more generally, it is the process of rejecting any input that is dissimilar with the training set. The training set of your model defines what is out of distribution \sidecite{yang2024generalized}.

OOD detection requires accurate epistemic uncertainty, since a model cannot make correct predictions outside its training set, and OOD inputs are expected to fall outside the model's knowledge.

The basic way to formulate OOD detection as a machine learning task, is to see it as a binary classification problem\marginnote{OOD Detection as Binary Classification}, where label $y = 0$ indicates an in-distribution input, and label $y = 1$ indicates an out of distribution input.

The binary classification decision can be based on the input $x$ itself, intermediate model features, etc, but it should be influenced somehow on the input features, as shown in $g(x, f)$ below.
\begin{equation}
	\label{eq:ood_score}
	g(x, f) = 
	\begin{cases} 
		0 		& \text{if } s(x, f) < \lambda \quad \text{In-Distribution Input}\\
		1       & \text{if } s(x, f) \geq \lambda	  \quad \text{Out-of-Distribution Input}
	\end{cases}
\end{equation}
Methods for out of distribution detection can be (learning) models themselves, or they can be non-learning formulas or equations based on some criteria. They usually use a scoring function $s$ that produces a score of OOD-ness,\marginnote{Note that some methods use the opposite direction, lower values being more likely to be OOD inputs.} with higher values meaning more likely to be an OOD input, but not necessarily being a probability or a bounded value.

Some basic desiderata for out of distribution detection methods are:
\begin{enumerate}
	\item OOD methods should be easy to apply to a trained model.
	\item OOD methods should not require model retraining.
	\item Good OOD performance should not harm ID performance. 
\end{enumerate}
We now give basic definitions in this topic:
\begin{marginfigure}
	\input{figures/trainset_holes.tikz}
	\caption{Example of a training set (shaded area) with holes in the distribution density in feature space, indicating that semantic or class variability samples are missing from the data.}
	\label{trainset_holes}
\end{marginfigure}
\begin{definitionbox}{In-Distribution Data}{id_data}
    These are the samples and semantic meaning implied by the training set.\\
    
    For example, if our training set has dogs and cats, in distribution would be the specific semantic meaning of the dogs and cats in the training set, which could be generated from particular species of dogs and cats, with specific background, animal pose, and illumination conditions.\\
    
    Note that training sets can have "holes" in feature space, as shown in Figure \ref{trainset_holes}. Since data defines the training set distribution, any instances missing from data can make these holes, for example on the cat/dogs example, missing breeds can form these holes.
\end{definitionbox}

\begin{definitionbox}{Out of Distribution Data}{ood_data}
    Any sample that is not inside the In-Distribution set.\\
    
    Example given our cats/dogs dataset example, OOD would be other
    semantic classes (foxes, ducks, humans, houses, etc) that are not part of the training set.
    
    Additionally samples of dogs and cats that are visually far from the dogs/cats in the training set would also be out of distribution, for example if the training set contains a particular set of breeds of dogs/cats, the missing breeds would be out of distribution.
\end{definitionbox}
Out of distribution data has a broad definition, and not all OOD samples are equal, there is the concept of Far OOD and Near OOD samples, depending on the feature space distance to the in-distribution samples, The definition below makes a formal definition and provides examples.

\begin{definitionbox}{Far vs Near OOD Samples}{far_near_ood}
	Not all OOD samples are equally difficult, this is where the concept of near and far out of distribution samples appears. This is drawn from \cite{yang2022openood}.\\
    
    \textbf{Near OOD samples} are the ones close to the training set, likely close to the boundary between the training set and any out of distribution samples. Near OOD samples are difficult to detect since they can be easily confused with In-Distribution samples.\\
    
    \textbf{Far OOD samples} are the ones far from the training set, far from the boundary. Far OOD samples are very easy to detect as they are very different from the training set.
\end{definitionbox}
\begin{marginfigure}
	\scalebox{0.8}{
		\input{figures/id_ood_difficulty.tikz}
	}
	\caption{Difficulty in the ODD detection task when considering Near and Far OOD samples. This figure shows the feature space, with the hatched area being the training set.}
\end{marginfigure}
\section{Evaluation in OOD Detection}
\begin{examplebox}[label=ood_dataset_examples]{Examples of Commonly Used Datasets for OOD Research}{ood_dataset_examples}
	For basic research in OOD detection methods, there are commonly used ID-OOD dataset combinations, like MNIST vs Fashion MNIST, CIFAR10 vs SVHN, ImageNet vs ImageNet-O, CIFAR-100 vs CIFAR-10, ImageNet vs iNaturalist, etc.\\
	
	There are some datasets used for OOD research that contain their own OOD splits, for example Camelyon17-WILDS (\cite{koh2021wilds}).
\end{examplebox}

\begin{table}
	\begin{tabular}{lp{4cm}p{4cm}}
		\toprule
		\textbf{ID Dataset} & \textbf{Near OOD Datasets} & \textbf{Far OOD Datasets}\\
		\midrule
		MNIST		& Not-MNIST			& Kuzushiji-MNIST, Fashion-MNIST\\
		CIFAR10		& CIFAR100,\newline TinyImageNet & MNIST, Fashion-MNIST, Texture, Places365\\
		ImageNet	& iNaturalist, ImageNet-O, OpenImages-O & Texture, MNIST, SVHN\\
		\bottomrule
	\end{tabular}
	\caption[Examples of In-Distribution datasets and their commonly used Near and Far OOD datasets]{Examples of In-Distribution datasets and their commonly used Near and Far OOD datasets, based on information from \cite{yang2022openood}. Note that in some cases images have to be removed from the OOD datasets if they semantically overlap with the ID dataset.}
\end{table}

\textbf{Obtaining OOD Data}. For evaluation, additional datasets are required to provide out of distribution data that is a good representation of what the model could see in the real-world.

Any dataset selected for OOD detection evaluation must, of course, be out of distribution\marginnote{You can check this by making predictions over your OOD dataset and confirm that accuracy is very low or error is very high.}, covering different semantic classes, have some kind of corruption, or lack some intersection with the ID dataset. But they must be the same format (color images, greyscale images, same input size), overall it
should be the same modality.

If you have a single classification dataset, one simple way to generate OOD data is to leave one or more classes out from the training set, and use those classes for OOD evaluation. Basically this means splitting the train, validation, and test sets, across classes. Example \ref{ood_dataset_examples} shows some commonly used dataset combinations.

\textbf{Metrics}. Since OOD detection is a binary classification task, methods produce a score that is used to identify ID and OOD samples, with likely an OOD sample obtaining a higher score than an ID sample.

Any metric meant for binary classification can be used to evaluate this task, such as precision, recall, F1 score, etc, but one metric is the most common, the area under the ROC curve. This is particularly important as the AUROC measures how separated are the two classes based on the input score, in this case the OOD detection score $s$ (in Eq. \ref{eq:ood_score}).

\section[Methods for OODD]{Methods for OOD Detection}

This section describes several commonly used methods to perform out of distribution detection. This is an active area of research.

\subsection{Model Uncertainty}

The premier method for out of distribution detection should be model uncertainty, as it directly measures model's knowledge for a specific prediction.

In classifications tasks, scores can be the maximum class probability, or the predicted distribution entropy.

For regression tasks\marginnote{OOD in Regression}, OOD detection can be performed using the standard deviation of the output distribution given by a model as a score.

\begin{examplebox}{Expectations of Model Uncertainty in OOD}{exp_ood_epistemic}
Epistemic uncertainty should be low inside the training set, likely to be zero.\\

Epistemic uncertainty should be high outside the training set, exactly signalling the out of distribution setting.
\end{examplebox}

\subsection{Training on OOD Data}
One possible approach for OOD Detection is to use an additional output head with a sigmoid activation, and
tune the extra output head to produce a high score on OOD data, and a low score on ID data.

This is basically to train a binary classifier from model features or inputs to discriminate between ID and OOD inputs.

The problem is that it fails to work due to generalization, what dataset can you use as OOD data? We provided examples in the evaluation section, and if you train a model on those datasets, it will likely perform very well in the ID/OOD classification tasks, but fail in datasets that are OOD but different from the OOD test sets. There is a generalization gap \marginnote{Generalization gap} since OOD data is an open set with almost infinite possible instances of data points.

So overall this method is not used in practice due to these gaps in generalization, and it is only advisable in settings where OOD data is well known and expected.

\subsection{Density Estimation}

Since we define ID and OOD datasets based on the training set contents, a simple method would be to use a density estimation model $\prob{X}$ on the training set, then the negative log-density is evaluated in a new input $x$ as an OOD score. For any point in the training set you would get a non-zero density, implying a non-zero but small log-density, but for points outside the training set, there would be zero density, and the log-density would very large or infinite, clearly separating ID and OOD samples.
\begin{equation}
	s(x, f) = -\log \prob{x}
\end{equation}
There are many ways to perform density estimation, for example Kernel density estimation, normalizing flows, or simply assuming a probability distribution and fitting its PDF to data.

Issues with this family of methods is that density estimation is a difficult problem that is fundamental in statistics, and often we do not know the density function or PDF for a set of data, so it becomes a model misspecification problem.
\subsection{Likelihood Ratios}

The previous section says that Density estimation on the training set is, in theory, the best OOD detection method possible, but density estimation is itself a difficult problem. \sidecite{ren2019likelihood} have shown that OOD performance with Density estimation methods is related to background statistics, and they propose a likelihood ratio.

Consider a density estimation model $\probvar{x}{\theta}$ trained on in-distribution data, which the authors call semantic component, and another density estimation model $\probvar{x}{\theta_o}$ trained on population-level background statistics, then the likelihood ratio $\text{LLR}(x)$ is defined as:
\begin{equation}
	\text{LLR}(x) = \log \frac{\probvar{x}{\theta}}{\probvar{x}{\theta_o}} = \log \probvar{x}{\theta} - \log \probvar{x}{\theta_o}
\end{equation}
\begin{marginfigure}
	\subfloat[Density Estimation]{
		\begin{tikzpicture}			
			\begin{axis}[width=\linewidth, xtick = {-1.96, 0.0, 1.64}, xlabel={Log-proba}, ylabel={Density}, domain=-2:2, xtick=\empty, ytick=\empty]
				\addplot+[red, mark=none, smooth] {gauss(0.0, 1.0)};            
				\addplot+[blue, mark=none, smooth] {gauss(0.05, 1.05)};
			\end{axis}
		\end{tikzpicture}
	}
	
	\subfloat[Likelihood Ratio]{
		\begin{tikzpicture}			
			\begin{axis}[width=\linewidth, xtick = {-1.96, 0.0, 1.64}, xlabel={Log-proba}, ylabel={Density}, domain=-3:3, xtick=\empty, ytick=\empty]
				\addplot+[blue, mark=none, smooth] {gauss(-0.5, 0.3)};            
				\addplot+[red, mark=none, smooth] {gauss(1.2, 0.8)};
			\end{axis}
		\end{tikzpicture}
	}
	\caption[Comparison of Density estimation with Likelihood ratios]{Comparison of Density estimation with Likelihood ratios, with {\color{red} ID} and {\color{blue} OOD} data being more separable when using a likelihood ratio.}
	\label{likelihood_ratios}
\end{marginfigure}
In the above definitions, $\probvar{x}{\theta}$ indicates a density estimation model with parameters $\theta$, which just makes an explicit dependency of the density model on parameters $\theta$.

\cite{ren2019likelihood} show that in a standard Density estimation model, it is common to see a problem when the distribution of log-density values overlap between ID and OOD data, which prevents its use as a OOD detection method.

The diagnosis by \cite{ren2019likelihood} is that the problem is interference by background statistics in the data, and by applying the likelihood ratio which compares the densities of semantic and background density estimation models at a given input $x$, this is much more effective at separating ID and OOD samples, which is exemplified in Figure \ref{likelihood_ratios}.

This does not fully solve the problem as to apply this method you have to discover the background statistics of your data and train a density estimation model for this.
\subsection{Energy-based Methods}

Energy-based methods in machine learning make connections with physics, and they are generative models that map inputs $x$ to non-probability scalar values called \emph{energy}. This has connections with the softmax function, and some energy-based methods have been proposed for out of distribution detection, in \sidecite{liu2020energy}.

The free energy score $E$ based on the denominator of the softmax function is given by:
\begin{equation}
	E(x, f) = -T \log \sum_i^C \text{exp}(f_i(x) / T)
\end{equation}
\begin{marginfigure}
	\begin{tikzpicture}			
		\begin{axis}[width=\linewidth, xtick = {-1.96, 0.0, 1.64}, xlabel={Free Energy}, ylabel={Frequency}, domain=-2.5:3, xtick=\empty, ytick=\empty]
			\addplot+[red, mark=none, smooth, samples=100] {gauss(-0.8, 0.15)};            
			\addplot+[blue, mark=none, smooth] {gauss(1.2, 0.7)};
		\end{axis}
	\end{tikzpicture}		
	\caption[Example of using the Free Energy score for OOD detection]{Example of using the Free Energy score for OOD detection, with {\color{red} ID} and {\color{blue} OOD} data being better separated with the prediction free energy.}
	\label{energy_score}
\end{marginfigure}
Where $T$ is a temperature value that define how uniform the output distribution is, and can be used as $T = 1$ for OOD detection.

The key experimental insight in this method is that OOD inputs usually get a lower free energy score than ID inputs, allowing for separation of ID/OOD inputs based on this score.

\section{Challenges}

I think by now it is generally clear that separating ID and OOD data distributions is difficult, particularly because of the empirical observation that there are ID samples that have high uncertainty, due to them being just difficult or mislabeled samples, and there are OOD samples that have low uncertainty, most likely due to approximation errors and other silly mistakes and assumptions in the OOD methods. Modelling the variability in data is a difficult problem.

Overall OOD detection is not a solved problem and I believe we are far from general solution that works most of the time, more research is needed. This is one of the reasons why ChatGPT and other LLMs produce hallucinations, and they are also difficult to detect, and there is no obvious solution that can be applied by Industry.

As shown by \cite{ovadia2019can}, there are no guarantees for uncertainty estimation performance, and real-world distribution shifts have to be considered in the lifecycle of a model.

\begin{marginfigure}
	\begin{tikzpicture}
		\begin{axis}[ymax=5, ymin=0, xmin=0.0, xmax=2.5, minor y tick num = 3, ylabel={Density}, xlabel={Entropy}, width=\linewidth, xtick=\empty, ytick=\empty]
			\addplot+[green, ybar, fill, mark=none, bar width=0.5] coordinates { (0, 4.5) (0.5, 0.5)};
			\addplot+[red, mark=none, smooth, samples=100] {gauss(1.3, 0.8)};
		\end{axis}
	\end{tikzpicture}
	\caption{Distribution of Entropy between ID and OOD datasets for MNIST vs Fashion MNIST, noting how there are OOD examples with very low uncertainty.}
	\label{ood_mnist_fminst_density}
\end{marginfigure}
As a quick example of OOD challenges, we trained a neural network with MC-Dropout on MNIST, which is a very simple and introductory task, and then evaluated it on Fashion MNIST, which is an OOD dataset where the MNIST model cannot make correct predictions.

\begin{figure*}[t]
	\subfloat[MNIST - In Distribution]{
		\includegraphics[width=\linewidth]{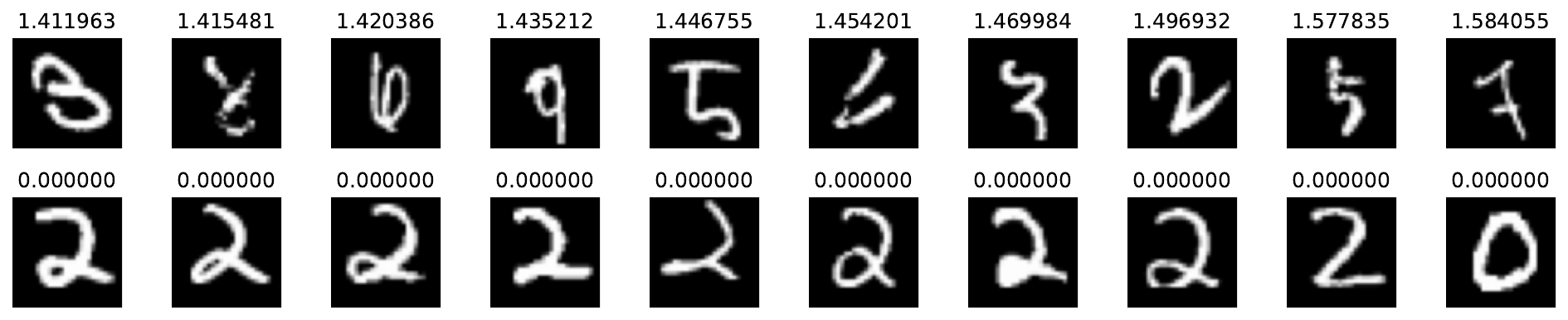}
	}
	
	\subfloat[Fashion MNIST - Out of Distribution]{
		\includegraphics[width=\linewidth]{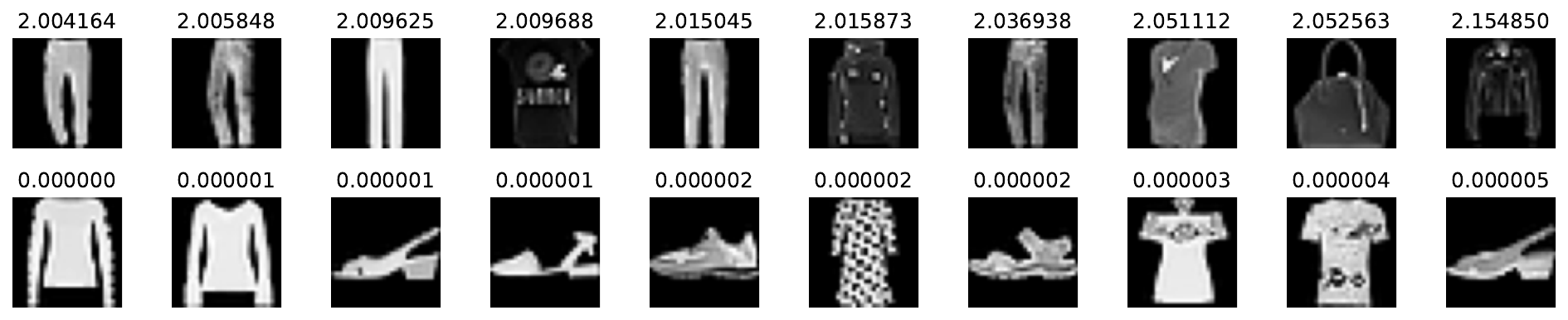}
	}
	\caption{Comparison of predicted entropy in a model trained on MNIST and evaluated on Fashion MNIST using MC-Dropout. This figure presents samples with highest (top) and lowest entropy (bottom) in each dataset, showing that there are always in-distribution samples with high uncertainty, and out of distribution samples with low uncertainty.}
	\label{ood_mnist_fminst_examples}
\end{figure*}

Figure \ref{ood_mnist_fminst_density} shows the entropy distribution of predicted probabilities for both datasets, noting that while there is no total separation between ID and OOD, there is partial separation and minimal overlap, leading to AUROC of 0.97. Then Figure \ref{ood_mnist_fminst_examples} shows qualitative examples for both datasets, a selection of minimum and maximum entropy. It is clear that in both datasets there are examples with high and low entropy, implying that some OOD examples still receive very low uncertainty, and some ID examples are naturally uncertain. In the case of MNIST, some uncertain examples are digits that are malformed or very different to the most common digits (intra-class variability), while the OOD examples with low uncertainty are simply mistakes made by the model.

MNIST and Fashion MNIST are simple datasets used for academic research, and even in this simple example it is not possible to completely separate ID and OOD examples. Results with other uncertainty methods like Ensembles are very similar.

%% file: figures/trainset_holes.tikz
\begin{tikzpicture}
    \filldraw[fill=gray, pattern=north east lines, pattern color=gray!70, name path=mainpath, even odd rule] plot [smooth cycle, tension=0.99]
    coordinates {(-.5,1.5)(.5,.5)(1.5,-0.7)(3,0)(4,2)(2,3)}
    (2.0, 2.0) circle (0.2)
    (1.0, 0.3) circle (0.1)
    (2.6, 0.5) circle (0.4);
\end{tikzpicture}

%% file: figures/id_ood_difficulty.tikz
\begin{tikzpicture} 
    \coordinate (a1) at (0,0);
    \coordinate (a2) at (3.3,-0.3);
    \coordinate (a3) at (4.5,0.5);
    \coordinate (a4) at (1.7,1.4);
    \coordinate (a5) at (3,3);
    \coordinate (a6) at (2.4,3.2);
    \coordinate (a7) at (1,2);
    \coordinate (O) at ($(a1)-(1,2)$);
    \draw[use Hobby shortcut, thick] ([out angle=-90]a1)..(a2)..([in angle=-90]a3); 
    \draw[use Hobby shortcut, thick] ([out angle=90]a3)..(a4)..(a5)..(a6)..(a7)..([in angle=90]a1);
        
    \fill[fill=gray, pattern=north east lines, pattern color=gray!70, use Hobby shortcut] ([out angle=-90]a1)..(a2)..([in angle=-90]a3);
    \fill[fill=gray, pattern=north east lines, pattern color=gray!70, use Hobby shortcut] ([out angle=90]a3)..(a4)..(a5)..(a6)..(a7)..([in angle=90]a1);
        
    %OOD Markers
    \draw plot[only marks, mark=x, mark size=4pt] (2.2, 1.6) node[right=5pt] {Near OOD};
    \draw plot[only marks, mark=x, mark size=4pt] (2.3, -0.9) node[right=5pt] {Near OOD};
        
    \draw plot[only marks, mark=x, mark size=4pt] (6.1, 1.0) node[below=5pt] {Far OOD};
    
    \draw plot[only marks, mark=x, mark size=4pt] (2.5, 6.0) node[below=5pt] {Far OOD};
        
    %ID Markers
    \fill (1, 0.6) circle (4pt) node[below=5pt] {ID};
\end{tikzpicture}

%% file: sections/appendix-distributions.tex
\chapter{Review of Probability Distributions}
\margintoc

\section{Probability}

\begin{definitionbox}{Probability}{}
    A probability is defined for an event as the ratio of possible ways it can happen to the total number of outcomes.    
    \begin{equation}
        P(E) = \frac{\text{\# of ways } E \text{ can happen}}{\text{\# of total outcomes}}
    \end{equation}    
    Probabilities are real numbers in the $[0, 1]$ range, and are interpreted as likelihood or chance that the event will happen.\\
    A simple example can be made with a six-sided die, P(D is odd) = P(even) = 1/2, P(D = 1) = 1/6, P(D $>$ 2) = 4/6, etc.
\end{definitionbox}

\begin{definitionbox}{Conditional Probability}{}
    It is the probability of an event $A$ given that another event $B$ has already occurred, which takes the assumption that both events have some sort of relationship.\\

    This is denoted by $\condprob{A}{B}$ and computed by definition as:
    \begin{equation}
        \condprob{A}{B} = \frac{\prob{A \cap B}}{\prob{B}}
    \end{equation}
    
    Note that in general $\prob{A} \neq \condprob{A}{B} \neq \condprob{B}{A} \neq \prob{B}$.

	\textbf{Independence}. Events are independent if there is no relationship between them, implying that:
	\begin{equation}
		\prob{A \cap B} = \prob{A}\prob{B}\qquad \text{i.i.f $A$ and $B$ are independent.}
	\end{equation}
    Only if $\condprob{A}{B} = \prob{A}$ and/or $\condprob{B}{A} = \prob{B}$, this indicates that $A$ and $B$ are independent.
\end{definitionbox}
Some basic properties and identities of probability are given below.
\begin{table}
    \begin{tabular}{ll}
        \textbf{Event}	& \textbf{Probability}\\
        E			& $\prob{E} \in [0, 1]$\\
        not E		& $\prob{\text{not } E} = 1 - \prob{E}$\\
        & \\
        A or B		& $\prob{A \cup B} = \prob{A} + \prob{B} - \prob{A \cap B}$\\
        & $\prob{A \cup B} = \prob{A} + \prob{B}$ if A and B are mut. exclusive.\\
        & \\
        A and B		& $\prob{A \cap B} = \condprob{A}{B} \prob{B} = \condprob{B}{A} \prob{A}$\\
        & $\prob{A \cap B} = \prob{A} \prob{B}$ if A and B are independent.\\
        & \\
        A given B	& $\condprob{A}{B} = \frac{\prob{A \cap B}}{\prob{B}} = \frac{\condprob{B}{A} \prob{A}}{\prob{B}}$ (Bayes Rule)
    \end{tabular}
    \caption{Common properties of probability.}
\end{table}

\section{Univariate Probability Distributions}

A probability distribution is a function that maps from some input variable to probability (discrete) or probability density (continuous) values, as a way to characterize how probability is distributed on the variable.

The input variable is called a \emph{Random Variable}, because its value depends on a random outcome or randomness. The possible values of the variable are events, which are subset of a \emph{sample space}, which is any predefined set (of any kind, real numbers, matrices, vectors, etc).

\begin{definitionbox}{Random Variable}{}
    A random variable is a mathematical concept used to model randomness, where it is a object which depend on a random event.    
    \begin{description}
        \item[Sample Space] A random value has a set of allowable values, usually called sample space and denoted by $\mathcal{A}$.
        \item[Mapping] A random variable is a mapping $X: \mathcal{A} \rightarrow \mathbb{R}$, where $\mathcal{A}$ are possible outcomes in the sample space, mapped to real numbers (their probability). 
    \end{description}    
    Random variables can be discrete or continuous, which defines their sample space.
\end{definitionbox}

\begin{definitionbox}{Probability Function}{}
    The PF is a function that usually defines a probability distribution, and is defined as:
    \begin{equation}
        f_X(x) = \prob{X = x}
    \end{equation}
    This is very similar to the mapping $X$ we defined previously in a random variable. This definition works well for discrete random variables, but it is problematic for continuous ones, because $\prob{X = x} = 0$ for all $x$.\\

    For discrete distributions, this is usually called the probability mass function.
\end{definitionbox}

Every probability distribution must follow Kolmogorov's Axioms.
\begin{description}
    \item[Probability values are between 0 and 1] \hfill \\
    \begin{equation*}
    	0 \leq \prob{X \in E} \leq 1 \qquad \forall E \in \mathcal{A}
    \end{equation*}
    \item[Sum of all events is 1] \hfill \\
    \begin{equation*}
    	\sum_{X \in \mathcal{A}} \prob{X} = 1 \qquad \int_{\mathcal{A}} P(X) dX = 1
    \end{equation*}
    \item[Disjoint Family of Sets] \hfill \\
    \begin{equation*}
    	\prob{X \in \cup_i E_i} = \sum_i \prob{X \in E_i}
    \end{equation*}
    For any disjoint family of sets $E_i \in \mathcal{A}$
\end{description}

\begin{definitionbox}{Probability Density Function}{}
    For continuous probability distributions, the PDF is a function that usually defines a probability distribution, and is defined as $f_X(x)$ that follows:
    \begin{equation}
        \prob{a \leq X \leq b} = \int_a^b f_X(x) dx
    \end{equation}
    
    For any $a \leq b$. PDFs model probability density instead of plain probability, so they can (and will be) bigger than one. This does not violate Kolmogorov's Axioms.
    
    Note that if $a = b$, then:
    \begin{equation}
        \prob{a \leq X \leq a} = \int_a^a f_X(x) dx = 0
    \end{equation}
\end{definitionbox}

\begin{definitionbox}{Cumulative Distribution Function (CDFs)}{}
    The cumulative distribution function (CDF) is a function defined by:
    \begin{equation}
        F_X(x) = \prob{X \leq x}
    \end{equation}    
    This function gives the probability that a random variable $X$ is less than a value $x$. A special property is:    
    \begin{equation}
        \prob{a < X \leq b} = F_X(b) - F_X(a)
    \end{equation}
\end{definitionbox}

Continuous PDFs and CDFs have a kind of duality, related by the derivative integral, as shown below:

\begin{equation}
    F_X(x) = \int_{-\infty}^{x} f_X(u) du
\end{equation}
\begin{equation}
    f_X(x) = \frac{d}{dx} F_X(x)
\end{equation}

\begin{definitionbox}{Inverse Cumulative Function or Quantile Function}{}
    Another important function is the inverse CDF, also known as quantile function:
    \begin{equation}
        F^{-1}(q) = \inf\{x: F(x) > q \}
    \end{equation}
    Where $q \in [0, 1]$ is a specific quantile. The $\inf$ method can be conceptually thought as taking the minimum.\\

    If $F(x)$ is a strictly increasing function, then $F^{-1}(q)$ is the unique value $x$ so $F(x) = q$ holds.\\

    Quantiles are equal divisions of the probability space, some have special names, for example percentiles divide in 100 divisions, or quartiles over 4 divisions. The percentile 50 is always the median ($F^{-1}(0.5)$).
\end{definitionbox}

\section{Algebra of Random Variables}
Expectation or expected value \marginnote{Expectation} is a linear operation defined for continuous distributions with PDF $X \sim f_X$ as:
\begin{equation}
    \expect{X} = \int_{-\infty}^{\infty} x \, f_X(x)dx
\end{equation}
And for discrete distributions as:
\begin{equation}
    \expect{X} = \sum_{i=0}^{\infty} x_i \, f_X(x_i)
\end{equation}
Using the expectation, two important properties of a probability distribution can be defined.
The expected value $\expect{X}$ is associated with the mean. While the variance of a distribution can be computed as:
\marginnote{Variance}
\begin{equation}
     \var{X} = \expect{X^2} - (\expect{X})^2
\end{equation}
The variance measures the distribution spread around the mean. The mean is usually denoted by $\mu$, and the variance by $\sigma^2$, the standard deviation is denoted as $\sigma$, that is, the square root of the variance.

Some important identities and transformations on random variables $X$ and $Y$:
\begin{align*}
    \var{X + Y} &= \var{X} + 2 \cov{X, Y} + \var{Y}\\
    \var{X - Y} &= \var{X} - 2 \cov{X, Y} + \var{Y}\\
    \var{X} 	&= \cov{X, X} = \expect{X^2} - (\expect{X})^2
\end{align*}
For a constant $k$:
\begin{align*}
    \var{X + k} &= \var{X}\\
    \var{k X}  &= k^2 \var{X}\\
\end{align*}
The law of total variance\marginnote{Law of Total Variance} relates the variance of a random variable $Y$ with the conditional variance.
\begin{equation}
    \var{Y} = \expect{\var{Y \, | \, X}} + \var{\expect{X \, | \, Y}}
\end{equation}
While similarly the law of total expectation\marginnote{Law of Total Expectation} connects with the conditional expectation.
\begin{equation}
    \expect{X} = \expect{\expect{X \, | \, Y}}
\end{equation}

\begin{definitionbox}{Addition of Probability Distributions}{}
    One important result in statistics is the addition/sum of two random variables, which is the convolution of their PDFs. If we have $X \sim f_X$, $Y \sim f_Y$, then $Z = X + Y$ has PDF:    
    \begin{equation}
        f_Z(z) = \int_{-\infty}^{\infty} f_X(x) f_Y(z - x) dx
    \end{equation}
    Which is the convolution of the PDFs, denoted by $\star$:
    \begin{equation}
        f_Z =  f_X \star f_Y
    \end{equation}
\end{definitionbox}

\begin{definitionbox}{Transformations on Random Variables}
    Passing random variables through a function is in general difficult to do, for example, if we have $X$ a distribution with PDF $f_X$, and a function $g$ that is increasing in the domain of $X$, then the PDF of $Y = g(X)$ is given by:    
    \begin{equation}
        f_Y(y) = f_X(g^{-1}(y))\left|\frac{d}{dy} g^{-1}(y)\right|
    \end{equation}
    This is of course only valid for continuous distributions. For a discrete distribution $X$ with probability mass function $f_X$, then the transformation is:
    \begin{equation}
        f_Y(y) = \sum_{x \in g^{-1}(y)} f_X(x)
    \end{equation}
\end{definitionbox}

\section{Multivariate Random Variables}

Random variables can also take the form of vectors, which are called random vectors: $X = (X_1,  2..., X_n)$, where $X_i$ is a random variable.

Distributions over random vectors are defined by the joint probability density function\marginnote{Joint Distribution}, or simply the joint distribution:
\begin{equation}
	\prob{(X_1, ..., X_n) \in A} = \int_A f_{X_1, ..., X_n}(x_1, ..., x_n) dx_1, ..., dx_n
\end{equation}
A marginal distribution\marginnote{Marginal Distribution} is the probability distribution for a subset of random variables in a random vector. In particular for a random vector, the density functions $f_{X_1}, ..., f_{X_n}$ for the univariate random variables $X_1, ..., X_n$ are called marginal distributions, or just marginals.

An important property related to marginals is marginalization\marginnote{Marginalization}, meaning that we can obtain the marginal distribution of one variable by taking an expectation of the joint distribution of all other variables:
\begin{equation}
	f_{X_1}= \int f_{X_1, ..., X_n}(X_1, ..., X_n) d(X_2, ..., X_n)
\end{equation}
For the two-dimensional case with random vector $(X, Y)$:
\begin{equation}
	f_X(x) = \int f_{X, Y}(x, y)dy
\end{equation}
\begin{equation}
	f_Y(y) = \int f_{X, Y}(x, y)dx
\end{equation}
The joint cumulative distribution function $F$\marginnote{Joint CDF} for a random vector is defined as:
\begin{equation}
	F_{X_1, ..., X_n}(x_1, ..., x_n) = \prob{X_1 \leq x_1, ..., X_n \leq x_n}
\end{equation}

Covariance is the generalization of variance for random vectors. Covariance is a measure of the joint variation between the random variables inside the random vector, and for random variables $X$ and $Y$ is defined as:
\begin{equation}
    \cov{X,Y} = \expect{(X - \expect{X})(Y - \expect{Y})}
\end{equation}
Covariance can be alternatively computed as:
\begin{equation}
    \cov{X,Y} = \expect{XY} - \expect{X}\expect{Y}
\end{equation}
Covariance is a matrix, where elements in the diagonal are variances $\sigma^2_{ii}$, and elements outside the diagonal are covariances between the corresponding $X_i$ and $X_j$ variables $\sigma^2_{ij}$.

The Covariance matrix has some important properties. It is a positive semi-definite matrix, which is conceptually similar to being positive for the real numbers, but this does not mean that elements $\Sigma_{ij}$ are positive.

In particular, if element $\Sigma_{ij}$ is positive, this indicates variables $X_i$ and $X_j$ are positively correlated, while a negative value indicate negative correlation.

There are multiple types of Covariance matrices. A Diagonal Covariance is one where $\Sigma_{ij} = 0$ for $i \neq j$, meaning that only diagonal terms are non-zero, while a Spherical Covariance is a Diagonal Covariance matrix where all diagonal covariances are equal.

If covariance is normalized with the standard deviation of each variable ($\sigma_X$, $\sigma_Y$), you obtain the correlation:
\begin{equation}
    \text{CORR}(X, Y) = \frac{\expect{(X - \expect{X})(Y - \expect{Y})}}{\sigma_X \sigma_Y}
\end{equation}

\section{Common Distributions}

\begin{definitionbox}{Gaussian Distribution}{}
    It is a univariate continuous distribution defined for $x \in \mathbb{R}$ with two parameters, mean $\mu$ and variance $\sigma^2$, and with PDF given by:
    \begin{equation}
        f_X(x) = \frac{1}{\sigma \sqrt{2 \pi}} e^{-0.5 (\frac{x-\mu}{\sigma})^2}
    \end{equation}
    The expectation and variance of the Gaussian distribution $X \sim \mathcal{N}(\mu, \sigma^2)$ is:
    \begin{align*}
        \expect{X} &= \mu\\
        \var{X} &= \sigma^2
    \end{align*}
    The multivariate version of the Gaussian distribution, defined for the random vector $\mathbf{x} \in \mathbb{R}^N$, with parameters mean vector $\mu$ and covariance matrix $\Sigma$ has PDF defined as:
    \begin{equation}
        f_X(x) = (2\pi)^{-n/2} \text{det}(\Sigma)^{-1/2} \text{exp}\left(-\frac{1}{2} (x - \mu)\Sigma^{-1}(x - \mu)\right)
    \end{equation}
\end{definitionbox}

\begin{definitionbox}{Uniform Distribution}{}
    Continuous or Discrete distribution where all values in a range $[a, b]$ are equally likely. It is parameterized by $a$ and $b$ and has PDF/mass fn as:
    \begin{equation}
        f_X(x) = \begin{cases}
            \frac{1}{b - a} & \text{ if } a \leq x \leq b\\
            0	& \text{ otherwise}
        \end{cases}
    \end{equation}
    
    It is usually denoted as $X \sim U(a, b)$.\\
    
    The mean is $\expect{X} = 0.5(a + b)$ and variance is $\var{X} = \frac{1}{12}(b - a)^2$.
    
\end{definitionbox}

\begin{definitionbox}{Binomial Distribution}{}
	It is a discrete distribution for the event: number of successes in a sequence of $n$ binary experiments, each with success probability $p$, and thus failure probability $1-p$. Its probability mass function is given by:
	\begin{equation}
		f_X(k) = \binom{n}{k} p^k(1-p)^{n-k}
	\end{equation}
	Where $\binom{n}{k}$ is the binomial coefficient, which can be computed as $\binom{n}{k} = \frac{n!}{k!(n-k)!}$.\\
	
	The mean is $\expect{X} = np$ and variance is $\var{X} = np(1-p)$.
\end{definitionbox}

\begin{definitionbox}{Categorical Distribution}{}
	This is a discrete distribution over an unordered set of $C$ categories, with each category having probability $p_k$, and constraint that $\sum_{k \in C} p_k = 1$. The PMF is given by:
	\begin{equation}
		f_X(k) = \prob{X = k} = p_k
	\end{equation}		
	For two categories ($C = 2$), it is equivalent to the Bernoulli distribution with parameters $p_0, p_1 = 1 - p_0$.\\
	
	The discrete uniform distribution is the Categorical distribution with $p_k = 1/C$.
\end{definitionbox}

\begin{definitionbox}{Empirical Distribution}{}
	The empirical distribution is the one that can be estimated from data samples. Given i.i.d random variables $X_1, ... X_n$, then the empirical CDF is:
	\begin{equation}
		\hat{F}_n(t) = \frac{\sum_i \mathbb{1}[X_i \leq t]}{n} = \frac{\text{number of elements in sample} \leq t}{n}
	\end{equation}
	Note how we define this distribution with its CDF, instead of the normal PDF. The PDF can be visualized with a density histogram, but it is prone to bias-variance trade-off when tuning the number of histogram bins. See Section \ref{appendix:numbins} for details.
\end{definitionbox}

\section{Important Theorems}

\begin{definitionbox}{Law of Large Numbers}{}
	Given a sequence of random variables $X_1, ..., X_n$ that are independent and identically distributed with expected value $\expect{X_1} = ... = \expect{X_n} = \mu$. Then the sample average $\bar{X_n}$:
	\begin{equation}
		\bar{X}_n = \frac{X_1 + ... + X_n}{n}
	\end{equation}
	Converges to the expected value as $n \rightarrow \infty$:
	\begin{equation}
		\lim_{n \rightarrow \infty} \bar{X}_n = \mu
	\end{equation}
	This law can be interpreted that in i.i.d. random variables, more number of samples implies that the sample average becomes closer to the true mean.\\
	
	One major application of the Law of Large Numbers is \textbf{Monte Carlo sampling}, as we increase the number of samples, the sample average gets closer to the true mean.
\end{definitionbox}

\begin{definitionbox}{Central Limit Theorem}{}
	Given a sequence of random variables $X_1, ..., X_n$ that are independent and identically distributed with population mean $\mu$ and variance $\sigma^2$. First let's define the sample mean or average:
	\begin{equation}
		\bar{X}_n = \frac{X_1 + ... + X_n}{n}
	\end{equation}
	The central limit theorem states that as $n \rightarrow \infty$, then the distribution of $\bar{X}_n$ is a Gaussian distribution:    
	\begin{equation}
		\bar{X}_n \sim \mathcal{N}\left(\mu, \frac{\sigma^2}{n}\right)
	\end{equation}
	That is the variance of the sample mean decreases with $\frac{1}{n}$, while the standard deviation decreases with $\frac{1}{\sqrt{n}}$.
	The CLT is a fundamental theorem in statistics and probability, as it describes why many real-world random variables are Gaussian distributed.
\end{definitionbox}

\section{Sampling}

\begin{definitionbox}{(Random) Sampling}{}
    Sampling is the process of generating values that follow a given probability distribution. This is done with random number generators and mathematics (that we do not cover).
    \begin{equation}
        x \sim f_X(\text{parameters})
    \end{equation}
    Where $x$ is a sample from distribution density function $f_X$ with given parameters.\\
    
    An important and often misunderstood concept, is that values produced by sampling only follow the distribution in aggregation, which can be visualized by making histograms of many samples.
\end{definitionbox}

\begin{definitionbox}{Density Estimation as Opposite of Sampling}{}
    Now imagine the opposite problem of sampling, we have many values or samples $x$ from a unknown distribution $f_X$, and we would like to estimate the distribution $f_X$ from these samples.\\
    
    Density estimation is the problem of estimating the PDF $f_X$ from i.i.d samples, without knowing the distribution that generated the data. This is unsupervised as there are no labels, only data points. It is one way to learn structure from data.\\
    
    A clear use case is generative models, we would like to learn the distribution that generated some data points, by only having access to those data points.  
\end{definitionbox}

\section{Estimation Theory}

Estimation theory is a field in Statistics about estimating parameters of probability distributions from data samples. We have seen already Maximum Likelihood estimation as the basis to derive some loss functions based on particular probability distributions.

Estimator Theory builds \textit{estimators}, which are functions that can estimate a given parameter of a probability distribution from data, and thus estimators are also random variables and can be studied with the tools we have previously seen.

\begin{definitionbox}{Statistic}{}
	A statistic\footnote{not to be confused with the field of Statistics}, or sample statistic, is a quantity computed as function from a sample of data.
\end{definitionbox}
Some examples of commonly used statistics are the sample mean and sample variance. For i.i.i data points $x_i$ samples from a given but unknown distribution, then the sample mean\marginnote{Sample Mean} is:
\begin{equation}
	\mu(x_i) = \frac{1}{N} \sum_i x_i
\end{equation}
And the sample variance\marginnote{Sample Variance} can be computed as:
\begin{equation}
	\sigma^2(x_i) = \frac{1}{N-1} \sum_i (x_i - \mu_x)^2
\end{equation}
And the sample Covariance matrix\marginnote{Sample Covariance} with elements $\cov{X}_{jk}$:
\begin{equation}
	\cov{X}_{jk} = \frac{1}{N-1} \sum (X_{ij} - \bar{X}_j)(X_{ik} - \bar{X}_k)
\end{equation}

\begin{definitionbox}{Estimator}{}
	An Estimator is a statistic related to a quantity, usually a parameter, of a probability distribution. An estimator for a parameter $\theta$ is usually denoted as $\hat{\theta}$.
\end{definitionbox}

\begin{definitionbox}{Unbiased Estimator}{}
	The bias of an estimator $\hat{\theta}$ is:
	\begin{equation}
		\text{Bias}(\hat{\theta}, \theta) = \expect{\hat{\theta} - \theta}
	\end{equation}
	An unbiased estimator is one with bias equal to zero, or equivalently:
	\begin{equation}
		\expect{\hat{\theta}} = \theta
	\end{equation}
\end{definitionbox}

The squared error of an estimator has a decomposition based on the bias and variance:
\begin{equation}
	\expect{(\hat{\theta} - \theta)^2} = \text{Bias}(\hat{\theta}, \theta)^2 + \text{Var}(\theta)
\end{equation}

\section{Confidence Intervals}
Confidence intervals are representations of uncertainty with an interval that is $\alpha$ \% likely to contain the true value.
\begin{equation}
	\prob{l_\alpha \leq y \leq u_\alpha} = \alpha
\end{equation}
Where $[l_\alpha, u_\alpha]$ are the bounds of the confidence interval, which is a function of the confidence level $\alpha$, and predicted mean $\hat{\mu}_i$ and variances $\hat{\sigma}^2_i$.\\

So let's assume that $F$ is the CDF of a Gaussian distribution. Then the computation of a $\alpha$ confidence interval has a pseudo-closed form:
\begin{align}
	l = \mu - |z_{\frac{\beta}{2}}| \sigma\\
	u = \mu + |z_{\frac{\beta}{2}}| \sigma
\end{align}
Where  $\beta = 1 + \alpha$ and $z_{\frac{\beta}{2}}$ is the $z$ score \marginnote{The z-score is given by $z = \frac{x - \mu}{\sigma}$} corresponding to the $\frac{\beta}{2}$ quantile. This can be computed with the quantile function (inverse of the CDF, $z = F^{-1}(\frac{\beta}{2})$), but for Gaussian distributions there are standard tables that you can use.

%% file: sections/appendix-derivation.tex
\chapter{Derivation of Losses}
\margintoc

\section{Gaussian Negative Log-Likelihood}

Assume that we have a ground truth $y$ and a model that predicts a Gaussian distribution with parameters mean $\mu(x)$ and variance $\sigma^2(x)$, and assume data points $y$ are i.i.d. Then we build the likelihood function as the product of the Gaussian PDF given those parameters, assuming the ground truth $y$ as true mean:
\begin{equation}
    L(y, \mu(x),\sigma^2(x)) = \prod_{i} (2 \pi \sigma_i^2)^{-\frac{1}{2}} e^{-\frac{1}{2} (\frac{y_i-\mu_i}{\sigma_i})^2}
\end{equation}
Then we build the log-likelihood by taking the logarithm on both sides:
\begin{align*}
    \log L(y, \mu(x),\sigma^2(x)) &= \sum_{i} \log \left((2 \pi \sigma_i^2)^{-\frac{1}{2}} e^{-\frac{1}{2} (\frac{y_i-\mu_i}{\sigma_i})^2}\right)\\
    &= \sum_{i} -\frac{1}{2} (\log 2 \pi \sigma^2) +\log e^{-\frac{1}{2} \left(\frac{y_i-\mu_i}{\sigma}\right)^2}\\
    &= \sum_{i} -\frac{1}{2} (\log 2 \pi \sigma^2) -\frac{1}{2} \left(\frac{y_i-\mu_i}{\sigma}\right)^2\\
    &= \sum_{i} -\frac{1}{2} (\log 2 \pi + \log \sigma^2) -\frac{1}{2} \left(\frac{y_i-\mu_i}{\sigma}\right)^2
\end{align*}
Now this is a log-likelihood that we want to maximize, but in general we want a loss function to minimize. Maximizing $L$ is equivalent to minimizing $-L$, so we flip the sign:
\begin{align*}
    -\log L(y, \mu(x),\sigma^2(x)) &= \sum_{i} 0.5 (\log 2 \pi + \log \sigma_i^2) +0.5 \left(\frac{y_i-\mu_i}{\sigma_i}\right)^2
\end{align*}
Now if we remove all constant terms (they do not contribute to gradients and do not change the optimal solution), we arrive at:
\begin{equation}
    -\log L(y, \mu(x),\sigma^2(x)) = \sum_{i} \log \sigma_i^2 + \frac{(y_i - \mu_i)^2}{\sigma_i^2} \hspace*{3.0cm}
\end{equation}
This loss is called the Gaussian Negative Log-Likelihood.

%% file: sections/appendix-numbins.tex
\chapter{Setting Number of Histogram Bins}
\label{appendix:numbins}

\begin{figure*}[!ht]
	\subfloat[$|B| = 5$]{
		\begin{tikzpicture}
			\begin{axis}[ybar, ymin=0, width=0.25\linewidth, ylabel={Density}, xlabel={}, xtick=\empty, ytick=\empty]
				\addplot+[hist={bins=5}] table [y index=0] {data/normal_data.csv};
			\end{axis}
		\end{tikzpicture}
	}
	\subfloat[$|B| = 10$]{
		\begin{tikzpicture}
			\begin{axis}[ybar, ymin=0, width=0.25\linewidth, xlabel={}, xtick=\empty, ytick=\empty]
				\addplot+[hist={bins=10}] table [y index=0] {data/normal_data.csv};
			\end{axis}
		\end{tikzpicture}
	}
	\subfloat[$|B| = 15$]{
		\begin{tikzpicture}
			\begin{axis}[ybar, ymin=0, width=0.25\linewidth, xlabel={}, xtick=\empty, ytick=\empty]
				\addplot+[hist={bins=15}] table [y index=0] {data/normal_data.csv};
			\end{axis}
		\end{tikzpicture}
	}
	\subfloat[$|B| = 20$]{
		\begin{tikzpicture}
			\begin{axis}[ybar, ymin=0, width=0.25\linewidth, xlabel={}, xtick=\empty, ytick=\empty]
				\addplot+[hist={bins=20}] table [y index=0] {data/normal_data.csv};
			\end{axis}
		\end{tikzpicture}
	}
	\subfloat[$|B| = 25$]{
		\begin{tikzpicture}
			\begin{axis}[ybar, ymin=0, width=0.25\linewidth, xlabel={}, xtick=\empty, ytick=\empty]
				\addplot+[hist={bins=25}] table [y index=0] {data/normal_data.csv};
			\end{axis}
		\end{tikzpicture}
	}
	\subfloat[$|B| = 30$]{
		\begin{tikzpicture}
			\begin{axis}[ybar, ymin=0, width=0.25\linewidth, xlabel={}, xtick=\empty, ytick=\empty]
				\addplot+[hist={bins=30}] table [y index=0] {data/normal_data.csv};
			\end{axis}
		\end{tikzpicture}
	}
	\caption[Example of a histogram built on a random sample of a standard Gaussian distribution as the number of bins $|B|$ is varied]{Example of a histogram built on a random sample of a standard Gaussian distribution as the number of bins $|B|$ is varied, showing the trade-off between oversmoothing and undersmoothing. Which value of $|B|$ do you think best represents the underlying data distribution?}
	\label{tradeoff_histogram_bins}
\end{figure*}
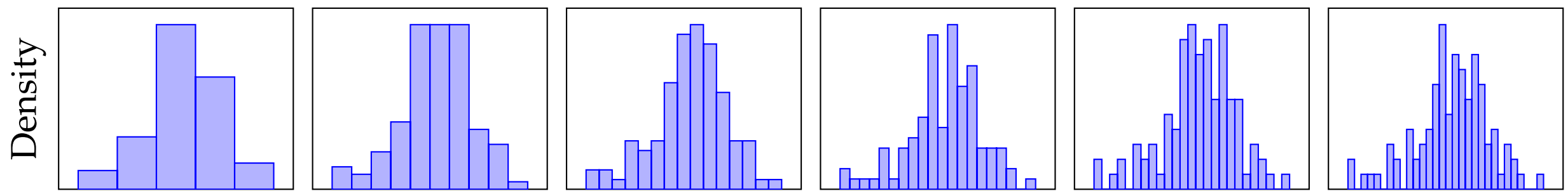

The number of bins in a histogram, denoted by $|B|$, is a hyperparameter that must be set by the user, and it is very relevant to computation of calibration plots and calibration error. If $|B|$ is too low, then the histogram is oversmoothed, and if $|B|$ is too large, then the histogram is undersmoothed. There is a trade-off between these two extremes which is very similar to a bias-variance trade-off. This is shown in Figure \ref{tradeoff_histogram_bins}.

\begin{marginfigure}
	\begin{tikzpicture}
		\begin{axis}[ybar, ymin=0, width=\linewidth, xlabel={}, ylabel={Density}, xtick=\empty, ytick=\empty]
			\addplot+[hist={bins=100}] table [y index=0] {data/normal_data.csv};
		\end{axis}
	\end{tikzpicture}
	\caption{Example of extreme undersmoothing, obtained with $|B| = 100$.}
	\label{extreme_undersmoothing}
\end{marginfigure}
It is possible to obtain misleading histograms that do not represent the true distribution of the data if $|B|$ is not properly set, and this can lead to misleading calibration error and plots. An example of extreme undersmoothing is available in Figure \ref{extreme_undersmoothing}, with its characteristic being many bins with zero samples.

An important question is then how to set the number of histogram bins? There are many methods in literature but they are all heuristics that are not guaranteed to always work. For a quick summary please refer to (\sidecite{crossvalidOptNumBins}) There is no single method that works all the time. What is recommended is to look at the histogram, vary the value of $|B|$, and decide visually what "looks best". The problem with this method is that you often do not know the true data distribution, and it is easy for a human to mislead themselves while setting this parameter.

An option to making histograms, is to perform density estimation, but these still need a parameter to be set.

\begin{marginfigure}
	\begin{tikzpicture}
		\begin{axis}[width = \linewidth, xlabel={Confidence}, ylabel={Accuracy}, xmin = 0.0, xmax = 1.0, ymin = 0.0, ymax = 1.0, ymajorgrids=false, xmajorgrids=false, grid style=dashed, legend pos = north east, legend style={font=\scriptsize}, x tick label style={font=\scriptsize, rotate=90}, y tick label style={font=\scriptsize}, name=main axis, width=\linewidth]
			
			\addplot+[mark = none, gray, dashed] coordinates { (0, 0) (1, 1)};
			\addplot+[mark = none, blue!30,ybar,fill] coordinates { (0.0,0.0) (0.1, 0.05) (0.2, 0.15) (0.3, 0.2) (0.4, 0.15) (0.5, 0.35) (0.6, 0.48) (0.7, 0.55) (0.8, 0.5) (0.9, 0.8)};
			
		\end{axis}
		
		\begin{axis}[ytick=\empty, xtick=\empty, width=136pt, height=2cm, anchor=south west, at=(main axis.north west), hide axis]						
			\addplot+[mark = none, gray!50, ybar, fill] coordinates { (0.0,0.1) (0.1, 0.1) (0.2, 0.15) (0.3, 0.2) (0.4, 0.15) (0.5, 0.2) (0.6, 0.1) (0.7, 0.1) (0.8, 0.2) (0.9, 0.4)};			
		\end{axis}
	\end{tikzpicture}
	\caption{Example of Calibration plot showing confidence distribution on top to guide the reader.}
	\label{calibration_plot_density}
\end{marginfigure}
For a calibration plot and computation of calibration error, it is not important to obtain an "optimal" value for $|B$|, what matters is to make sure all bins have an approximate equal number of samples and that there are no bins with zero samples, but these guidelines are not always possible, for example if a model never predicts confidences in some ranges.

For calibration plots, showing the distribution of confidences in the X axis, can help interpret the calibration plot in case that the confidence distribution is skewed, an example is provided in Figure \ref{calibration_plot_density}. This shows which bins have more or less samples and can reveal if the calibration plot is based on an incorrect histogram.